\documentclass[acmlarge,screen,10pt]{acmart}

\newcommand{\ignore}[1]{ }

\usepackage{amsmath,amsfonts}
\usepackage{algorithmic}
\usepackage{algorithm}
\usepackage{array}
\usepackage[caption=false,font=normalsize,labelfont=sf,textfont=sf]{subfig}

\usepackage{tikz}
\usetikzlibrary{arrows.meta, positioning, shapes.geometric}
\usetikzlibrary{arrows.meta, positioning}

\renewcommand\footnotetextcopyrightpermission[1]{}

\usepackage{graphicx}

\usepackage[prologue,table,xcdraw,dvipsnames,svgnames,x11names]{xcolor}

\usepackage[edges]{forest}
\usepackage{longtable}

\usepackage{multido}
\newcommand{\blackbullets}[1]{\multido{\i=1+1}{7}{\ifnumgreater{\i}{#1}{\color{lightgray}}{}{\LARGE{\textbullet}}\kern 0.40em}}

\usepackage{eso-pic}

\usepackage{textcomp}
\usepackage{stfloats}
\usepackage{url}

\usepackage{tabularx}

\usepackage{verbatim}
\usepackage{graphicx}

\usepackage{adjustbox}

\usepackage{nicematrix}

\usepackage{flushend}

\usepackage{circledsteps}
\newcommand\myCircled[2][]{\ifmmode
\Circled[fill color=black,inner color=white,#1]{\mathsf{#1}}
\else
\Circled[fill color=black,inner color=white,#1]{\sffamily#1}
\fi
}

\usepackage{circledsteps}
\newcommand\myCircledSercan[2][]{\ifmmode
\Circled[fill color=black,inner color=white,#1]{\mathsf{#2}}
\else
\Circled[fill color=black,inner color=white,#1]{\sffamily#2}
\fi
}

\definecolor{navy}{RGB}{0,51,102}

\usepackage{comment}
\usepackage{amsmath}
\usepackage{mathtools}

\usepackage{ragged2e}

\usepackage{bbding}
\usepackage{pifont}

\usepackage[most]{tcolorbox}
\usepackage{tikz}
\usetikzlibrary{arrows.meta, positioning, shadows.blur, fit}

\newcommand{\red}{}

\newcommand{\blue}{}
\renewcommand{\red}{\textcolor{red}}

\renewcommand{\blue}{\textcolor{blue}}

\newcommand{\navy}{}
\renewcommand{\navy}{\textcolor{Navy}}
\definecolor{myGreenA}{RGB}{62,176,86}

\usepackage{float}

\usepackage{booktabs}

\usepackage{amsfonts}

\usepackage{graphbox} 

\usepackage{nccmath}

\usepackage{multirow}
\usepackage{colortbl}
\usepackage{hhline}

\usepackage{enumitem}

\usepackage{rotating}

\usepackage{pifont}

\definecolor{stOneColor}{HTML}{F5C542}   
\definecolor{stTwoColor}{HTML}{2FBF71}   
\definecolor{stThreeColor}{HTML}{A77AD8} 
\definecolor{stFourColor}{HTML}{FF8F95}  
\definecolor{stFiveColor}{HTML}{38BDE3}  
\definecolor{stSixColor}{HTML}{E6D900}   

\newcommand{\stone}{\textcolor{stOneColor}{\ding{108}}}    
\newcommand{\sttwo}{\textcolor{stTwoColor}{\ding{110}}}    
\newcommand{\stthree}{\textcolor{stThreeColor}{\ding{115}}}
\newcommand{\stfour}{\textcolor{stFourColor}{\ding{117}}}  
\newcommand{\stfive}{\textcolor{stFiveColor}{\ding{168}}}  
\newcommand{\stsix}{\textcolor{stSixColor}{\ding{88}}}     

\definecolor{a_best}{RGB}{102,255,102}
\definecolor{bitstream}{RGB}{255,73,51}
\definecolor{a_best_fluc}{RGB}{237,2,140}

\usepackage{makecell}

\usepackage[prependcaption, textsize=footnotesize]{todonotes}

\usepackage{tablefootnote}

\usepackage{algorithm}
\usepackage{algorithmic}

\usepackage{soul}

\usepackage[caption=false,font=normalsize,labelfont=sf,textfont=sf]{subfig}

\usepackage{fixltx2e}

\usepackage{stfloats}

\newcommand{\RNum}[1]{\uppercase\expandafter{\romannumeral #1\relax}}

\def\arrvline{\hfil\kern\arraycolsep\vline\vline\kern-\arraycolsep\hfilneg}

\usepackage{etoolbox}

\AtBeginDocument{%
  }

\setcopyright{acmcopyright}
\copyrightyear{2026}
\acmYear{2026}
\acmDOI{}

\acmJournal{CSUR}
\acmVolume{xx}
\acmNumber{xx}
\acmArticle{xxx}

\newcommand{\HV}{{$\boldsymbol{\mathcal{HV}}$}}

\makeatletter
\let\ACM@origbaselinestretch\baselinestretch
\makeatother

\begin{document}

\title{Learning from Hypervectors: A Survey on Hyperdimensional Encoding
}

\author{Sercan Aygun}
\email{sercan.aygun@louisiana.edu}
\orcid{0000-0002-4615-7914}
\affiliation{%
  \institution{University of Louisiana at Lafayette}
  \city{Lafayette}
  \state{LA}
  \country{USA}
  \postcode{70503}
}
\author{Mehran Moghadam}
\email{moghadam@case.edu}
\orcid{0000-0002-1325-1664}
\author{M. Hassan Najafi}
\email{najafi@case.edu}
\orcid{0000-0002-4655-6229}
\affiliation{%
  \institution{Case Western Reserve University}
  \city{Cleveland}
  \state{OH}
  \country{USA}
  \postcode{44106}
}

\author{Mohsen Imani}
\email{m.imani@uci.edu}
\orcid{0000-0002-5761-0622}
\affiliation{%
  \institution{University of California Irvine}
  \city{Irvine}
  \state{CA}
  \country{USA}
  \postcode{92697}}


\renewcommand{\shortauthors}{Aygun et al.}

\begin{abstract}
Hyperdimensional computing (HDC) 
is an emerging computing paradigm that imitates the brain's structure 
to offer a powerful and efficient processing and learning model. In HDC, the data are encoded with long vectors, called \textit{hypervectors}, 
which typically range from $1{,}000$ to $10{,}000$ in length.
The literature provides several encoding methods to generate orthogonal or correlated hypervectors, based on the application.
Existing surveys focus on overall HDC system aspects like inputs, primary computations, and outputs.
However, this study takes a more specific approach. It zeroes in on the HDC system input and the generation of hypervectors, directly influencing the hypervector encoding. This survey brings together various hypervector generation and encoding methods, exploring the limitations, challenges, and potential benefits they entail. 
\end{abstract}



\received{02 December 2023}
\received[revised]{21 June 2026}
\received[accepted]{13 July 2026}


\AddToShipoutPictureFG*{%
  \AtPageUpperLeft{%
    \put(0,-60){%
      \makebox[\paperwidth][c]{%
        \normalfont \Large
        \textcolor{Blue}{This work has been accepted for publication in ACM Computing Surveys (CSUR).}
      }%
    }%
  }%
}

\maketitle

\section{Introduction}
Hyperdimensional Computing (HDC), also known as Vector Symbolic Architecture (VSA), is an emerging, brain-inspired computing paradigm that has recently gained considerable attention.
It has been applied to a wide range of cognitive tasks~\cite{cognitiveHDC}, including text analysis~\cite{AbbasLang}, recommendation systems~\cite{resSysHDC}, genome research~\cite{DNAimani}, security~\cite{10044775}, and Internet of Things (IoT) applications~\cite{BehnamIoT}, among others. 
Much of this interest stems from 
its unconventional approach to data processing, which places it
alongside other emerging  paradigms such as stochastic computing (SC), unary computing, approximate computing, 
and quantum computing \cite{largeSurvey, 8122049, 7348659, leon2023approximate}. 
In HDC, scalar data are encoded into long hypervectors ($\boldsymbol{\mathcal{HV}}$s)
that map the input from the scalar domain into a 
hyperspace. 
Because $\boldsymbol{\mathcal{HV}}$s are processed through simple logic elements rather than complex circuitry, basic arithmetic operations, such as multiplication and addition for binary $\boldsymbol{\mathcal{HV}}$s, reduce to logical \texttt{XOR}s and counters.
One significant advantage of operating on long $\boldsymbol{\mathcal{HV}}$s is their robustness against soft 
errors \cite{ScaleHD}:
unlike conventional binary representations, HDC has no especially error-prone bits, such as the \textit{most}- and \textit{least}-significant bits or the sign bit.
This fault-robustness is particularly valuable in in-memory computing (IMC), resistive RAM, 
and \texttt{FinFET}-based architectures~\cite{rahimi3D, reramRahimi}, underscoring the potential of HDC as a reliable solution for advanced computing platforms. 
In short, HDC represents information as large, fixed-length vectors, encoding data points, symbols, and concepts within a high-dimensional vector space.


\vspace{-4pt}
\subsection{Representation and Operation Conventions at-a-Glance.} 
\vspace{-2pt}
In HDC, data units are 
represented as high-dimensional vectors whose components follow a chosen representation scheme. In the \textit{binary} setting, each component takes a value from $\{0,1\}$, corresponding to logic levels, whereas in the \textit{bipolar} setting, each component takes a value from $\{-1,+1\}$. The two domains are directly related: a common hardware-oriented interpretation maps logic-$1$ to $+1$ and logic-$0$ to $-1$, although the exact mapping and the associated operations depend on the chosen representation. $\boldsymbol{\mathcal{HV}}$s of dimensionality $D$ up to $10{,}000$ are commonly used to encode data~\cite{HV_Design, 9614302}. A larger $D$ improves representational capacity and reduces accidental similarity between unrelated $\boldsymbol{\mathcal{HV}}$s, but it also increases storage and computational cost; the dimensionality ($D$) and the encoding type therefore jointly govern the \textit{accuracy}--\textit{efficiency} tradeoff. 

Orthogonality is central to HDC because it allows distinct features or symbols to be represented with low mutual similarity, as in letters for text processing~\cite{AbbasLang}, pixel positions for cognitive image tasks~\cite{binaryImageHDC}, and temporal patterns for voice recognition~\cite{voiceHD}. For this reason, random or pseudo-random vectors are widely used as atomic primitives, since they are approximately orthogonal in high-dimensional spaces~\cite{Zou2022}. Depending on the application, however, controlled correlation among $\boldsymbol{\mathcal{HV}}$s may also be useful when related scalar values or semantically similar inputs should remain close in hyperspace.  Several methods have been proposed to generate such $\boldsymbol{\mathcal{HV}}$s~\cite{Kanerva2009}. A common approach begins with an initial seed vector and derives additional vectors through random bit-flip operations~\cite{HDCluster}, producing correlated or similar vectors for nearby scalar values and increasingly dissimilar vectors for distant ones. The specific vector generation strategy depends on the data type and application context. In language classification, for instance, near-orthogonal random vectors are used to represent symbols, such as letters. Logical shifts are then applied to letter $\boldsymbol{\mathcal{HV}}$s to encode \textit{$N$-gram} structure while preserving contextual relationships~\cite{AbbasLang}.

\begin{figure*}[!t]
  \centering
  \newsavebox{\leftpanelbox}
  \sbox{\leftpanelbox}{%
    \includegraphics[width=0.56\textwidth]{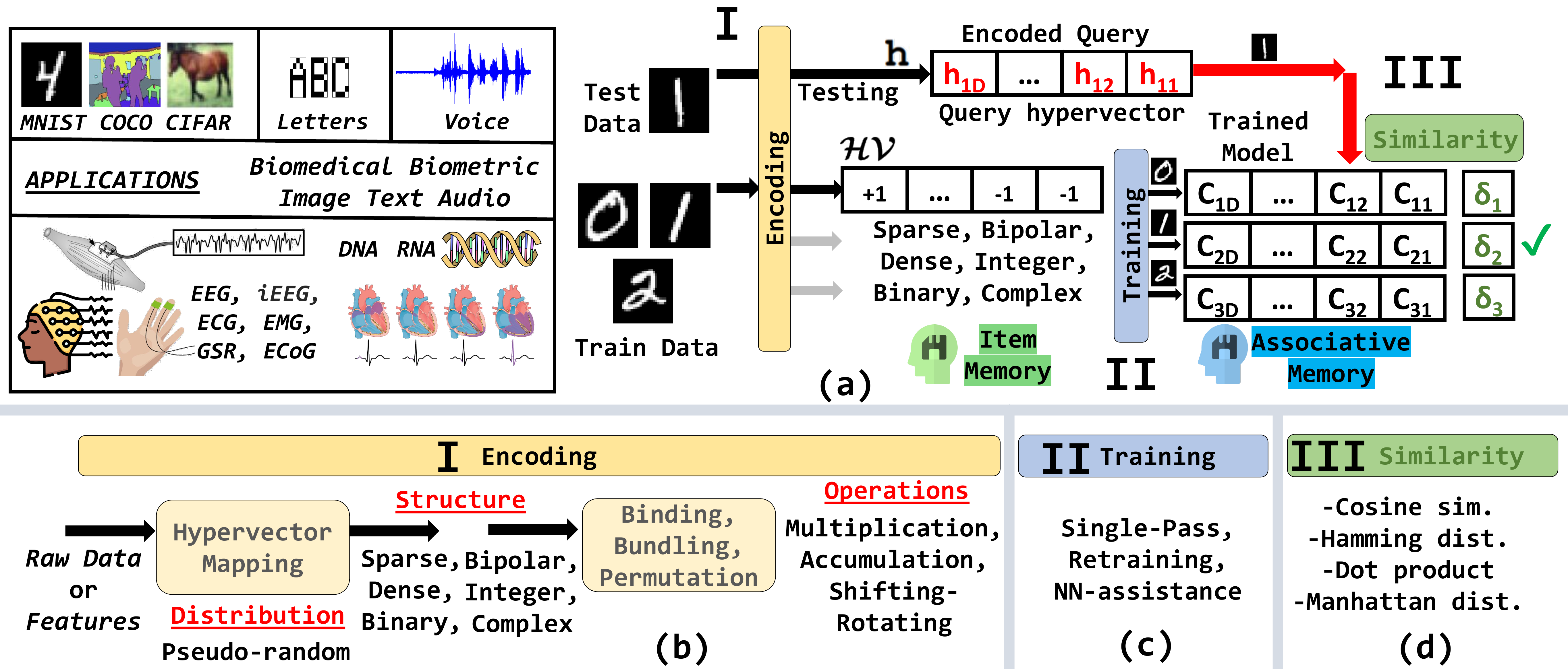}%
  }
  
  \begin{minipage}[b]{0.56\textwidth}
    \centering
    \usebox{\leftpanelbox}
  \end{minipage}%
  \hfill
  \begin{minipage}[b]{0.42\textwidth}
    \begin{tcolorbox}[
      height=\dimexpr\ht\leftpanelbox+\dp\leftpanelbox\relax, 
      valign=center, 
      colback=white,
      colframe=black!55,
      boxrule=0.47pt,
      arc=1mm,
      left=3pt,
      right=3pt,
      top=1.5pt,
      bottom=1.5pt,
      boxsep=1pt,
      title=\footnotesize\textbf{\textcolor{Navy}{HDC Notation Quick Reference}},
      coltitle=black,
      colbacktitle=gray!12
    ]
    \scriptsize
    \setlength{\parskip}{1.2pt}
    \linespread{0.92}\selectfont 

    \noindent \textbf{\navy{Binary:}} $\boldsymbol{\mathcal{HV}}\in\{0,1\}^{D}$ \quad \textbf{\navy{Bipolar:}} $\boldsymbol{\mathcal{HV}}\in\{-1,+1\}^{D}$

    \noindent\textbf{\navy{Domain Mapping:}}
    Binary and bipolar domains are related by $0\leftrightarrow -1$ and $1\leftrightarrow +1$.

    \noindent\textbf{\navy{Binding}} associates two $\boldsymbol{\mathcal{HV}}$s using $\oplus$ (\texttt{XOR}) in the binary domain or $\odot$ (Hadamard product) in the bipolar domain. 

    \noindent\textbf{\navy{Bundling}} accumulates (bipolar) / pop-counts (binary) $\boldsymbol{\mathcal{HV}}$s.
    \textbf{\textit{\navy{sign}}} maps accumulated bipolar values back to $\{-1,+1\}^{D}$; 
    
    \textbf{\textit{\navy{majority thresholding}}} maps binary counts back to $\{0,1\}^{D}$.

    \noindent\textbf{\navy{Permutation}} cyclic shifts or index permute to encode \textit{order}, \textit{position}, or \textit{sequence} information denoted as $\rho$.
    \end{tcolorbox}
  \end{minipage}
  \vspace{-0.75em}
  \caption{Overview of an HDC system for classification, together with a compact reference for the representation and operation conventions used throughout this survey. The main stages are: {\large{\texttt{\textbf{I}}}} encoding incoming data, {\large{\texttt{\textbf{II}}}} training with the input dataset, and {\large{\texttt{\textbf{III}}}} computing similarity between class-based $\boldsymbol{\mathcal{HV}}$s ($C_{\mathrm{class}}$) and the query \HV{}. The right-side quick reference summarizes \textit{binary}/\textit{bipolar} domains, binding operators, bundling and thresholding conventions, and permutation notation.}
  \label{general_structure}
\end{figure*}

The computational framework of HDC relies on a small set of representation-dependent operations: \textit{Binding} forms associations between $\boldsymbol{\mathcal{HV}}$s and is implemented as element-wise \texttt{XOR} in the binary domain and as element-wise multiplication, i.e., the Hadamard product, in the bipolar domain. \textit{Bundling} aggregates multiple $\boldsymbol{\mathcal{HV}}$s through component-wise accumulation in the bipolar domain or population counting of logic-$1$s in the binary domain. The resulting vector is then mapped back to the target domain using the sign function $\mathrm{sgn}(\cdot)$ for bipolar $\boldsymbol{\mathcal{HV}}$s or majority thresholding for binary $\boldsymbol{\mathcal{HV}}$s. In the binary domain, this thresholding can be implemented by comparing the population count at each dimension against a majority threshold. \textit{Permutation}, commonly implemented as cyclic shifting or index permutation, repositions components to encode order, position, or sequence information while preserving dimensionality. For example, in word or $N$-gram encoding, each letter is represented by its own letter $\boldsymbol{\mathcal{HV}}$, and shifted copies of these $\boldsymbol{\mathcal{HV}}$s distinguish the first, second, third, and subsequent positions in the sequence. These simple, robust, and parallelizable operations support holistic learning~\cite{Kanerva2009} and facilitate energy-efficient hardware implementations, motivating the use of HDC in emerging and synergistic computing paradigms. Fig.~\ref{general_structure} provides an at-a-glance summary of these representations and operation conventions together with the overall HDC workflow. Section~\ref{section_overview} then describes the workflow in detail and subsection~\ref{sec:formal_notation} formalizes the corresponding notation and definitions.

\subsection{Survey Scope and Literature Search Methodology}
\begin{figure*}[!t]
  \centering
  \includegraphics[width=455pt]{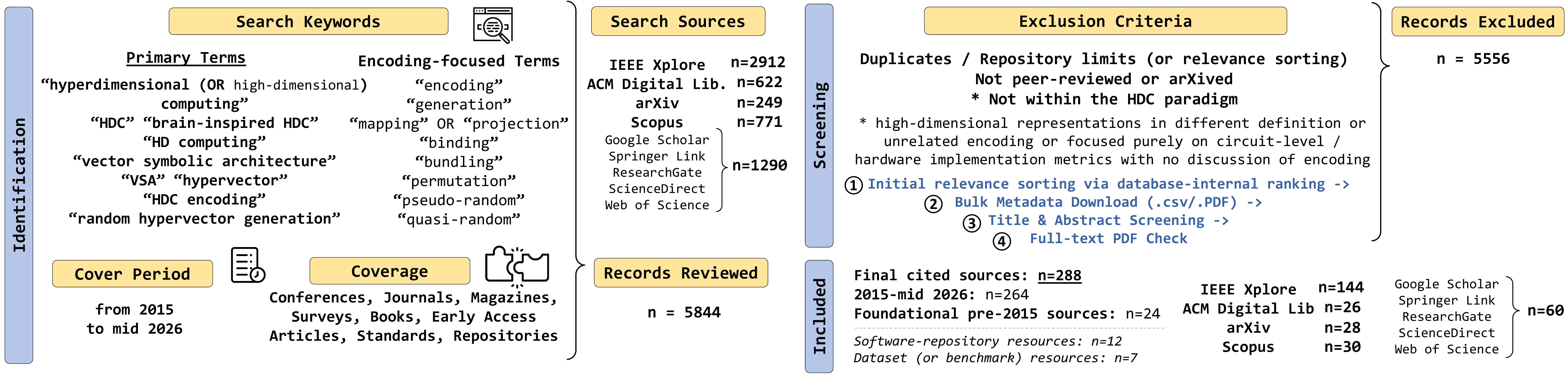}
  \vspace{-1em}
  \caption{PRISMA-inspired literature search, screening, and inclusion workflow used in this survey.}
 \label{taxonomy}
\end{figure*}

To make 
the scope and coverage of this survey transparent and reproducible, we describe the literature search, screening, and inclusion methodology used to identify the works discussed here. The primary focus of this survey is \textbf{\textit{\HV{} generation}} and \textbf{\textit{encoding strategies in HDC}}. Accordingly, we adopt a targeted, encoding-oriented perspective that complements existing system-level and application-level HDC/VSA surveys~\cite{10044775,KleykoPART1,KleykoPART2,10038612,9354795,ParhiSurvey,GRITSENKO,HDC_Survey_Thomas_JAIR2021,HDC_Survey_Stock_PLOS2024,HDC_Survey_Cumbo_PeerJ2025,HDC_Survey_Du_ACM_TECS2026,HDC_Survey_Xu_IEEE_TCOM2026,HDC_Survey_Salik_Elsevier2026,HDC_Survey_Salahudeen_Access2026}. Fig.~\ref{taxonomy} summarizes the search and screening workflow using a PRISMA-inspired structure.

\noindent\textbf{Search sources and terms.} The primary database searches covered \textit{IEEE Xplore}, \textit{ACM Digital Library}, \textit{arXiv}, and \textit{Scopus}. In addition, the complementary sources \textit{Google Scholar}, \textit{SpringerLink}, \textit{ResearchGate}, \textit{ScienceDirect}, \textit{Web of Science}, \textit{GitHub}, together with backward and forward citation tracing, were used for coverage validation, author-uploaded versions, repositories, datasets, software artifacts, and recently posted works. The primary search terms included ``\textit{hyperdimensional computing}'', ``\textit{HD computing}'', ``\textit{brain-inspired HDC}'', ``\textit{vector symbolic architecture}'', ``\textit{VSA}'', ``\textit{hypervector}'', ``\textit{HDC encoding}'', and ``\textit{random hypervector generation}''. These were combined with encoding-focused terms, including ``\textit{encoding}'', ``\textit{generation}'', ``\textit{mapping}'', ``\textit{projection}'', ``\textit{binding}'', ``\textit{bundling}'', ``\textit{permutation}'', ``\textit{pseudo-random}'', and ``\textit{quasi-random}''. Domain-specific terms, such as ``\textit{EEG classification}'', ``\textit{language processing}'', ``\textit{image HDC}'', ``\textit{DNA pattern matching}'', and ``\textit{biomedical sensing}'', were also used to verify coverage in application-oriented HDC studies. The reference lists of well-known surveys~\cite{10044775,KleykoPART1,KleykoPART2,10038612,9354795,ParhiSurvey,GRITSENKO,HDC_Survey_Thomas_JAIR2021,HDC_Survey_Stock_PLOS2024,HDC_Survey_Cumbo_PeerJ2025,HDC_Survey_Du_ACM_TECS2026,HDC_Survey_Xu_IEEE_TCOM2026,HDC_Survey_Salik_Elsevier2026,HDC_Survey_Salahudeen_Access2026} were further inspected to improve coverage of foundational and representative works.

\noindent\textbf{Time range and coverage.}
The main search window covered works from 2015 to mid-2026, and included conference papers, journal articles, early-access articles, books, surveys, theses, software repositories, and benchmark resources. Foundational works predating this window, including sparse distributed memory~\cite{kanerva1993sdm}, binary spatter coding~\cite{10.1007/3-540-61510-5_146}, holographic reduced representations~\cite{377968}, and other SOTA works, were retained through manual inspection because of their foundational influence on modern HDC encoding practices.

\noindent\textbf{Screening and exclusion procedure.}
The screening procedure followed four stages. First, candidate records were retrieved using the search terms above and sorted by the database-internal relevance ranking where available; this was particularly useful for IEEE Xplore, Scopus, ScienceDirect, and arXiv, where the relevance ranking helped prioritize records more closely connected to HDC and encoding. Second, the available metadata, including titles, abstracts, keywords, venue information, publication years, and DOI, were downloaded in CSV, BibTeX, text, or PDF formats. Third, titles and abstracts were screened against the inclusion and exclusion criteria. Finally, the remaining candidates were verified through full-text inspection. A work was included if it: (i) proposed, employed, or systematically compared one or more \HV{} generation, mapping, or encoding methods; (ii) appeared in a peer-reviewed venue, as a substantive preprint, or as a relevant software, dataset, repository resource; and (iii) was 
situated within the HDC (VSA) paradigm. Works were not carried forward if they were duplicates, fell outside the HDC paradigm, used high-dimensional representations in unrelated settings without explicit HDC framing, focused solely on circuit-level or hardware metrics without discussing encoding, or lacked sufficient relevance to \HV{} generation, representation, mapping, or composition. Works at the intersection of HDC and emerging paradigms, such as SC~\cite{StocHD, StochasticHD}, neuromorphic computing~\cite{NeuroHDC_TVLSI2026, 9892030, HyperSpikeASIC}, and quantum computing~\cite{10655760, poduval2024hyperdimensionalquantumfactorization}, were included when encoding was a defined contribution or an important part of the method.

\noindent\textbf{Search statistics and final selection.} The database searches produced $5{,}844$ raw candidate records: $2{,}912$ from IEEE Xplore, $622$ from ACM Digital Library, $249$ from arXiv, $771$ from Scopus, and $1{,}290$ from the complementary sources. These raw counts include cross-database overlap and should therefore be interpreted as database-level candidate records rather than unique papers. After relevance sorting, duplicate handling, title-abstract screening, and full-text inspection, $5{,}556$ raw records were not carried forward. The final bibliography contains 288 cited sources and resources: 264 from the 2015--mid-2026 search window and 24 foundational pre-2015 works. The references include 12 software-repository resources and 7 dataset or benchmark resources.

This work provides a concise and comprehensive review of the \textit{\textbf{encoding}} process in HDC systems. It is structured as follows: Section~\ref{section_overview} provides an overview of HDC systems and relevant background. Section~\ref{section_vector_generation} reviews prior surveys, introduces the proposed multi-stage encoding taxonomy, and discusses representative \HV{} mapping and composition strategies. Section~\ref{challenges} identifies and discusses critical challenges, limitations, and future opportunities related to encoding in HDC, drawing insights from diverse research efforts. Section~\ref{extras} further offers practical guidance to the reader with a list of available platforms, HDC evaluation metrics, and a summary of state-of-the-art works. Finally, Section~\ref{conclusions} presents the study's conclusions and findings. A glossary of key terms is provided in the Appendix to support clarity in terminology throughout the paper.

\vspace{-5pt}
\section{Overview of HDC Systems}
\label{section_overview}
This section provides the background needed to understand
HDC systems and
presents an overview of the key sub-modules that constitute the HDC pipeline.
Fig.~\ref{general_structure} 
shows a general depiction of the HDC system, consisting of three primary steps. 
The first step, shown in Fig.~\ref{general_structure}~(a)~{\large{\texttt{\textbf{I}}}}, is the encoding. 
After mapping the input data (or features) to the atomic $\boldsymbol{\mathcal{HV}}$s, the encoding stage performs 
some encoding operations on 
the $\boldsymbol{\mathcal{HV}}$s.
These operations 
include \textit{Binding}, \textit{Bundling}, and \textit{Permutation}~(Fig.~\ref{general_structure}~(b)). The second step, shown in 
Fig.~\ref{general_structure}~(a)~{\large{\texttt{\textbf{II}}}}, corresponds to the \textit{Training} phase. In this step, all of the related $\boldsymbol{\mathcal{HV}}$s from the same category are merged together to create a unique class $\boldsymbol{\mathcal{HV}}$ to represent any unique 
data class. Lastly, the third step, shown in Fig.~\ref{general_structure}~(a)~{\large{\texttt{\textbf{III}}}}, is the \textit{Similarity} calculation, which measures 
the similarity (e.g. \texttt{cosine}, \texttt{hamming distance}, \texttt{dot product}, etc.) between the newly encoded \HV{} 
and the class $\boldsymbol{\mathcal{HV}}$s. 
This survey focuses on the \textbf{\textit{encoding step}} in Fig.~\ref{general_structure} with detailed insights targeting the initial vector mapping and continuous encoding. 
The initial stage of encoding involves embedding data into vectors, called ``\textit{mapping}''. This process, known as $\boldsymbol{\mathcal{HV}}$ generation or projection, generates a vector based on a pseudo-random source. Then, multiple vectors are combined holistically for a learning process.

\begin{figure}[t]
\centering
\scriptsize
\begin{minipage}[t]{0.53\linewidth}
\vspace{2pt}
\centering
\includegraphics[width=\linewidth]{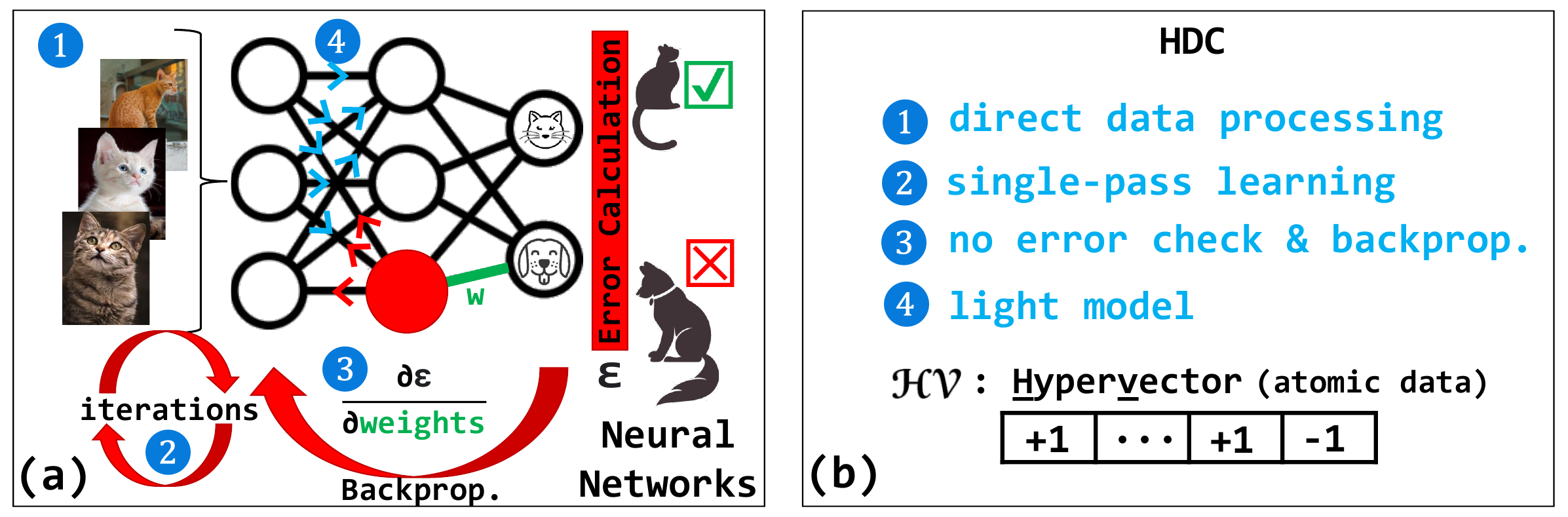}
\end{minipage}
\hfill
\begin{minipage}[t]{0.46\linewidth}
\vspace{0pt}
\centering
\textbf{Comparison: HDC and ML-based Representation Learning}
\vspace{0.3em}
\renewcommand{\arraystretch}{0.98}
\setlength{\tabcolsep}{2.2pt}
\begin{tabular}{lll}
\toprule
\textbf{Feature} & \textbf{HDC Encoding} & \textbf{ML Representation} \\
\midrule
\rowcolor{Lavender!30}
Vector Type & Binary/Bipolar & Real-valued dense \\
Learning & Non-iterative & Gradient-based \\
\rowcolor{Lavender!30}
Dimensionality & High ($\geq$1,000) & Low--mid (32--1024) \\
Training Data & Few-shot & Large datasets \\
\rowcolor{Lavender!30}
Hardware Cost & Low & High \\
Interpretability & Medium--High & Low \\
\rowcolor{Lavender!30}
Noise Robustness & High \& Inherent & Variable \& Costly \\
Domains & Edge, IoT, security & NLP, vision, general AI \\
\bottomrule
\end{tabular}
\end{minipage}
\vspace{-1.2em}
\caption{Comparison of \textit{learning}: HDC vs. NNs, together with a compact summary of HDC encoding and ML-based representation learning.
}
\label{fig:hdc_vs_nn}
\end{figure}

HDC offers 
an efficient alternative to traditional machine learning (ML) models through four key properties, as demonstrated in Fig.~\ref{fig:hdc_vs_nn}: streamlined arithmetic operations \cite{StocHD}, non-iterative learning \cite{onlinehd}, model tuning without optimization \cite{10.1145/3503541}, and compact model size \cite{HDC_Aygun_DATE2023}.
Conventional neural networks (NNs) rely on three fundamental operations during the learning phase: (i) multiplication, (ii) addition, and (iii) activation (i.e., applying a non-linear function such as ReLU or sigmoid to introduce complexity into the model). In contrast, HDC's model acquisition uses simpler logical operations.
Conventional ML with NNs requires optimization through computationally intensive steps. 
As noted in Fig.~\ref{general_structure}~(c), HDC uses single-pass learning~\cite{ma2022hyperdimensional}.
Training data are evaluated once in most approaches, and the holistic model is obtained. 
In some cases, HDC models incorporate NN assistance to improve learning
efficiency~\cite{10.1145/3400302.3415696}.
The 
HDC model consists of a 
set of $\boldsymbol{\mathcal{HV}}$s, each representing a class vector for the classification task. Unlike using positional weights in matrices connecting numerous neurons across consecutive layers in NN systems, HDC uses 
$D$-sized $\boldsymbol{\mathcal{HV}}$s for each class. 
When the training is complete, 
the obtained model is checked with the testing data as in conventional ML systems. 
Encoding in HDC is a comprehensive step applied to both training and testing data.
For any class, the query data is encoded into a $D$-sized $\textbf{{\texttt{h}}}$ypervector. 
During the decision phase, the query vector is compared with the $\boldsymbol{\mathcal{HV}}$ of each class
already trained in the model.
Fig.~\ref{general_structure}~(d) shows different similarity measures used for checking 
the similarity between query $\textbf{{\texttt{h}}}$ypervector and the class $\boldsymbol{\mathcal{HV}}$s. The simplicity and efficiency of the HDC paradigm, along with its single-pass training advantage, 
have spurred its adoption in various 
domains. Table~\ref{table_HDC_apps} presents an overview of prior HDC studies across diverse domains. Compared to NNs, HDC’s lightweight learning has made it a growing focus in these diverse research fields.

\begin{table}[!t]
\centering
\scriptsize
\setlength{\aboverulesep}{0pt} 
\setlength{\belowrulesep}{0pt} 
\setlength{\extrarowheight}{0pt}
\caption{First Glance of Prior HDC Works Classified based on Prominent Targets}
\vspace{-1.5em}
\resizebox{\linewidth}{!}{
\begin{tabular}{llll}
\toprule
\textbf{Target} & \textbf{SOTA Works} & \textbf{Target} & \textbf{SOTA Works}\\
\midrule

\rowcolor{Lavender!30}
\textbf{Image Processing/Computer Vision}
&
\cite{LocalityEN,cognitiveHDC,QuantHD,kang2022openhd,BRIC,symbolichdc,BinHD,10089842,Abbas_Class_recall,LeHDC,ma2022hyperdimensional,10.1145/3495243.3558757,10.1145/3489517.3530669,VQA,montone2017hyperdimensional,10.1007/978-3-030-89820-5_3,10.1145/3370748.3406560,Watkinson2021,duan2022braininspired,10024980,hdnn,gluing,PerfHD,neuro2023,10030789,SchubertDescriptors,menet2023mimonets,hersche2023probabilistic,uHD,Mehran-No-mul,HDFace,10.1145/3615991.3616404,9614302,yu2022understanding,ID-VSA_Moghadam_TVLSI2026,AddOn-IHD_Masum_DAC2025}
&
\textbf{Language/Text Processing}
&
\cite{rahimi3D,AbbasLang,okko,Abbas_Class_recall,RahimiRecentLanguage,9869459,VQA,montone2017hyperdimensional,10.1007/978-3-030-89820-5_3,10198153,10.1145/3477495.3531761,HDC_Aygun_DATE2023,ID-VSA_Moghadam_TVLSI2026}
\\

\textbf{Biomedical/Biometrics}
&
\cite{Imani_Hierarchical,10.1145/3400302.3415696,Abbas-ExG,Zou2022,sandwich,8771622,QuantHD,kang2022openhd,BRIC,SearcHD,BinHD,Abbas_Biosignal,abbas-ieeg,8584751,Menon2022,HV_Design,moin2018emg,Watkinson2021,AbbasEmbedding,10.1145/3503541,10050560,10.1007/s11036-017-0942-6,10.3389/fneur.2021.701791,10.3389/fneur.2022.816294,HyperSpec,HwAwHDC,SpecHD,9614302,ID-VSA_Moghadam_TVLSI2026,AMS-HD_TBME2026,AddOn-IHD_Masum_DAC2025}
&
\textbf{Cross-paradigm Computing}\textsuperscript{\dag}
&
\cite{StocHD,HDC-IM,QubitHD,StochasticHD,Mehran-No-mul,HDC_Aygun_DATE2023,orchard2023hyperdimensional,Zou2022_,9892030,HyperSpikeASIC,uHD,HDFace,PhotoHDC,ID-VSA_Moghadam_TVLSI2026,AMS-HD_TBME2026,NeuroHDC_TVLSI2026}
\\

\rowcolor{Lavender!30}
\textbf{Security/Reliability Issues}
&
\cite{9943057,10089842,10044775,cai2022hyperlock,10.1007/978-3-030-65745-1_22,9869459,gungor2022reshd,10.1145/3576914.3587484,cyberhd,Prive-HD,yang2020adversarial,WANG2022125042,HDTest,PoisonHD,chen2021adversarial,HD-I-IoT,drophd2024genssler,reasoning2023iccad,10137331,park2023hyperdimensional,HDC_ind,VLSI-SoC22,Nam2023EfficientML,lightningtalk,10038683,10247811,10720539,10908571,jeong2025exploitingboostinghyperdimensionalcomputing,10415375,Radiation_HDC_ACM_TECS2026}
&
\textbf{Graph Processing}
&
\cite{cognitiveHDC,10.5555/3539845.3540189,RelHD,9943057,9473227,HyperGRAF,9892300,10.5555/3016100.3016172,10.1145/3622797}
\\

\textbf{Memory-centric Computing}
&
\cite{10.1145/3524067,cognitiveHDC,StocHD,SearcHD,Karunaratne2020,MIMHD,reramRahimi,Imani_Dual,hdnn,ST-HDC,9642237,abbas-robust,Langenegger2023,10038683,9911329,9892243,Thomann2022,cai2022hyperlock,https://doi.org/10.1002/inf2.12416,10.1145/3503541,HDC-IM,10.3389/fnins.2022.867568,tutorial2023esweek,10198153,9656637,wang2023computinginmemorybased,AddOn-IHD_Masum_DAC2025,LogHD_DATE2026}
&
\textbf{DNA - Genome Application}
&
\cite{geniehd,DNAimani,a13090233,BioHD,9586253,9786145,10137331}
\\

\rowcolor{Lavender!30}
\textbf{IoT \& Edge AI}
&
\cite{HD-I-IoT,BehnamIoT,HDC_Arch,imani-efficient,10.1145/3576914.3587484,9943057,neuralhd,10.1145/3489517.3530669,HV_Design,9857180,Genssler2022BrainInspiredHC,lightningtalk,9831446,SenHDC_TVLSI2025,SenseHD_TCAD2026,FlexibleHD_SA2026,NeuroHDC_TVLSI2026,LogHD_DATE2026,DystoHD,DecoHD_DATE2026,Radiation_HDC_ACM_TECS2026}
&
\textbf{Robotics}
&
\cite{9811939,hassan2023efficient,sensorimotor,Neubert2019,10.1117/12.2585772,SchubertDescriptors}
\\

\bottomrule
\end{tabular}
}
\justify{\scriptsize{
\dag: Cross-paradigm Computing deals with exploiting Quantum Computing, Neuromorphic Computing, Unary Computing, and Stochastic Computing under the umbrella of the emerging computing paradigms in the context of HDC.
}}
\label{table_HDC_apps}
\vspace{-1.0em}
\end{table}

\subsection{HDC's Comparison to Modern Learning Phenomena: Representation Learning}
\label{sec:rep_learning_comparison}

Representation learning has become a foundational component of modern ML, particularly in deep learning, where the goal is to automatically learn dense and continuous feature representations from raw input data. Common approaches include autoencoders, word embeddings (e.g., \textit{word2vec}, \textit{GloVe}), and deep contextualized models such as \textit{BERT} and \textit{GPT} \cite{mikolov2013distributed, pennington2014glove, devlin2019bert}. These methods typically map inputs into low-dimensional latent spaces optimized for downstream performance. In contrast, HDC uses high-dimensional ($\geq$ 1,000 dimensions) vectors to represent data. These $\boldsymbol{\mathcal{HV}}$s are not learned through backpropagation but are either randomly generated, handcrafted, or constructed using data-driven heuristics. HDC representations are holistic, compositional, and noise-resilient.

There are several structural and functional differences between HDC encoding and traditional representation learning methods in ML~\cite{info15080490}; \ Fig.~\ref{fig:hdc_vs_nn} (right-side table) presents a point-by-point comparison. While ML embeddings typically operate in low to mid-dimensional spaces (ranging from 32 to 1024 dimensions), HDC employs high-dimensional vectors, often ranging from 1,000 to 10,000 dimensions. HDC representations are usually sparse and binary or bipolar in nature, whereas ML encodings are dense and real-valued. From a learning perspective, ML representations are optimized through iterative, gradient-based methods, while HDC relies on rule-based or randomized, non-iterative encoding. In terms of use cases, HDC excels in scenarios that demand lightweight and efficient processing, such as on-device learning~\cite{10.1145/3489517.3530669}, where low power consumption and rapid inference are critical \cite{AbbasLang}. It is also particularly suited for privacy-sensitive applications~\cite{9869459, Prive-HD} due to the symbolic nature of its encodings. Furthermore, HDC offers inherent robustness for hardware implementations, including in-memory computing and fault-tolerant architectures \cite{Karunaratne2020, 10038683}. In contrast, ML-based learned representations offer higher accuracy in large-scale, data-rich environments. They support contextual encoding, particularly in domains like natural language processing (NLP) and computer vision using transformer-based models, and provide tunable control through extensive training data and parameter optimization. Recent efforts have explored hybrid models that seek to combine the benefits of both paradigms. These include leveraging ML-derived dense embeddings as inputs to HDC systems, or applying HDC principles to operate on compressed latent features \cite{AyarNICEllm, austin_ESL}. Such integrations have the potential to bridge the efficiency of HDC with the representational richness of modern ML techniques. 

\subsection{Formal Definitions and Notations}
\label{sec:formal_notation}

To support technical rigor throughout this survey, we formally define the representational types and core operations of HDC in mathematical terms. Table~\ref{HV_types_ops} summarizes the common \HV{} representation types and their associated operations.

\vspace{-0.5em}
\subsubsection{Hypervector Types} 
Let $\mathbf{h}$ denote a $D$-dimensional \HV{}. In the binary domain, $\mathbf{h}\in\{0,1\}^{D}$, whereas in the bipolar domain, $\mathbf{h}\in\{-1,+1\}^{D}$.
The two representations are isomorphic under the element-wise mapping $0 \leftrightarrow -1$ \& $1 \leftrightarrow +1$, but they differ in their natural encoding semantics, as detailed below.

\vspace{-0.5em}
\subsubsection{Binding Operation} 
Binding is a compositional operation that combines multiple \HV{s} to produce a result that is approximately orthogonal to each operand, enabling reversible association. 
For \textit{Binary} \HV{s}, binding is 
element-wise \texttt{XOR}: $(\mathbf{h}_a\oplus \mathbf{h}_b)_i=h_{a,i}\oplus h_{b,i}$, where $\mathbf{h}_a, \mathbf{h}_b \in\{0,1\}^D$.
For \textit{Bipolar} \HV{s}, binding corresponds to element-wise (Hadamard) multiplication: $(\mathbf{h}_a\odot \mathbf{h}_b)_i=h_{a,i}\odot h_{b,i}$, where $\mathbf{h}_a, \mathbf{h}_b \in\{-1,+1\}^D$.
Both are self-inverse, that is: $\mathbf{h}_a=(\mathbf{h}_a\oplus \mathbf{h}_b)\oplus \mathbf{h}_b$ and $\mathbf{h}_a=(\mathbf{h}_a\odot \mathbf{h}_b)\odot \mathbf{h}_b$.
Throughout this survey, we use $\oplus$ exclusively for \texttt{XOR}-based binding and $\odot$ for multiplicative binding.

\vspace{-0.5em}
\subsubsection{Bundling and Sign Operations}
Bundling aggregates a set of \HV{s} $\{\mathbf{h}_1, \ldots, \mathbf{h}_n\}$ into a holistic summary vector via component-wise summation followed by binarization (sign).
For binary \HV{s}, threshold-based majority vote replaces the sign function in bipolar form; output is 1 if more than half the operands carry a 1 in that position, 0 otherwise. Unlike binding, bundling is generally not invertible; it serves as the primary mechanism for constructing class \HV{s} during training.

\vspace{-0.5em}
\subsubsection{Permutation Operation}
Permutation encodes positional or sequential information by shuffling vector elements. It is most commonly defined as a cyclic shift by $k$ positions, denoted as $\rho^k(\mathbf{H})$, where the components of the vector $\mathbf{H}$ are shifted cyclically to the right or left. The resultant \HV{} after $k$ cyclic shifts keeps near-orthogonality to the original \HV{}, which is the basis for position-sensitive encoding in $N$-gram processing (see Section~\ref{rb_ng_encoding}).

\begin{table}[!t] 
\centering  
\scriptsize 
\caption{Summary of \HV{} Types and Operations} 
\vspace{-1.2em} 
\setlength{\aboverulesep}{0pt} 
\setlength{\belowrulesep}{0pt} 
\setlength{\extrarowheight}{0pt} 
\renewcommand{\arraystretch}{1} 
\resizebox{\linewidth}{!}{ 
\begin{tabular}{@{} l l l l l l l @{}}
\toprule
\textbf{HV Type} & \textbf{Domain} & \textbf{Binding} & \textbf{Unbinding} & \textbf{Bundling} & \textbf{Binarization} & \textbf{Similarity} \\
\midrule

\rowcolor{Lavender!30}
Binary & $\{0, 1\}^D$ & \begin{tabular}[c]{@{}l@{}} \texttt{XOR}:\\ $(\mathbf{h}_a \oplus \mathbf{h}_b)_i$ \end{tabular} & \begin{tabular}[c]{@{}l@{}} \texttt{XOR}\\ again \end{tabular} & Majority vote / pop-count & \begin{tabular}[c]{@{}l@{}} Majority threshold \end{tabular} & \begin{tabular}[c]{@{}l@{}} Hamming Distance:\\ $\frac{1}{D} \|\mathbf{h}_a \oplus \mathbf{h}_b\|_1$ \end{tabular} \\

Bipolar & $\{-1, +1\}^D$ & \begin{tabular}[c]{@{}l@{}} Hadamard:\\ $(\mathbf{h}_a \odot \mathbf{h}_b)_i = h_{a,i}h_{b,i}$ \end{tabular} & \begin{tabular}[c]{@{}l@{}} Hadamard \\ again \end{tabular} & $\sum_k \mathbf{h}_k$ & $\text{sgn}(\cdot)$ & \begin{tabular}[c]{@{}l@{}} Cosine: \\ $\frac{1}{D} \sum_i h_{a,i}h_{b,i}$ \end{tabular} \\

\rowcolor{Lavender!30}
Real-valued & \begin{tabular}[c]{@{}l@{}} $\mathbb{R}^D$ \\ (unit-norm) \end{tabular} & \begin{tabular}[c]{@{}l@{}} Hadamard \end{tabular} & \begin{tabular}[c]{@{}l@{}} Approximate \\ (re-normalize) \end{tabular} & \begin{tabular}[c]{@{}l@{}} $\frac{1}{n} \sum_k \mathbf{h}_k$ \end{tabular} & \begin{tabular}[c]{@{}l@{}} $\text{sgn}(\cdot)$ or\\ quantize \end{tabular} & \begin{tabular}[c]{@{}l@{}} Cosine:\\  $\mathbf{h}_a \cdot \mathbf{h}_b / (\|\mathbf{h}_a\| \|\mathbf{h}_b\|)$ \end{tabular} \\

\begin{tabular}[c]{@{}l@{}} Complex \\ (phasor) \end{tabular} & \begin{tabular}[c]{@{}l@{}} $\mathbb{C}^D$ \\ $(|h_j| = 1)$ \end{tabular} & \begin{tabular}[c]{@{}l@{}} Phase add: \\ $(\mathbf{h}_a \odot \mathbf{h}_b)_j = e^{i(\theta_{a,j}+\theta_{b,j})}$ \end{tabular} & \begin{tabular}[c]{@{}l@{}} Conj. mult.: \\ $\mathbf{h}_c \odot \overline{\mathbf{h}_b}$ \end{tabular} & \begin{tabular}[c]{@{}l@{}} Superposition$\rightarrow$renormalize: \\ $\frac{\sum_k h_{k,j}}{|\sum_k h_{k,j}|}$ \end{tabular} & \begin{tabular}[c]{@{}l@{}} Normalize to \\ unit circle \end{tabular} & \begin{tabular}[c]{@{}l@{}} Inner product: \\ $\frac{1}{D} \text{Re}(\mathbf{h}_a \cdot \overline{\mathbf{h}_b})$ \end{tabular} \\
\bottomrule
\end{tabular}
}
\label{HV_types_ops}
\end{table}

\vspace{-0.5em}
\subsubsection{Other Hypervector Types.}
Beyond binary and bipolar \HV{s}, some HDC methods use real-valued or complex-valued representations to support continuous projections, phasor-based operations, or NN-assisted encoding.
\newline
\noindent {\textbf{\textit{Real-Valued Hypervectors}}.} A real-valued \HV{} is a vector $\mathbf{h} \in \mathbb{R}^D$, typically normalized to unit length: $\|\mathbf{h}\| = 1$. Components are most commonly drawn from a standard normal distribution $\sim \mathcal{N}(0,1)$. Real-valued \HV{s} arise naturally in Gaussian-based encoding frameworks such as \texttt{FebHD}~\cite{febhd}, \texttt{NeuralHD}~\cite{neuralhd}, \texttt{DistHD}~\cite{disthd}, and \texttt{ManiHD}~\cite{manihd}, where each dimension $h_i = \cos(\mathbf{B}_i \cdot \mathbf{F})$ is a nonlinear projection of an input feature vector $\mathbf{F}$ through a Gaussian random vector $\mathbf{B}_i$.
Binding for real-valued \HV{s} can be implemented using element-wise multiplication.
Bundling is performed by component-wise averaging: $\mathbf{h}_{\mathrm{bundle}} = \frac{1}{n}\sum_{k=1}^n \mathbf{h}_k$. Similarity is measured by cosine similarity as: $\mathrm{sim}(\mathbf{h}_a, \mathbf{h}_b) = \mathbf{h}_a \cdot \mathbf{h}_b / (\|\mathbf{h}_a\| \|\mathbf{h}_b\|)$. Real-valued \HV{s} are compatible with accelerated GPU-based matrix operations, but sacrifice the bitwise hardware efficiency of binary and bipolar representations.

\noindent {\textbf{\textit{Complex (Phasor) Hypervectors}}.}
A complex or phasor \HV{} is a vector $\mathbf{h} \in \mathbb{C}^D$ in which each component lies on the unit circle: $h_j = e^{\mathrm{i}\theta_j}$, with phase $\theta_j \in [-\pi, \pi]$. This representation is used in VSAs based on Holographic Reduced Representations (HRR) like~\cite{377968} and in the reinforcement learning based HDC frameworks such as~\cite{QHD}.
Binding for phasor \HV{s} is element-wise complex multiplication, which corresponds to phase addition: $(\mathbf{h}_a \odot \mathbf{h}_b)_j = e^{\mathrm{i}(\theta_{a,j} + \theta_{b,j})}$. 
Unbinding is achieved via conjugate multiplication: $\mathbf{h}_a = \mathbf{h}_c \odot \overline{\mathbf{h}}_b$ (where $\overline{\mathbf{h}}_b$ denotes component-wise complex conjugate), since $e^{\mathrm{i}\theta} \cdot e^{-\mathrm{i}\theta} = 1$. 
Bundling is performed via component-wise superposition with renormalization to the unit circle: $h_{\mathrm{bundle},j} = ({\sum_k h_{k,j}})\, /\, |\sum_k h_{k,j}|$. 
Similarity is the real part of the normalized inner product: $\mathrm{sim}(\mathbf{h}_a, \mathbf{h}_b) = \frac{1}{D}\,\mathrm{Re}(\mathbf{h}_a \cdot \overline{\mathbf{h}}_b)$. 
Complex \HV{s} naturally encode rotational or periodic structure and support circular convolution binding via fast Fourier transforms, but require hardware platforms that support complex arithmetic.

\section{Prior Surveys, Encoding Taxonomy, and Hypervector Mapping}
\label{section_vector_generation}

Before introducing the proposed encoding taxonomy, we position this survey with respect to the foundations of HDC and the prior survey literature. HDC traces its conceptual foundations back to the seminal contributions of Pentti Kanerva, who introduced the principles of sparse distributed memory (SDM) as an influential model for distributed information storage and retrieval using high-dimensional vectors~\cite{kanerva1993sdm}. Kanerva further developed these concepts through binary spatter coding, and Plate introduced HRR~\cite{377968}; all of these providing practical frameworks for encoding structured data into high-dimensional distributed representations~\cite{10.1007/3-540-61510-5_146}. His extensive studies paved the way for the effectiveness of hyperdimensional vector representations for robust and noise-tolerant computing~\cite{Kanerva2009, KanervaDesignTest}. Additionally, Kanerva illustrated how HDC naturally maps semantic relationships and cognitive concepts within vector spaces~\cite{Kanerva2010WhatWM}, highlighting its diverse application domains.

From a data science perspective, vector initialization plays a crucial role in shaping the overall HDC model.
In a SOTA survey on HDC, 
Kleyko et al.~\cite{KleykoPART1} acknowledge the existing research \cite{Plate1997ACF, Kanerva2009, Schlegel2022, abbas-hdc-nano, Neubert2019} and underscore the lack of a survey work dedicated to $\boldsymbol{\mathcal{HV}}$ for various types of data representations, such as numeric, 2D image, sequence, and others. 
Vector generation techniques tailored to specific data types have been relatively overlooked in favor of other aspects of HDC.
Undoubtedly, the encoding using $\boldsymbol{\mathcal{HV}}$s constitutes a critical step that profoundly impacts both the accuracy and efficiency of HDC. 
Dynamic $\boldsymbol{\mathcal{HV}}$ generation using efficient hardware is critical for reducing hardware costs and enabling on-edge learning~\cite{10.1145/3489517.3530669}.
The overall effectiveness and learning performance of the HDC models 
heavily rely on the quality of encoding the input data into $\boldsymbol{\mathcal{HV}}$s~\cite{HV_Design}. 
Accurate and efficient encoding is crucial for reliable, high-performance HDC operations across applications~\cite{10038612}.
Kleyko et al. delve into the structure of the 
$\boldsymbol{\mathcal{HV}}$s, 
such as dense, sparse, and binary. 
They discuss the trade-offs involved in the selection of $\boldsymbol{\mathcal{HV}}$ density and some $\boldsymbol{\mathcal{HV}}$ \textit{mapping} (or \textit{projection} alternatively) characteristics~\cite{Abbas_Class_recall}. Such insights shed light on optimizing the encoding and its impact on HDC performance and applications.

\begin{table}[!t]
\centering
\caption{Comparison of Previous SOTA Surveys for Key Contributions of This Survey}
\vspace{-1em}
\setlength{\aboverulesep}{0pt} 
\setlength{\belowrulesep}{0pt} 
\setlength{\extrarowheight}{0pt}
\resizebox{\linewidth}{!}{
\begin{tabular}{lcccccc} 
\toprule
\textbf{HDC Surveys} & \textbf{Year} & \begin{tabular}[c]{@{}c@{}}\textbf{Vector} \\ \textbf{Mapping}\end{tabular} & \begin{tabular}[c]{@{}c@{}}\textbf{Random Source}\\\textbf{Variety Discussion}\end{tabular} & \begin{tabular}[c]{@{}c@{}}\textbf{Open-source}\\\textbf{Platform Discussion}\end{tabular} & \begin{tabular}[c]{@{}c@{}}\textbf{Cross-paradigm}\\\textbf{Computing}\end{tabular} & \begin{tabular}[c]{@{}c@{}}\textbf{Application} \\ \textbf{Classifications}\end{tabular} \\ 
\midrule

Gritsenko et al.~\cite{GRITSENKO} & 2017 & \blackbullets{1} & \blackbullets{1} & \blackbullets{1} & \blackbullets{1} & \blackbullets{1} \\

Lulu and Parhi~\cite{ParhiSurvey} & 2020 & \blackbullets{3} & \blackbullets{1} & \blackbullets{1} & \blackbullets{2} & \blackbullets{2} \\

Thomas et al.~\cite{HDC_Survey_Thomas_JAIR2021} & 2021 & \blackbullets{7} & \blackbullets{3} & \blackbullets{1} & \blackbullets{5} & \blackbullets{2} \\

Kleyko et al. (part 1)~\cite{KleykoPART1} & 2022 & \blackbullets{6} & \blackbullets{6} & \blackbullets{1} & \blackbullets{2} & \blackbullets{7} \\

Kleyko et al. (part 2)~\cite{KleykoPART2} & 2022 & \blackbullets{7} & \blackbullets{5} & \blackbullets{2} & \blackbullets{3} & \blackbullets{7} \\ 
 
Hassan et al.~\cite{9354795} & 2022 & \blackbullets{3} & \blackbullets{2} & \blackbullets{5} & \blackbullets{1} & \blackbullets{3} \\ 
Ma et al.~\cite{10044775} & 2023 & \blackbullets{1} & \blackbullets{2} & \blackbullets{1} & \blackbullets{2} & \blackbullets{4} \\ 

Chang et al.~\cite{10038612} & 2023 & \blackbullets{1} & \blackbullets{1} & \blackbullets{1} & \blackbullets{3} & \blackbullets{6} \\ 

Stock et al.~\cite{HDC_Survey_Stock_PLOS2024} & 2024 & \blackbullets{5} & \blackbullets{2} & \blackbullets{3} & \blackbullets{4} & \blackbullets{6} \\

Cumbo and Chicco~\cite{HDC_Survey_Cumbo_PeerJ2025} & 2025 & \blackbullets{3} & \blackbullets{2} & \blackbullets{7} & \blackbullets{2} & \blackbullets{7} \\

Du et al.~\cite{HDC_Survey_Du_ACM_TECS2026} & 2026 & \blackbullets{6} & \blackbullets{2} & \blackbullets{1} & \blackbullets{7} & \blackbullets{6} \\

Xu et al.~\cite{HDC_Survey_Xu_IEEE_TCOM2026} & 2026 & \blackbullets{6} & \blackbullets{5} & \blackbullets{2} & \blackbullets{7} & \blackbullets{4} \\

Salik et al.~\cite{HDC_Survey_Salik_Elsevier2026} & 2026 & \blackbullets{7} & \blackbullets{2} & \blackbullets{1} & \blackbullets{6} & \blackbullets{7} \\

Salahudeen et al.~\cite{HDC_Survey_Salahudeen_Access2026} & 2026 & \blackbullets{7} &  \blackbullets{4} & \blackbullets{1} & \blackbullets{6} & \blackbullets{3} \\
 
\rowcolor{Lavender!30}
\textcolor{Navy}{\textbf{This work}} & \textcolor{Navy}{2026} & \textcolor{Navy}{\blackbullets{7}} &  \textcolor{Navy}{\blackbullets{7}} &  \textcolor{Navy}{\blackbullets{5}} & \textcolor{Navy}{\blackbullets{6}} & \textcolor{Navy}{\blackbullets{7}} \\ 
\bottomrule
\end{tabular}
}
\label{status_bar}
\end{table}

Our survey addresses a
gap in the literature, focusing specifically on $\boldsymbol{\mathcal{HV}}$  generation and encoding in HDC systems.
Previous surveys have presented the overall structure and motivations of HDC (see Table~\ref{status_bar}). However, 
this work goes into the 
$\boldsymbol{\mathcal{HV}}$ generation process. 
Chang et al.~\cite{10038612} review
HDC from the perspective of edge intelligence and cloud collaboration, including discussions on security and privacy. Within HDC research, they 
categorize the hardware environment into three primary domains of microprocessor, field-programmable gate arrays (FPGA), and computing-in-memory. 
While hardware comparisons of HDC implementations include some details on
encoding, a stronger focus on $\boldsymbol{\mathcal{HV}}$ generation is necessary.
A recent two-part survey by Kleyko et al. \cite{KleykoPART1, KleykoPART2} 
provides a comprehensive analysis on HDC.
The first part of their survey~\cite{KleykoPART1} focuses on exploring the opportunities for data transformation during $\boldsymbol{\mathcal{HV}}$ encoding. The primary aim of this part 
is to equip the readers with a foundational knowledge of basic notations, computational models, and input transformations. 
They also provide an overview of HDC models, emphasizing essential operations like binding, unbinding, and
superposition.
Lulu and Parhi~\cite{ParhiSurvey}, in their SOTA survey, define two main benchmark metrics for HDC: 
\textit{accuracy} and  \textit{efficiency}.
The \textit{accuracy} depends on three factors: \textit{encoding}, \textit{retraining}, and \textit{non-binary}. They discuss \textit{encoding} as 
one of the critical factors affecting accuracy.
\textit{Efficiency}, on the other hand, is discussed 
under two categories: 
algorithm and hardware. Three factors affect the algorithm: 
binarization, quantization, and sparsity.
\textit{$\boldsymbol{\mathcal{HV}}$} generation, the first step in encoding, affects both accuracy and
efficiency, depending on whether the vectors are sparse or dense.
In addition to these surveys and benchmarks, Kleyko~et~al.~\cite{KleykoPART1} highlight the studies in 
\cite{Kanerva2009}, \cite{KanervaDesignTest}, and \cite{Neubert2019} as tutorial-like introductory articles on HDC. 

More recently, the literature has also seen a rapid growth in both broad and highly specialized surveys exploring HDC from various interdisciplinary angles. Du et al.~\cite{HDC_Survey_Du_ACM_TECS2026}, for example, provide a comprehensive cross-layer perspective on VSAs, centered on the co-design of cognitive algorithms and brain-inspired hardware acceleration. 
While this survey strongly emphasizes cross-paradigm computing (specifically, bridging HDC with neuromorphic spiking arrays, IMC, and SC) and thoroughly categorizes applications, it offers little detail on the variety of vector generation sources and open-source platforms. 
Salik et al.~\cite{HDC_Survey_Salik_Elsevier2026} likewise deliver an extensive overview of HDC architectures for general AI applications. 
This survey presents a rigorous classification of applications by learning paradigms (supervised, unsupervised, semi-supervised) and provides detailed mathematical formulations of vector mapping strategies, including fractional power encoding and Siamese encoders. As with Du et al., it explores cross-paradigm fusions with vision-language models and quantum computing but lacks dedicated coverage of open-source HDC frameworks.

Other recent surveys have focused on integrating HDC with large-scale data frameworks and classical ML models. Xu et al.~\cite{HDC_Survey_Xu_IEEE_TCOM2026} explore a novel intersection between HDC and modern communication networks, proposing an Integrated Learning and Communication (ILAC) framework. Their survey excels in detailing how HDC integrates with large AI models, federated learning, and 6G wireless networks. Furthermore, it explicitly contrasts standard pseudo-random sampling with structured mathematical approaches such as low-discrepancy (LD) sequences. 
Salahudeen et al.~\cite{HDC_Survey_Salahudeen_Access2026} provide a highly specialized review devoted exclusively to HDC for regression tasks, offering an in-depth theoretical dive into feature mapping strategies, such as learnable Random Fourier Feature (RFF) encoders, and bridging HDC with classical machine learning paradigms (e.g., Kernel Ridge Regression and Random Forests). 
Thomas et al.~\cite{HDC_Survey_Thomas_JAIR2021} adopt a comparably rigorous theoretical perspective, detailing the mathematical foundations of vector mapping, continuous vector spaces, and distance preservation.

Finally, the flexibility of HDC has led to several domain-specific reviews, particularly within the biological sciences. Cumbo and Chicco~\cite{HDC_Survey_Cumbo_PeerJ2025} and Stock et al.~\cite{HDC_Survey_Stock_PLOS2024} both present comprehensive reviews focusing on bioinformatics and biomedicine. Cumbo and Chicco conduct a meta-analysis of HDC in medical informatics, notably advocating for standardized open-source practices by explicitly reviewing platforms such as \texttt{hdlib} and \texttt{Torchhd}. Stock et al. similarly develop a robust taxonomy of HDC in genomics and drug discovery, detailing specialized vector mapping techniques for biological data, such as DNA sequence encoding and k-mers. Because of their domain-specific focus, these reviews offer excellent application taxonomies but generally assume standard identically distributed initialization, without hardware-level random source varieties.
\color{black}

Compared to these SOTA works, this survey explores vector generation and mapping based on different random sources.
Additionally, it explores open-source platforms and applications categorized under diverse domains, including biomedical/biometrics, image processing and computer vision, security/reliability-related efforts, memory-centric computing, IoT \& edge, graph processing, emerging computing paradigms, symbol-based processing (primarily language and text processing), genome processing such as DNA, and the robotics in the HDC literature as of mid 2026. Focusing particularly on cross-paradigm emerging methods, Kleyko et al., in their Part II survey \cite{KleykoPART2}, provide an in-depth discussion of cognitive computing and its architectures in conjunction with HDC systems. They specifically elaborate on integrating 
spiking neural networks and HDC. That survey stands as a SOTA examination within its genre, providing a comprehensive overview of the HDC architecture. However, there is a notable absence of information regarding the synergy between other emerging paradigms: SC \cite{StocHD, StochasticHD}, approximate computing \cite{leon2023approximate}, in-(near-)memory computing~\cite{hdnn, tutorial2023esweek}, and HDC. 
Our survey examines previous efforts in \textit{bit-stream processing}, like SC, that align with HDC vector mapping and encoding.
This aspect distinguishes our work by offering 
new perspectives on bit-stream processing with similar features  
to holistic hyperdimensional vector processing. 
Our work 
supplements the recent literature by presenting some new viewpoints. Regarding applications, 
our survey classifies and discusses various applied learning
approaches, as shown
in~\cite{10044775, 10038612, KleykoPART1, KleykoPART2}. Notably, \cite{10044775} and \cite{10038612} are relatively focused surveys concentrating on security and edge computing applications, respectively. 
Most previous works have limitations in vector mapping, random source discussions, open-source platforms, and cross-paradigm computing opportunities. 
In what follows, we begin by discussing the taxonomy of encoding pipelines, followed by the foundational data processing methods in HDC.

\subsection{Multi-Stage Encoding Pipeline Taxonomy}
\begin{figure*}[t]
  \centering
  \includegraphics[width=\linewidth]{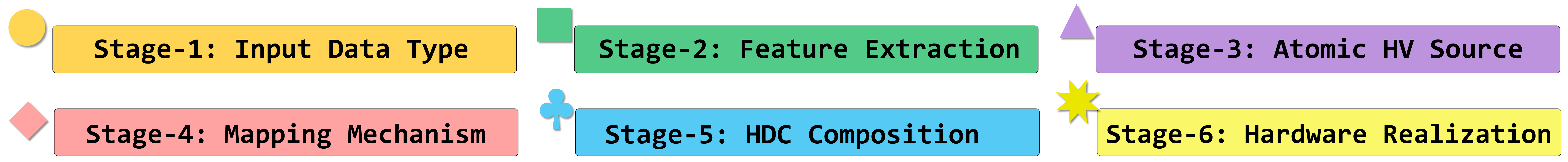}
  \vspace{-1.em}
  \caption{An overview of the comprehensive multi-stage encoding pipeline taxonomy.}
%
%
%
 \label{fig:pipeline_taxonomy}
\end{figure*}

We 
establish a principled multi-stage taxonomy that captures the full HDC encoding pipeline from raw input to hardware realization. This pipeline, illustrated in Fig.~\ref{fig:pipeline_taxonomy}, separates conceptually distinct 
decisions that are frequently combined in the literature. These stages should be interpreted as methodological decision points in the encoding workflow rather than as a 
mandatory sequence; depending on the application, some stages may be merged, skipped, or implemented implicitly.



\noindent \textbf{\textit{\stone~Stage-1: Input Data Type.}} The first stage determines the structural properties the encoding must preserve, which subsequently constrains the choices in the downstream stages. Four primary data categories arise in the HDC literature: \textit{Image pixels or Video frames} (spatial arrays requiring joint position-and-intensity encoding~\cite{cognitiveHDC,binaryImageHDC,LocalityEN}); \textit{Text or Symbols} (categorical tokens requiring orthogonal base \HV{s} and often permutation-based sequence encoding~\cite{AbbasLang,RahimiRecentLanguage}); \textit{Biosignals or Real-Time Signals} (continuous or streaming temporal data requiring position-sensitive or sliding-window encoding~\cite{Imani_Hierarchical,Abbas_Biosignal,abbas-ieeg,8584751}); and \textit{Graphs, DNA, or Genomes} (structured data requiring compositional binding and relational representations~\cite{HyperGRAF,RelHD,DNAimani,geniehd,BioHD}). Multi-modal inputs combine two or more of these types and are handled by cross-modal binding or feature-fusion strategies~\cite{VQA,gluing}.

\noindent \textbf{\textit{\sttwo~Stage-2: Feature Extraction.}}
The second stage captures the preprocessing steps that transform raw inputs into a form compliant with \HV{} mapping. Four core strategies appear in the pipeline: \textit{Discretization or Quantization} (binning continuous scalar values into discrete levels or compact feature codes~\cite{QuantHD,HV_Design,10546832}); \textit{Local Binary Patterns (LBP)} (extracting local texture or signal transitions~\cite{kaya20141d}); \textit{Deep Hashing and NN Outputs} (leveraging trained NNs to output dense or binary feature vectors~\cite{symbolichdc,gluing,neuralhd}); and \textit{Frequency Transforms (DFT)} (extracting spectral features, such as discrete Fourier coefficients, from continuous signals prior to hyperdimensional projection~\cite{voiceHD,9892462}).

\noindent \textbf{\textit{\stthree~Stage-3: Atomic HV Source.}}
The third stage describes the foundational statistical or algorithmic engine used to initialize the hyperdimensional space. This stage defines the mathematical nature of the generated base vectors: \textit{Pseudo-Random Sequences} (the standard stochastic baseline for generating approximately orthogonal binary or bipolar \HV{s}~\cite{Kanerva2009,HDCluster,HDC_Arch}); \textit{Low-Discrepancy (LD) Sequences} (deterministic, quasi-random vectors that can improve reproducibility or reduce initialization variability in some settings~\cite{uHD,Mehran-No-mul,Sobol_Aygun_TCAD,ID-VSA_Moghadam_TVLSI2026}); \textit{Gaussian Distributions} (continuous, real-valued initializations used in projection-based or neural-assisted HDC encoders~\cite{febhd,neuralhd,disthd,manihd}); and \textit{Thermometer or Cellular Automata} (structured, deterministic generation used for creating correlated level \HV{s}, compact scalar encodings, or dynamically evolved states~\cite{HV_Design,StocHD,StochasticHD}).

\noindent \textbf{\textit{\stfour~Stage-4: Mapping Mechanism.}}
The fourth stage dictates how the extracted features or scalars from Stage-2~\sttwo{} are embedded into the hyperdimensional space using the atomic \HV{s} from Stage-3~\stthree. Mechanisms include: \textit{Random Projection (Sparse or Dense)} (linear embedding via a random or sparse projection matrix~\cite{RandomProj,AbbasEmbedding,BRIC,LocalityEN}); \textit{Radial Basis Function (RBF), Cosine, or Random Exponential} (kernel-inspired and complex phasor-based embeddings, e.g., $h_i = \mathrm{sgn}(\cos(\mathbf{B}_i \cdot \mathbf{F}))$~\cite{febhd,neuralhd,RBF}); \textit{Fractional Power or Sigmoidal} (coordinate mapping and random nonlinear activations~\cite{KleykoPART1}); \textit{Selective Kanerva or Scatter Codes} (memory-inspired indexing and landmark activation~\cite{travnik2017representing,okko}); \textit{Cellular-Automata or Structured-State Mapping} (state-evolution-based transformations for structured or deterministic embeddings~\cite{CA-HV,Menon2022}); \textit{Stochastic or Bit-stream Mapping} (encoding through unary or stochastic bit-stream representations~\cite{StocHD,StochasticHD}); and \textit{End-to-End Learned Projections} (leveraging trained networks to jointly optimize the mapping space~\cite{symbolichdc,gluing,10.1145/3400302.3415696}).


\noindent \textbf{\textit{\stfive~Stage-5: HDC Composition.}}
The fifth stage captures how multiple mapped \HV{s} are mathematically combined into holistic, encoded representations. This encompasses core algebraic primitives shared across nearly all HDC systems--\textit{Binding} (\texttt{XOR} as $\oplus \mathbf{H}$ or Hadamard as $\odot \mathbf{H}$), \textit{Bundling} (majority vote or accumulation as $\Sigma \mathbf{H}$), and \textit{Permutation} (cyclic shifts as $\rho^k\mathbf{H}$)--as well as advanced structural composition and signal processing transforms~\cite{Kanerva2009,KleykoPART1,KleykoPART2}. Examples include the \textit{Kronecker or Tensor Product} for multi-modal feature combinations and \textit{Circular Convolution} for HRR/phasor-style compositional representations~\cite{377968,9892462,Abbas_Class_recall}.

\noindent \textbf{\textit{\stsix~Stage-6: Hardware Realization.}}
Finally, the sixth stage records the physical implementation substrate on which the encoding pipeline is executed. The hardware ecosystem spans \textit{FPGAs or Dedicated ASICs} (which typically favor pseudo-random binary \HV{s} and \texttt{XOR} binding~\cite{HDC_Arch}); \textit{In-Memory or In-Storage Computing} (leveraging RRAM or SRAM arrays for efficient matrix-vector operations~\cite{HDC-IM,Karunaratne2020}); \textit{In-Sensor or Near-Sensor Processing} (for extreme-edge data encoding)~\cite{SenHDC_TVLSI2025,SenseHD_TCAD2026}; \textit{emerging non-volatile memories (NVMs)} such as RRAM and FeFET~\cite{Imani_Dual,9656637}; and forward-looking \textit{Neuromorphic or Quantum Substrates}~\cite{9892030,10655760}. The choice of hardware realization strongly influences which Stage-3~\stthree{} sources and Stage-5~\stfive{} operations are the most energy- and area-efficient.

\color{black}

\subsection{Encoding with Data Type Considerations~\stone}
This section categorizes encoding strategies in HDC based on the type of input data.

\subsubsection{Image-Based Encoding}
Image data, whether grayscale or binary, is mapped to $\boldsymbol{\mathcal{HV}}$s through position-based or value-aware schemes. Binary images are often encoded using position $\boldsymbol{\mathcal{HV}}$s, where each pixel location corresponds to a pre-generated $\boldsymbol{\mathcal{HV}}$, and the pixel intensity is bound accordingly~\cite{binaryImageHDC}. Grayscale images, on the other hand, may be processed with scalar-aware methods~\cite{HV_Design}, where each pixel value is mapped to a scalar $\boldsymbol{\mathcal{HV}}$~\cite{10024980}, enabling finer differentiation of 
intensity levels. Common approaches include thermometer encoding~\cite{thermometer}, unary-based methods~\cite{uHD, shoushtari2024all}, and deterministic encoding~\cite{Mehran-No-mul}. The SOTA research follows two general procedures for encoding 2D images:
1) utilizing raw pixel values \cite{SHEARer} and 2) leveraging ML 
techniques to extract and utilize features \cite{Zou2022_, sandwich, QuantHD}. 
The first approach leads to lightweight hardware designs but may have lower accuracy.
On the other hand, the latter approach achieves 
higher accuracy but requires feature engineering and additional 
hardware considerations for the feature extraction process. In the context of pixel-based processing, the SOTA works explore two types of information \cite{ParhiSurvey}: (i) pixel values, often referring to grayscale values, and (ii) pixel positions. The first one involves generating level $\boldsymbol{\mathcal{HV}}$s with numerical significance, while the second requires orthogonal $\boldsymbol{\mathcal{HV}}$s to represent unique pixel positions~\cite{Mehran-No-mul}.

Kleyko et al.~\cite{KleykoPART1} introduce fractional power encoding for encoding 2D images.
This approach involves producing 
two randomly assigned base $\boldsymbol{\mathcal{HV}}$s, denoted as $\boldsymbol{\mathcal{A}}$ and $\boldsymbol{\mathcal{B}}$, for the $x$ and $y$ axes, respectively. Subsequently, for each position $(u,v)$ in the image, the corresponding $\boldsymbol{\mathcal{HV}}$ denoted as $\boldsymbol{\mathcal{W}}$ is computed by raising the base $\boldsymbol{\mathcal{HV}}$ to the power of the respective coordinate values (bound together). Thus, the expression for the resulting $\boldsymbol{\mathcal{HV}}$ 
at position $(u,v)$ is: $\boldsymbol{\mathcal{W}} = \boldsymbol{\mathcal{A}}^u\oplus\boldsymbol{\mathcal{B}}^v$.
\vspace{-0.5em}


\subsubsection{Text-Based Encoding}
Text data is typically encoded at the character or token level. In language classification or NLP tasks, each character or word is represented with a unique, (nearly) orthogonal $\boldsymbol{\mathcal{HV}}$~\cite{AbbasLang}. Sequences are processed by permuting and bundling token vectors~\cite{Verges2025HDReview}. Circular shifts or position-based permutations are applied to capture order sensitivity. Tokenizers or language models (e.g., \textit{BERT}) can also serve as front-end encoders~\cite{AyarNICEllm, AyarGLSVLSI24llm}, producing semantic $\boldsymbol{\mathcal{HV}}$s from embeddings~\cite{austin_ESL}. When encoding semantic meaning, hyperdimensional systems must also address synonymy and polysemy, often using context-based disambiguation or embedding techniques~\cite{Berster2012WSD}.

\subsubsection{Signal, Time-Series, or Bio-symbolic Encoding}
For biosignals (EEG~\cite{9175328}, iEEG~\cite{hdTorch, sandwich, 8584751, abbas-ieeg}, ECG~\cite{10820450}) or other time-series inputs~\cite{9892158}, encoding involves representing temporal patterns while maintaining sequence sensitivity. Raw signals may be quantized into scalar bins~\cite{10546832}, each with a corresponding $\boldsymbol{\mathcal{HV}}$. To preserve time dynamics, sliding windows~\cite{HyperSense} and position shifts~\cite{9531748} are used before bundling. In some cases, frequency domain transforms are applied, and spectral features are encoded instead of raw signals~\cite{voiceHD}. Beyond physiological time-series data, biological symbolic inputs such as DNA sequences, genome strings, and methylation profiles can also be encoded using HDC. In this setting, nucleotides, $k$-mers, genomic positions, or methylation-derived features are mapped to atomic or position-aware \HV{s}, which are then composed for pattern matching, search, or classification~\cite{DNAimani,geniehd,BioHD,9586253,9786145,a13090233}. Thus, bio-symbol encoding extends sequence-sensitive HDC principles from temporal signals to biological alphabets and structured biomedical profiles.
Representative datasets and benchmark resources corresponding to these input categories are summarized in the Appendix.

\subsection{Features to Holistic Representation~\sttwo~and their Mapping~\stfour}
Several studies have proposed feature-driven $\boldsymbol{\mathcal{HV}}$ generation approaches, especially for scenarios where raw input is unavailable or high-level feature representations are preferred for holistic representations. These methods aim to construct meaningful $\boldsymbol{\mathcal{HV}}$s while preserving essential data properties such as similarity and variance. Rachkovskij et al.~\cite{thermometer} introduced multiple methods, including thermometric, subtractive-additive, and distributed stochastic encoding. These techniques prioritize preserving scalar relationships during mapping and are well-suited for numerical features. In addition, methods like \textit{Distance-Preserving Distributed Encoding}~\cite{Abbas_Class_recall} and Residue Hyperdimensional Computing (RHDC)~\cite{RHDC} have shown promise in encoding numerical data while maintaining semantic similarity. These methods adjust the $\boldsymbol{\mathcal{HV}}$ generation to ensure that small differences in input values correspond to small Hamming distances between $\boldsymbol{\mathcal{HV}}$s, improving robustness in real-valued feature domains. With foundational perspectives, Rachkovskij et al.~\cite{thermometer} discuss the potential of sparse binary distributed encoding in vector mapping from scalars.
To preserve the similarity of identical sequence elements in nearby positions and be equivariant regarding the sequence shifts, 
Rachkovskij
and Kleyko~\cite{9892462} propose a distributed sequence representation specific to the \textit{Fourier Holographic Reduced Representations} HDC model.
Their method model human identification of visual word images and measure the similarity of the $\boldsymbol{\mathcal{HV}}$s generated for each word.

From a more hardware perspective, Basaklar et al.~\cite{HV_Design} designed an optimized $\boldsymbol{\mathcal{HV}}$ generation technique for low-dimensional feature representations. Each feature is quantized into $M$ levels, and each level is assigned a bipolar $\boldsymbol{\mathcal{HV}}$ by flipping a fixed number of bits from the initial level. The final $\boldsymbol{\mathcal{HV}}$ is constructed by summing all level $\boldsymbol{\mathcal{HV}}$s corresponding to the input features. Dimensionality reduction is performed using t-distributed Stochastic Neighbor Embedding (t-SNE) algorithm, to maintain class separability. 

For the holistic representation, several non-linear encoding methods~\cite{RBF,rbf2}, 
inspired by the RBF kernel trick, have been
proposed.
These methods, introduced in various works such as \texttt{Dual}~\cite{Imani_Dual}, \texttt{FebHD}~\cite{febhd}, \texttt{DistHD}~\cite{disthd}, \texttt{NeuralHD}~\cite{neuralhd}, \texttt{RE-FHDC}~\cite{re-hfdc}, and \texttt{ManiHD}~\cite{manihd}, encode each dimension of the data by calculating the dot product of the feature vector ($F$) with a randomly generated vector ($B_i$) sampled from a Gaussian distribution with mean $\mu=0$ and standard deviation $\sigma=1$, denoted as $h_i = \cos{(B_i \cdot F)}$. The final holistic $\boldsymbol{\mathcal{HV}}$ is binarized using a \texttt{sign} function. The same method has been employed in \texttt{DistHD} and \texttt{CyberHD}~\cite{cyberhd} to identify and regenerate the insignificant dimensions, resulting in reduced dimensionality in the $\boldsymbol{\mathcal{HV}}$s and enhancing the overall system efficiency.
In \texttt{DistHD}, the $\boldsymbol{\mathcal{HV}}$s are generated based on 
the form $h_i = \cos{(B_i\cdot F +c)}\times \sin{(B_i \cdot F)}$ with $c \sim Uniform[0,2\pi]$.

As another feature-driven method, Salamat~et~al.~\cite{F5-HD} proposed a new $\boldsymbol{\mathcal{HV}}$ generation method for each feature vector $A_i$, which contains $\textit{N}$ elements of the form $A_i=\{a_1,a_2,...,a_N\}$. Each $a_i$ demonstrates a feature value $\boldsymbol{\mathfrak{F}_i}$ with \textit{l} levels as $a_i\in (\boldsymbol{\mathfrak{F}_1},\boldsymbol{\mathfrak{F}_2},...,\boldsymbol{\mathfrak{F}_l})$. In this case, the lowest feature value ($\boldsymbol{\mathfrak{F}_1}$, minimum value within a set of feature values) is 
assigned a random $\boldsymbol{\mathcal{HV}}$ called 
$\boldsymbol{\mathcal{H}_1}$ with size $D$ and 
the highest feature value ($\boldsymbol{\mathfrak{F}_l}$, maximum value within a set of feature values) 
is assigned a random $\boldsymbol{\mathcal{HV}}$ 
called $\boldsymbol{\mathcal{H}_\textit{l}}$ by randomly flipping $D/2$ 
bits. Starting from the first $\boldsymbol{\mathcal{HV}}$ 
the remaining $\boldsymbol{\mathcal{HV}}$s 
are generated by flipping $\frac{D/2}{\textit{l}-1}$ bits at each step. Afterward, each feature value $\boldsymbol{\mathfrak{F}_i}$ is mapped to its associated $\boldsymbol{\mathcal{HV}}$ as $\boldsymbol{\mathcal{H}_i}$. In order to take the position and/or temporal position of each input feature into account, the \textit{permutation} operation (\textbf{$\rho$}) is 
performed. The final holistic $\boldsymbol{\mathcal{HV}}$ denoted as $\boldsymbol{\mathcal{H}_C}$ 
is generated by $\boldsymbol{\mathcal{H}}_C=\sum_{t=1}^{T}\rho^{t}\boldsymbol{\mathcal{H}}_t$.

Symbolic holistic representations, such as the letter processing example, require 
orthogonal, i.e., dissimilar vectors. However, if the application involves 
numerical data, then 
similarity or correlation between some data points is required. For 
dissimilarity, random dense vectors with 50\% of the binary values with $1$s are utilized for orthogonality. 
When the $1$ and $0$ vector elements are equiprobable, the vector is called \textit{dense}.
For numerical data, 
closer 
values might need similar $\boldsymbol{\mathcal{HV}}$ representations, and a sparse vector is a better 
choice. If each $\boldsymbol{\mathcal{HV}}$ differs from the other by a few bits, then the sparsity can be obtained in the vector representation \cite{RACHKOVSKIJ201364, 8012434}. Hersche~et~al.~\cite{AbbasEmbedding}, and Cannings and Samworth~\cite{RandomProj} propose random projection methods for feature encoding using dense bipolar vectors. 
This encoding may be inefficient for hardware due to extensive addition and multiplication. Using sparse
random projection~\cite{HDC_Survey_Thomas_JAIR2021} for holistic representations, where only s\% of $\boldsymbol{\mathcal{HV}}$ elements are generated randomly, improves
efficiency.
Choosing the proper sparsity factor can significantly reduce 
the number of arithmetic operations. 
Further improvement could be achieved 
by preserving the benefit of sparsity in random projection and omitting random access in generating feature $\boldsymbol{\mathcal{HV}}$s as a hardware-friendly approach with 
predefined non-zero indices of $\boldsymbol{\mathcal{HV}}$s \cite{BRIC, LocalityEN}. In another work considering the orthogonality concept, Kleyko et al.~\cite{HoloGN} propose generating orthogonal $\boldsymbol{\mathcal{HV}}$ by using a base random $\boldsymbol{\mathcal{HV}}$
and performing multiple circular shifts in each iteration.
In other SOTA work proposed by Kleyko et al.~\cite{Abbas_Class_recall}, two types of encoding are employed for black-and-white images of letters and hand gesture recognition from a sequence of electromyography (\textbf{EMG}) signals as features.
In the \textit{Dissimilar Distributed Encoding} method, each feature is randomly assigned a dense binary $\boldsymbol{\mathcal{HV}}$.
Based on the feature value being 0 or 1, the corresponding $\boldsymbol{\mathcal{HV}}$ undergoes a circular $0$-bit or $1$-bit shift, respectively. All of these $\boldsymbol{\mathcal{HV}}$s will be orthogonal to each other, ensuring distinctiveness.
In the \textit{Distance-Preserving Distributed Encoding} method, each feature is assigned an $\boldsymbol{\mathcal{HV}}$ with a randomly determined number of bit-flips.
Unlike the first 
method, the adjacent $\boldsymbol{\mathcal{HV}}$s exhibit similarities to each other as their distance is preserved.

In holistic representation from features, the $D$ dimension is also a critical metric for the hardware efficiency. Basaklar~et~al.~\cite{HV_Design} proposed
an optimized $\boldsymbol{\mathcal{HV}}$ design with lower dimensionality compared to conventional high-dimension HDC for resource-restrained wearable IoT devices.  
They assumed a baseline HDC structure with 
input features $F = \{f_1,f_2,...,f_N\}$ and $S$ training samples. Each training sample $x_s$ in sample space $S$ is an $N$-tuple of the form $x_s = \{x_s^1,x_s^2,...,x_s^N\}$ in which each $x_s^n$ is the corresponding value of the feature $f_n$. To construct the sample $\boldsymbol{\mathcal{HV}}$s, first, the level $\boldsymbol{\mathcal{HV}}$s are generated by quantizing each feature value to $M$ levels. The first level of the $f_n$ is assigned to a random bipolar $D$-dimension $\boldsymbol{\mathcal{HV}}$ ($L_n^1$) and the other consecutive level $\boldsymbol{\mathcal{HV}}$s (\,$L_n^m , \forall  m \in \{2,...,M\}$)\, are generated by randomly flipping $b = D/2(M-1)$ bits. Finally, the sample $\boldsymbol{\mathcal{HV}}$ ($X_s$) is generated by adding the associated level $\boldsymbol{\mathcal{HV}}$ assigned to the related input feature as $X_s = \sum_{n=1}^{N} L_n^m \ \forall m \in \{1,...,M\}$. In a similar manner, the class $\boldsymbol{\mathcal{HV}}$s are generated by combining sample $\boldsymbol{\mathcal{HV}}$s belonging to the same class. This method uses varying bit flips for each $\boldsymbol{\mathcal{HV}}$ level, treating $\boldsymbol{\mathcal{HV}}$ generation as a multi-objective optimization problem, with goals to maximize training accuracy and minimize class similarity.


Local Binary Pattern (LBP) encoding is a feature-dependent method originally developed for texture analysis~\cite{10.1007/978-3-540-24670-1_36}, which has also been adapted to generate $\boldsymbol{\mathcal{HV}}$s. In essence, LBP examines an input signal (e.g., a window of a time series or image patch) by thresholding each point against its local neighbors to produce a binary pattern. This binary code can then be interpreted as an $\boldsymbol{\mathcal{HV}}$ (or used as an index into an $\boldsymbol{\mathcal{HV}}$). For example, the 1D-LBP technique proposed by Kaya et al.~\cite{kaya20141d} generates binary features from EEG signal segments. These patterns are then used directly as $\boldsymbol{\mathcal{HV}}$s to encode EEG or other biomedical signals in HDC systems. In HDC contexts, an LBP-based $\boldsymbol{\mathcal{HV}}$ might be produced by initializing a length-$D$ all-zero vector and setting ones in positions corresponding to ``local binary 1'' outputs (essentially resulting in a sparse code). LBP-based $\boldsymbol{\mathcal{HV}}$ generation incorporates domain knowledge by preserving locality structure: It encodes whether a signal rises or falls compared to its neighboring values. This has proven effective in neuromorphic and IoT applications. For instance, the Laelaps neuromorphic processor~\cite{sandwich} and subsequent works by Burrello et al.~\cite{8584751,abbas-ieeg} employ LBP encoding to convert sensor inputs into $\boldsymbol{\mathcal{HV}}$s.
The primary advantage of LBP-generated $\boldsymbol{\mathcal{HV}}$s is that they capture meaningful local pattern information (edges, transitions, etc.),  making the resulting representations more expressive than random $\boldsymbol{\mathcal{HV}}$s for certain data (e.g., images, time-series). They also provide deterministic encoding--the same local pattern always results in the same $\boldsymbol{\mathcal{HV}}$, improving consistency. However, LBP has certain limitations. It is data-type specific, primarily applicable to ordered data such as images or 1-D sequences, and is not suitable for arbitrary symbolic inputs. Also, the raw LBP codes typically have relatively low dimensionality (often only a few dozen bits), so they must be projected or expanded to match the higher dimensionality $D$ required in most HDC models. Despite these drawbacks, LBP-based encoding remains a lightweight, interpretable, and effective approach that integrates well with HDC, as evidenced by its use in several biosignal HDC systems.


Deep hashing-based feature $\boldsymbol{\mathcal{HV}}$ generation is another method that uses trained NNs to produce embeddings (hash codes) that serve as $\boldsymbol{\mathcal{HV}}$s. In this approach, a deep network (often a convolutional neural network--CNN or multi-layer perceptron) is 
trained to output a compact binary code (e.g., 32 or 64 bits) for each input, preserving similarity relationships in that code space. These output codes – effectively high-dimensional binary feature vectors – become the $\boldsymbol{\mathcal{HV}}$s for the inputs. 
Mitrokhin et al.~\cite{symbolichdc} introduced this approach by using hashing NNs to map images to binary $\boldsymbol{\mathcal{HV}}$s, followed by symbolic reasoning on those $\boldsymbol{\mathcal{HV}}$s. 
Here, the NN learns to extract 
meaningful features from the input data, resulting in data-dependent, informative 
$\boldsymbol{\mathcal{HV}}$s, in contrast to 
purely random vectors. Once generated, these $\boldsymbol{\mathcal{HV}}$s can be processed using standard HDC operations, combining deep learning and HDC. The method 
brings the advantage of learned embeddings--semantically or visually similar inputs will map to similar $\boldsymbol{\mathcal{HV}}$s, improving the efficiency and accuracy of subsequent HDC operations (like classification).
It essentially biases the $\boldsymbol{\mathcal{HV}}$ space to better reflect the data distribution. Another benefit is dimensionality reduction, producing 
compact $\boldsymbol{\mathcal{HV}}$s (e.g., 128 bits), compared to the $1{,}000+$ dimensions typically used in random HDC. 
However, this approach also introduces certain limitations. Most notably, it requires a trained NN, which entails computational cost and compromises the simplicity and energy efficiency often associated with pure HDC systems.
The encoding process is no longer 
transparent or random; it depends on trained weight parameters that must be 
stored. Additionally, the approach ties the $\boldsymbol{\mathcal{HV}}$ representations to a specific data domain (e.g., tuned for images or text), losing some generality. Nonetheless, deep hashing has proven effective in integrating with HDC frameworks: e.g., a deep network can generate initial $\boldsymbol{\mathcal{HV}}$s which are then fed into an HDC inference layer. Overall, deep hash-based $\boldsymbol{\mathcal{HV}}$ generation produces task-tailored $\boldsymbol{\mathcal{HV}}$s by combining the representational power of deep learning with the combinatorial advantages of HDC, achieving improvements in tasks like vision and language where learned features consistently outperform raw random encodings.

In addition to handcrafted or signal-processing-based features, learned neural representations can also serve as front-end features for HDC encoding. In this setting, an NN is not necessarily used as the final classifier; instead, its intermediate or output activations provide structured feature vectors that are subsequently mapped into the HDC space. For instance, the output activations of an NN layer can be directly represented as a $\boldsymbol{\mathcal{HV}}$. Instead of training a network specifically for a compact hash, this method uses a pre-trained NN (or multiple networks) to encode its final high-dimensional activation pattern as an $\boldsymbol{\mathcal{HV}}$. Sutor et al.~\cite{gluing} describe encoding the entire vector of output signals from a NN’s final layer (just before its classifier) into an $\boldsymbol{\mathcal{HV}}$, which can then be combined with others by bundling~\cite{gluing}.
In practice, if an NN produces an $N$-dimensional real output (for example, logits for $N$ classes or feature embeddings from a penultimate layer), that $N$-length vector can be converted or interpreted as an $\boldsymbol{\mathcal{HV}}$ in the HD space (possibly by binarizing each dimension or by directly using the real outputs as components). Those $\boldsymbol{\mathcal{HV}}$s can be bundled across multiple networks or time steps to form consensus or accumulated $\boldsymbol{\mathcal{HV}}$s.
The key idea is to fuse NN predictions at the $\boldsymbol{\mathcal{HV}}$ level. One advantage of this approach is to model ensembling and incremental learning: by representing different NN outputs as $\boldsymbol{\mathcal{HV}}$s, one can combine them through superposition (majority or consensus summation) to improve overall decision robustness. This allows multiple models to collaborate without altering their internal parameters – the fusion occurs in hyperdimensional space as a holistic approach. It also enables online learning: as new network outputs come in, the $\boldsymbol{\mathcal{HV}}$ can be updated by adding and normalizing (which corresponds to a kind of memory of past outputs). Sutor et al. showed that combining multiple vision CNNs in this way improves classification accuracy while inheriting the fast, iterative update properties of HDC.
Another benefit is that this method can leverage powerful deep features (learned by networks on large data) within an HDC framework – essentially getting the “best of both worlds.” The limitations include the dependence on those NNs (one must have them trained beforehand), and the size of the output vectors might be relatively low-dimensional (hundreds of neurons) compared to typical HDC $\boldsymbol{\mathcal{HV}}$s. To truly embed an $N$-dimensional network output into a $D$-dimensional $\boldsymbol{\mathcal{HV}}$ (with $D \gg N$), additional expansion or encoding might be necessary (such as binding the output with a random projection $\boldsymbol{\mathcal{HV}}$ to inflate dimensionality). Moreover, if using raw real-valued outputs, one might binarize or thermometrically encode them to fit binary HDC operations. Despite these considerations, neural-output $\boldsymbol{\mathcal{HV}}$s have been effective for hybrid systems. For example, Daniels et al.~\cite{daniels2023efficient} combine deep network features with hyperdimensional classifiers for efficient and parallel learning. Overall, this method treats an NN as a feature generator and uses HDC to merge and manipulate those feature vectors, capitalizing on HDC’s noise tolerance and combinatorial flexibility to augment NN-based AI systems.


\subsection{Source-Driven Hypervector Encoding~\stthree~and their Mapping~\stfour}
In what follows, we go deeper into the details of various generation source-aware encoding and the corresponding mapping.

\vspace{-0.5em}
\subsubsection{Pseudo-Random}
The most commonly used source for $\boldsymbol{\mathcal{HV}}$ generation is 
based on pseudo-randomness. In this method, each component of a $\boldsymbol{\mathcal{HV}}$ is determined using a pseudo-random number generator, which typically produces independent and identically distributed binary or bipolar values. 
Due to the high dimensionality of $\boldsymbol{\mathcal{HV}}$, two independently generated pseudo-random $\boldsymbol{\mathcal{HV}}$s are \textit{nearly} orthogonal, 
making them ideal for representing distinct symbols and maintaining separation between encoded data. However, the near-orthogonality also brings a limitation for the best-performing orthogonal vector selections, requiring an optimization. 
Numerous HDC frameworks utilize pseudo-random initialization for their item memory $\boldsymbol{\mathcal{HV}}$s due to its simplicity and effectiveness in achieving near-orthogonality without significant computational overhead.
For example, \texttt{VoiceHD}~\cite{voiceHD} and \texttt{AdaptHD}~\cite{adapthd} frameworks extensively utilize pseudo-random $\boldsymbol{\mathcal{HV}}$s to represent distinct phonetic features. 
Similarly, \texttt{DOMINO}~\cite{domino} and \texttt{ST-HDC}~\cite{ST-HDC} 
leverage pseudo-random $\boldsymbol{\mathcal{HV}}$s to encode sensor data and textual information, demonstrating robust performance in 
classification tasks. 

The primary advantages of pseudo-random $\boldsymbol{\mathcal{HV}}$s lie in their computational efficiency, ease of implementation, and statistical properties, without complex algorithmic overhead. Additionally, the deterministic nature of pseudo-random generation (via seed-controlled pseudo-random number generators) enhances reproducibility and facilitates debugging in research and hardware implementations.
However, pseudo-random $\boldsymbol{\mathcal{HV}}$s inherently lack semantic structure or data-dependent similarity. 
Similar input data 
are not necessarily mapped to similar $\boldsymbol{\mathcal{HV}}$s, which can potentially limit performance in tasks where semantic or proximity-based relationships are crucial. In addition, reliance on pseudo-randomness can lead to excessive model execution and runtime to achieve the desired level of accuracy. 
Furthermore, hardware implementations may incur overhead from pseudo-random number generation circuitry for edge learning, affecting resource utilization and power efficiency~\cite{ID-VSA_Moghadam_TVLSI2026}.
Despite these limitations, pseudo-random $\boldsymbol{\mathcal{HV}}$s remain widely adopted due to their favorable balance of simplicity, orthogonality, and computational efficiency, forming a fundamental baseline for evaluating more advanced encoding methods.

\vspace{-0.5em}
\subsubsection{Gaussian}
Instead of using binary components, $\boldsymbol{\mathcal{HV}}$s can be drawn from a continuous distribution, such as the Gaussian distribution. In this approach, each $\boldsymbol{\mathcal{HV}}$ is a D-dimensional real-valued vector (often with components sampled from a standard normal distribution $\sim\mathcal{N}(0,1)$) that may be normalized to unit length. Like random binary $\boldsymbol{\mathcal{HV}}$s, 
Gaussian-distributed $\boldsymbol{\mathcal{HV}}$s in high dimensions are nearly orthogonal. 
The use of Gaussian $\boldsymbol{\mathcal{HV}}$s is particularly beneficial when dot-product or cosine similarity is used as the comparison metric, as it avoids discretization and enables direct interfacing with analog signals or neuromorphic hardware. For example, Hersche et al.~\cite{AbbasEmbedding,10.1145/3370748.3406560} utilize real-valued $\boldsymbol{\mathcal{HV}}$s drawn from a Gaussian distribution to encode biosignal features in an analog HDC classifier.
One key advantage of Gaussian $\boldsymbol{\mathcal{HV}}$s is their compatibility with standard linear algebra operations, enabling superposition and similarity computations via efficient vector–matrix operations that can benefit from hardware acceleration.  
Moreover, they can represent graded information by the actual component values, rather than bit flips. However, 
using continuous $\boldsymbol{\mathcal{HV}}$s sacrifices the bitwise simplicity and noise-immunity that binary $\boldsymbol{\mathcal{HV}}$s provide. Implementations must handle floating-point (or fixed-point) precision and potentially re-normalize $\boldsymbol{\mathcal{HV}}$s to maintain vector magnitude. In summary, Gaussian $\boldsymbol{\mathcal{HV}}$s offer a continuous 
representation that retains the randomness and orthogonality of binary $\boldsymbol{\mathcal{HV}}$s, while enhancing integration for certain computing substrates, at the cost of increased hardware complexity and potential degradation in robustness  due to cumulative numerical errors during operations like iterative bundling. 

\vspace{-0.5em}
\subsubsection{Thermometer/Unary Code}
Thermometer encoding (also known as unary encoding) is a technique for representing a scalar value in a distributed binary format that preserves ordinal relationships. To generate an $\boldsymbol{\mathcal{HV}}$ via thermometer encoding, one decides on an ordering of $\boldsymbol{\mathcal{HV}}$ components and then sets the first $k$ components to 1 (with the remaining components set to 0) to encode a value $k$ (out of a maximum $D$).
For example, if $D=10$ and the input value 
corresponds to 4, the resulting $\boldsymbol{\mathcal{HV}}$ would be 1111000000~\cite{uHD}. This encoding scheme ensures that similar values produce $\boldsymbol{\mathcal{HV}}$s with a high degree of overlap. For example, the $\boldsymbol{\mathcal{HV}}$s for values 4 and 5 share several leading 1s. As a result, numeric proximity is naturally preserved when measured using Hamming or cosine similarity. In HDC systems, thermometer-coded $\boldsymbol{\mathcal{HV}}$s are used to encode continuous quantities (e.g., sensor readings, pixel intensities) into binary $\boldsymbol{\mathcal{HV}}$s while maintaining similarity relationships~\cite{Abbas-ExG}.
Rosato et al.~\cite{9533805} demonstrate an HDC-based classification system that uses thermometer-coded $\boldsymbol{\mathcal{HV}}$s to represent real-valued features.
The key advantage of thermometer encoding is that it introduces gradual similarity: the inner product or overlap between two thermometer $\boldsymbol{\mathcal{HV}}$s 
reflects the distance between the original scalar values. This property can improve accuracy in tasks where ranking or ordinal relationships are important, compared to using completely random vectors for numerical data. Thermometer $\boldsymbol{\mathcal{HV}}$s are also simple, deterministic, and computationally efficient to generate~\cite{shoushtari2024all}. However, there are limitations. The $\boldsymbol{\mathcal{HV}}$s are not mutually orthogonal (in fact, they are highly correlated by design), which can reduce vector space capacity if overused. 
There is also a resolution trade-off; the dimensionality $D$ determines how many distinct levels can be represented (a larger $D$ allows finer granularity but increases computational cost). In practice, thermometer encoding 
is often combined with other techniques 
(e.g., random projections or bundling) to balance order preservation with orthogonality.
Overall, thermometer $\boldsymbol{\mathcal{HV}}$s offer a non-linear, order-preserving mapping of scalar data into HDC space 
and have been successfully employed for encoding sensor inputs for HDC classifiers.

\vspace{-0.5em}
\subsubsection{Random Exponential}
Random exponential $\boldsymbol{\mathcal{HV}}$ generation 
involves constructing $\boldsymbol{\mathcal{HV}}$s whose components are random complex phases located on the unit circle (recall Table~\ref{HV_types_ops}).
These $\boldsymbol{\mathcal{HV}}$s can be viewed as a holistic phasor representation. Similar to real-valued random vectors, two independently generated random phasor $\boldsymbol{\mathcal{HV}}$s are nearly orthogonal--their real inner product is close to zero.  
One motivation for using complex $\boldsymbol{\mathcal{HV}}$s is to support circular convolution binding via element-wise multiplication: when phases encode information, multiplying two $\boldsymbol{\mathcal{HV}}$s adds their phases, implementing the binding operation common in VSA (e.g., in Holographic Reduced Representations). For instance, the QHD framework for reinforcement learning represents states with complex $\boldsymbol{\mathcal{HV}}$s (random unit-length phasors)~\cite{QHD}.
A key advantage of random exponential $\boldsymbol{\mathcal{HV}}$s is that they naturally represent rotational or periodic data. A rotation by angle $\phi$ in the data space corresponds to a fixed phase shift across all components, yielding a structured transformation in the $\boldsymbol{\mathcal{HV}}$. This encoding also facilitates algebraic operations, where phase addition corresponds to an analog of \texttt{XOR} in the complex domain. Additionally, complex $\boldsymbol{\mathcal{HV}}$s can be beneficial in hardware implementations that leverage complex arithmetic or when modeling oscillatory phenomena. A limitation, however, is that working with complex-valued $\boldsymbol{\mathcal{HV}}$s is 
more complicated than binary or real-valued $\boldsymbol{\mathcal{HV}}$s: similarity measures must be carefully defined  (e.g., using the complex dot product or the real part of the correlation), and many HDC hardware platforms do not natively support complex arithmetic. Moreover, the interpretability of each dimension’s phase is limited (similar to random real-valued $\boldsymbol{\mathcal{HV}}$s). Despite these challenges, random exponential $\boldsymbol{\mathcal{HV}}$s have been used in scenarios where phase information or rotational binding is required, effectively expanding HDC beyond real vector spaces into complex domains.

\vspace{-6pt}

\subsubsection{Low-Discrepancy (LD) Sequences}
LD sequence-based \HV{} generation produces \HV{s} in a deterministic, quasi-random manner. Rather than relying on a pseudo-random number generator, methods in this category utilize quasi-random sequences such as Sobol \cite{Mehran-No-mul,uHD,Sobol_Aygun_TCAD}, van der Corput \cite{shoushtari2024all,ID-VSA_Moghadam_TVLSI2026}, or Hadamard matrices~\cite{AddOn-IHD_Masum_DAC2025} to generate the \HV{} components. These sequences can reduce the clustering and gaps typical of pseudo-random sampling, so the resulting \HV{s} fill the high-dimensional unit cube more uniformly.
The primary advantage of LD-based generation lies in the quality of the point set it produces in a single pass. Because the sequence is built to minimize discrepancy, one deterministic sweep yields \HV{} components that cover the space evenly, avoiding the clustering and gaps that any individual pseudo-random or true-random draw can exhibit. With independent and identically distributed (i.i.d.) sampling, the coverage of a single draw depends on the chosen seed, so a reliably well-distributed set can require trying several seeds or averaging over multiple runs; an LD sequence guarantees even coverage on the first pass and removes this dependence. 
Reproducibility is a second, separate benefit. A fixed-seed pseudo-random generator is itself reproducible, so the gain here is specifically over true-random sources, which cannot be reproduced at all, and over deployments that reseed between runs: a deterministic LD generator returns the same well-distributed set on every execution without storing entropy or managing seeds, which is valuable for benchmarking and for hardware where reseeding or re-execution carries a real cost.
Moghadam et al. \cite{Mehran-No-mul,shoushtari2024all,ID-VSA_Moghadam_TVLSI2026} demonstrate that LD-generated \HV{s} are sufficient in a single execution pass, whereas pseudo-random methods may require multiple runs to reach a stable accuracy level. 
However, this deterministic generation introduces its own hardware overhead; unlike simple Linear Feedback Shift Registers (LFSRs) used for pseudo-random generation, LD sequences often require Look-Up Tables (LUTs) for direction numbers or counters, whose memory footprint scales linearly with $D$.\\ 
With respect to accuracy, the works in \cite{Mehran-No-mul,uHD,Sobol_Aygun_TCAD,shoushtari2024all,ID-VSA_Moghadam_TVLSI2026,AddOn-IHD_Masum_DAC2025,ayarSobol,AMS-HD_TBME2026} report improved accuracy-versus-dimensionality trade-offs compared to pseudo-random baselines on specific benchmarks, including language and speech recognition, time series, image, code, and biomedical classification. These gains are attributed to the reduced probability of accidental inter-vector correlations at moderate dimensionalities ($D \approx 512 - 2{,}048$), where random \HV{s} may not reliably achieve near-orthogonality. 
These benefits, however, 
are specific to the evaluated tasks and dimensionality ranges. At high dimensionalities ($D \ge 5{,}000$), pseudo-random vectors are already near-orthogonal with very high probability, which substantially narrows the gap between LD and random initialization. Similarly, while the 
comparisons in \cite{Mehran-No-mul,uHD,shoushtari2024all,Sobol_Aygun_TCAD,ID-VSA_Moghadam_TVLSI2026,AddOn-IHD_Masum_DAC2025, ayarSobol} involve a limited set of datasets, broader validation across highly complex domains such as graph or recursive structure encoding, multi-modal sensor fusion, and distributed federated learning protocols remains an open research challenge. Claims about LD-sequence superiority should therefore be interpreted as task- and dimension-conditional rather than universal.

\noindent A further consideration is predictability. Deterministically generated \HV{s} are, by construction, fully reproducible given the generator parameters. While this aids debugging and reproducibility, it also means that the \HV{} space is not entropic in the information-theoretic sense: an adversary with knowledge of the generator type and seed can reconstruct the entire item memory. In security-critical HDC applications, such as privacy-preserving inference \cite{9869459,Prive-HD,lightningtalk} or authentication~\cite{cai2022hyperlock,10044775}, this predictability may increase vulnerability relative to systems built on high-entropy pseudo- or true-randomness. This concern does not diminish the utility of LD sequences for standard classification tasks, but it should inform deployment decisions.

\noindent In summary, LD sequences represent a promising and theoretically motivated direction for structured \HV{} initialization rather than a universally superior substitute for pseudo- or true-random generation. Their observed benefits depend on the application domain, dimensionality, number of classes or symbols, similarity metric, encoding pipeline, and hardware realization. 
Table~\ref{tab:ld_vs_pr} summarizes the conditions and application constraints that may favor LD sequence generation, pseudo- or true-random initialization, 
highlighting trade-offs in dimensionality scaling, reproducibility, hardware-generator complexity, and security. Broader validation remains an open challenge across graph and recursive-structure encoding, multi-modal sensor fusion, federated or distributed learning, adversarial and security-sensitive deployments, and diverse hardware substrates. 



\color{black}

\begin{table*}[t]
\centering
\large

\caption{Qualitative Tradeoffs of LD Sequences vs. Pseudo-Random and True-Random Initializations}
\vspace{-1em}
\setlength{\aboverulesep}{0pt} 
\setlength{\belowrulesep}{0pt} 
\setlength{\extrarowheight}{0pt}

\resizebox{\linewidth}{!}{%
\begin{tabular}{@{} 
    >{\raggedright\arraybackslash}m{1.7cm} 
    >{\raggedright\arraybackslash}m{3.5cm} 
    >{\raggedright\arraybackslash}m{2.5cm} 
    >{\raggedright\arraybackslash}m{2.8cm} 
    >{\raggedright\arraybackslash}m{4cm} 
    >{\raggedright\arraybackslash}m{4cm} 
    >{\raggedright\arraybackslash}m{3.8cm} 
    >{\raggedright\arraybackslash}m{3.9cm} 
@{}}
\toprule
\textbf{\HV{} Source} & 
\textbf{Dimensionality of \HV{s} ($\mathbf{D}$)} & 
\textbf{Run-to-run reproducibility} & 
\textbf{Executions for validation} & 
\textbf{Security-critical deployment} & 
\textbf{Hardware generator complexity} & 
\textbf{High-$\mathbf{D}$ scaling overhead} & 
\textbf{Current validation gaps} \\
\midrule

\rowcolor{Lavender!30}
\textbf{Pseudo-Random} & 
Effective at sufficiently large $D$; near-orthogonality holds with high probability & 
Reproducible with fixed seed; seed-dependent otherwise & 
Usually needs multiple seeds for statistical confidence & 
Suitable if seeds are protected and entropy is sufficient & 
Low--Medium: LFSR+comparator & 
Usually modest; depends on parallelism and generator period & 
Standard baseline, but seed- and workload-dependent \\

\textbf{True Random} & 
Effective at large $D$; task- and technology-dependent & 
Weak unless stored or replayed & 
Requires repeated trials or stored realizations & 
Unpredictable; attractive for security-sensitive settings & 
Depends on entropy source or emerging device & 
Depends on generation, storage, and readout cost & 
Limited HDC-specific validation across workloads and hardware \\

\rowcolor{Lavender!30}
\textbf{LD Sequences} & 
Reported benefits mainly at moderate $D$; task-dependent & 
Deterministic and reproducible by construction & 
May reduce repeated seed trials; validation remains task-dependent & 
Potentially vulnerable if the generator is known & 
Medium--High: LUT/counter or deterministic logic & 
Implementation-dependent; LUT/counter overhead may grow with $D$ & 
Broader validation needed across tasks, datasets, dimensions, and hardware \\

\bottomrule
\end{tabular}%
}
\label{tab:ld_vs_pr}
\end{table*}

\subsubsection{Non-linear approaches}
Non-linear \HV{} generation methods aim to encode real-valued or vector inputs into hyperdimensional space through a fixed non-linear transform. Typically, this process involves applying random projections followed by a non-linear activation (such as thresholding) to produce a binary $\boldsymbol{\mathcal{HV}}$. The approach was pioneered by Rasanen~\cite{okko}, who developed a method to 
encode continuous multivariate sensory inputs into $\boldsymbol{\mathcal{HV}}$s. In his framework, a set of random hyperplanes (or random neurons) 
partitions the input space. A given continuous input vector is projected onto this ensemble, 
and each unit outputs a 1 or 0 depending on whether the input lies above or below its threshold. The resulting bits form the $\boldsymbol{\mathcal{HV}}$ corresponding to that input. Conceptually, this method resembles a one-layer random NN (an extreme learning machine) that produces a binary, HDC-friendly embedding. The $\boldsymbol{\mathcal{HV}}$s generated in this way are non-linear functions of the original input, allowing complex input relationships to be captured beyond simple linear encodings like thermometer coding. Chang et al.~\cite{8771622} apply a similar strategy for physiological sensor fusion, mapping multiple Biosignals (EEG, GSR, etc.) into a single $\boldsymbol{\mathcal{HV}}$ by passing the signals through 
non-linear transform modules (e.g., comparators, saturating functions) before high-dimensional encoding.
The major advantage of non-linear $\boldsymbol{\mathcal{HV}}$ generation is its expressiveness; it can encode higher-order combinations of features. Two inputs that are only slightly different might produce very different $\boldsymbol{\mathcal{HV}}$s if they fall on opposite sides of a random decision boundary, thereby allowing the model to distinguish subtle patterns (much like random forest or random basis functions in kernel machines). It also allows HDC to handle analog inputs naturally without requiring a fine quantization; 
The random thresholds 
discretize the space in a high-dimensional nonlinear fashion. Moreover, 
non-linear transforms often preserve local similarity to some extent (by using many thresholds, most of which an input pair will agree on if the inputs are close). Non-linear transforms can thus approximate kernel similarity in the $\boldsymbol{\mathcal{HV}}$ space (drawing inspiration from random Fourier features and related theory~\cite{RBF}). However, there are trade-offs. 
The $\boldsymbol{\mathcal{HV}}$s from non-linear encoders 
are harder to interpret, as each bit reflects an arbitrary threshold on a random combination of original features. There are also efficiency concerns: generating $\boldsymbol{\mathcal{HV}}$ requires computing multiple projections of the input (though these can be made very simple, e.g., weighted sums or even just comparisons to random prototypes). As an example, this might entail more operations than a thermometer code in hardware. Another consideration is that once fixed, the random projection is static – it may not be optimal for all inputs. Learning or tuning the projections, as done in an NN, could further improve performance, but would move the approach closer to learned models. 
Despite these challenges, non-linear generation has proven effective in practice, particularly in 
one-shot learning of continuous inputs where non-linear encoding outperforms linear ones. 
In summary, non-linear $\boldsymbol{\mathcal{HV}}$ generation uses random but fixed transformations to embed data into $\boldsymbol{\mathcal{HV}}$s, combining the power of random features with the simplicity of HDC.

\subsection{Record-based and $N$-gram-based HDC Composition~\stfive}
\label{rb_ng_encoding}

Following the multi-stage taxonomy in Fig.~\ref{fig:pipeline_taxonomy}, this subsection focuses on Stage-5~\stfive, where mapped \HV{s} are composed into holistic representations. At this stage, 
the HDC literature commonly relies on two fundamental composition templates: record-based encoding and $N$-gram-based encoding~\cite{Imani_Hierarchical,HDC_Arch,Abbas_Biosignal,ParhiSurvey}. These approaches are not tied to a single input modality; rather, they specify how feature, position, level, or symbol \HV{s} are algebraically combined after the initial mapping step. Record-based method, for each input feature $i$, employs position (or symbol) $\boldsymbol{\mathcal{HV}}$s denoted as $\boldsymbol{\mathcal{P}_i}$, which exhibit orthogonality with respect to each other. Level $\boldsymbol{\mathcal{HV}}$s denoted as $\boldsymbol{\mathcal{L}_i}$, demonstrate correlation for neighboring pairs. Equation~\ref{eq1} expresses the binary \HV{} setting, an \texttt{XOR} operation binds each level and position vector and the summation accumulates these bound vectors across all $N$ features. The \texttt{majority} function applies the binarization step.
\vspace{-0.5em}

\begin{equation}
\label{eq1}
\boldsymbol{\mathcal{H}} = \text{majority} \left( \sum_{i=1}^{N}(\boldsymbol{\mathcal{L}_i}\oplus\boldsymbol{\mathcal{P}_i}) \right).
\end{equation}

The $N$-gram-based method uses the permutation operation $\rho$ for each
level $\boldsymbol{\mathcal{HV}}$ ($\boldsymbol{\mathcal{L}_i}$).
In this approach, the resulting class $\boldsymbol{\mathcal{HV}}$ is generated by \texttt{XOR}ing 
the permuted level $\boldsymbol{\mathcal{HV}}$s (binary \HV{s}). The final $\boldsymbol{\mathcal{HV}}$ in this case is represented in Equation~\ref{eq2}, where $\rho^k$ represents the $k$-times circular permutation of a $\boldsymbol{\mathcal{HV}}$, used to preserve positional information in the $N$-gram encoding:

\begin{equation}
\label{eq2}
    \boldsymbol{\mathcal{H}} = \boldsymbol{\mathcal{L}_1}\oplus\rho\boldsymbol{\mathcal{L}_2}...\oplus\rho^{N-1}\boldsymbol{\mathcal{L}_N}.
\end{equation}

\begin{figure*}[t]
\centering
\includegraphics[width=\linewidth]{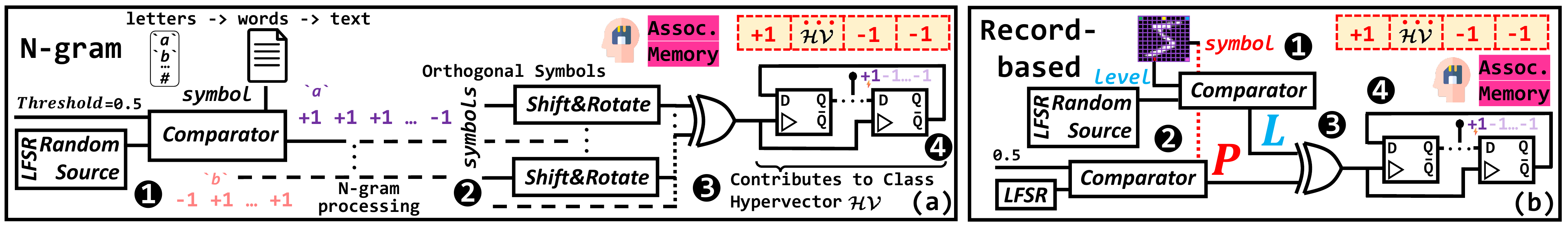}
\vspace{-2em}
\caption{Text and image processing pipelines in HDC. (a) Text Processing: Letters are converted into orthogonal symbol vectors via a pseudo-random source and a comparator, followed by $N$-gram encoding: \textit{shifting-and-rotating}, \textit{XOR}, and \textit{accumulate} operations. (b) Image Processing: Pixel values are processed using intensity ($L$, level) and symbol-based position ($P$) vector encoding.}
\label{VSA_general} 
\end{figure*}

Fig.~\ref{VSA_general} illustrates the general hardware architecture (relation from~\stfive~to~\stsix) of the \textit{$N$-gram} and \textit{record-based} approaches, applied to text data (Fig.~\ref{VSA_general}(a)) and image processing (Fig.~\ref{VSA_general}(b)), respectively. The first step in HDC is vector generation. In Fig.~\ref{VSA_general}(a), this involves generating vectors for symbolic values (e.g., letters), while in Fig.~\ref{VSA_general}(b), two types of vectors are generated: one for positions ($\boldsymbol{P}$, symbolic) and another for pixel intensity scalars ($\boldsymbol{L}$, numerical). Symbolic representations are typically created by comparing a random source to an unbiased threshold (e.g., 0.5 probability) across $D$ dimensions. Scalar values, in contrast, are proportionally mapped to vector representations. Both types of \HV{s} can be dynamically generated using LFSRs~\cite{lfsr_first_time}, or retrieved from memory as pre-stored \HV{s}~\cite{Karunaratne2020}.

The encoding stage then applies a series of operations: \textit{permutation} (Step 2 in Fig.~\ref{VSA_general}(a), used only in $N$-gram encoding), \textit{binding} (Step 3 in both Fig.~\ref{VSA_general}(a) and (b)), and \textit{bundling} (Step 4 in both cases)~\cite{Kazemi_2022}. The permutation step enhances orthogonality by cyclically shifting vector elements, thereby encoding positional information (e.g., no shift, 1-shift, 2-shift, etc.)~\cite{AbbasLang}. Although pseudo-random $\boldsymbol{\mathcal{HV}}$ generation is widely used, it may result in low orthogonality between vectors, which can degrade downstream accuracy~\cite{shoushtari2024all}. The binding operation combines multiple $\boldsymbol{\mathcal{HV}}$s via element-wise multiplication or \texttt{XOR}, while the bundling step aggregates them into a single holistic vector through a majority-based operation, preserving composite semantics in a memory-efficient form. 


\subsection{Hardware-Aware Encoding Approaches~\stsix}
Beyond data-specific strategies, several studies have proposed generalized feature-driven (refer to pre-extracted features) approaches that consider both the nature of the data and the underlying ML features, along with the corresponding hardware requirements. These methods often shift the focus from encoding raw input to constructing holistic representations under hardware constraints. Hardware efficiency plays a crucial role in real-world HDC applications. Several techniques have been introduced to optimize encoding operations for hardware implementations, especially in FPGA or IMC environments~\cite{hdnn}. To reduce memory utilization, Salamat et al.~\cite{F5-HD} suggested applying permutations not at the bit level, but in segments of size S. This segment-based permutation significantly reduces Block RAM (BRAM) usage on FPGAs, facilitating efficient hardware realization. Schmuck et al.~\cite{abbas-remat} proposed a memory-free $\boldsymbol{\mathcal{HV}}$ generation module using seed $\boldsymbol{\mathcal{HV}}$s and logic operations. Instead of storing pre-generated $\boldsymbol{\mathcal{HV}}$s, their approach rematerializes $\boldsymbol{\mathcal{HV}}$s on the fly using majority-based operations. Their design demonstrates improvements in area-efficiency (2.39$\times$ reduction in configurable logic blocks - CLBs) and throughput (986$\times$ gain), enabling continuous learning directly on-chip.

\begin{table*}[!t]
\centering
\tiny
\setlength{\aboverulesep}{0pt} 
\setlength{\belowrulesep}{0pt} 
\setlength{\extrarowheight}{0pt}

\caption{Selection Guideline on HDC Encoding Methods with Hardware Cost and Taxonomy Symbols}
\vspace{-1.5em}
\renewcommand{\arraystretch}{0.6}
\resizebox{\linewidth}{!}{
\begin{tabular}{llll}
\toprule
\textbf{Encoding} & \textbf{Accuracy vs. Dimensionality} & \textbf{Hardware Cost} & \textbf{Best-Suited Domain}  \\
\midrule

\stfive~Record-based~\cite{Imani_Hierarchical,HDC_Arch,Abbas_Biosignal,ParhiSurvey}            
& Medium & Low & Multivariate sensor streams  
\\

\rowcolor{Lavender!30}
\stfive~N-gram~\cite{AbbasLang,ParhiSurvey,HDC_Arch}                  
& High (text) & Medium & Language and other symbol sequences    
\\

\stfour~RBF (cosine)~\cite{febhd,neuralhd,disthd,manihd}            
& High (continuous) & High & Regression or continuous-valued data   
\\

\rowcolor{Lavender!30}
\stfour~Random projection~\cite{RandomProj,AbbasEmbedding,BRIC,LocalityEN}       
& Medium & Medium--High & General-purpose vision tasks           
\\

\stthree~Cellular automaton~\cite{Menon2022,SchubertDescriptors}      
& Medium & Very Low & Spatio-temporal signal processing      
\\

\rowcolor{Lavender!30}
\stfour~Sigmoidal activation function~\cite{febhd,neuralhd} 
& Medium & Medium & Analog/continuous sensor data           
\\

\stfive~Circular convolution~\cite{377968,KleykoPART1}    
& Medium & Medium & Symbolic reasoning tasks 
\\

\rowcolor{Lavender!30}
\stsix~Stochastic computation~\cite{StocHD,StochasticHD}  
& Medium & Very Low & Resource-constrained (low-power) IoT   
\\

\stfour~Scatter code~\cite{kanerva1993sdm,KleykoPART1}            
& Medium & Low & Associative memory keys (addresses)    
\\

\rowcolor{Lavender!30}
\sttwo~Discrete Fourier Transform~\cite{voiceHD} 
& Medium & High & Frequency-domain signal analysis       
\\

\stfive~Kronecker product~\cite{daniels2023efficient}       
& Medium & High & Multi-modal feature combinations       
\\

\rowcolor{Lavender!30}
\stfour~Selective Kanerva coding~\cite{travnik2017representing,okko} 
& Medium & Low & Robotics and continuous control        
\\

\stfour~Randomized activation function~\cite{RBF,febhd} 
& Medium & Medium & General continuous features     
\\

\rowcolor{Lavender!30}
\sttwo~Feature quantization~\cite{QuantHD,HV_Design,10546832}    
& Medium & Low & Discretized (quantized) sensor data    
\\

\stfour~End-to-end projection~\cite{symbolichdc,gluing,10.1145/3400302.3415696}   
& High & High & Deep learning feature fusion            
\\

\bottomrule
\end{tabular}
}
\label{tab:hdc-encodings}
\end{table*}

Imani et al.~\cite{Imani_Hierarchical} introduced a multi-encoder HDC framework that adaptively switches between record-based and $N$-gram encoding based on data complexity. Their method reduces $\boldsymbol{\mathcal{HV}}$ dimensionality and lowers energy consumption and latency overheads, making it suitable for embedded deployment. Other works, such as those by Hersche et al.~\cite{AbbasEmbedding} and Cannings and Samworth~\cite{RandomProj}, employed sparse random projections for encoding, significantly reducing the number of arithmetic operations. Further optimizations, such as using fixed, pre-determined non-zero indices~\cite{BRIC, LocalityEN}, eliminate the need for random access logic during encoding, resulting in more deterministic and hardware-friendly designs. Since different encoding families exhibit different accuracy--dimensionality behavior, hardware cost, and domain suitability, Table~\ref{tab:hdc-encodings} provides a compact qualitative guideline for selecting HDC encoding.



\subsection{Encoding Strategies in Practice: Taxonomy-Based Literature Map, Examples, and Decision Framework}
\label{sec:encoding_practice}

Having introduced the multi-stage encoding taxonomy, we next connect the proposed stages to concrete HDC studies and practical encoding choices. Table~\ref{table_pop1} offers a holistic view of the HDC encoding landscape by categorizing the surveyed literature according to the proposed multi-stage taxonomy in Fig.~\ref{fig:pipeline_taxonomy}. For compactness, the table focuses on the algorithmic stages most directly tied to encoding: feature extraction, mapping, composition, and atomic \HV{} source generation.

Across these stages, a clear pattern emerges. Foundational HDC architectures typically bypass complex feature-extraction or mapping stages, relying primarily on high-dimensional pseudo-random atomic sources combined through standard binding (C1), bundling (C2), and permutation (C3). More recent application-specific frameworks, by contrast, increasingly integrate advanced feature extraction (F1--F5), nonlinear or learned mappings (M1--M9), or structured atomic \HV{} sources such as Gaussian, low-discrepancy, or phasor-based initialization. By condensing these operations into a structured stage-wise layout, Table~\ref{table_pop1} serves as a compact reference that allows researchers to quickly identify the end-to-end encoding pipeline and operational dependencies of SOTA HDC models.
\color{black}

\begin{table*}[!t]
\centering

\scriptsize
\renewcommand{\arraystretch}{1.2} 
\renewcommand{\tabularxcolumn}[1]{m{#1}}
\setlength{\aboverulesep}{0pt} 
\setlength{\belowrulesep}{0pt} 
\setlength{\extrarowheight}{0pt}

\caption{Representative Studies Categorized by the Multi-Stage HDC Encoding Taxonomy}
\label{table_pop1}
\vspace{-1.5em}
\resizebox{\linewidth}{!}{
\begin{tabularx}{\linewidth}{ >{\raggedright\arraybackslash}X c c c c }
\toprule
\multirow{3}{*}{\textbf{Representative Studies}} & \multicolumn{4}{c}{\textbf{Dominant Encoding Stage}} \\
\cmidrule(lr){2-5}
&\begin{tabular}[c]{@{}c@{}} \textbf{\sttwo}\\ \textbf{Feature}\end{tabular} 
&\begin{tabular}[c]{@{}c@{}} \textbf{\stfour}\\ \textbf{Mapping} \end{tabular} 
&\begin{tabular}[c]{@{}c@{}} \textbf{\stfive}\\ \textbf{Composition} \end{tabular} 
&\begin{tabular}[c]{@{}c@{}} \textbf{\stthree}\\ \textbf{\HV{} Source} \end{tabular} \\ 
\midrule

\rowcolor{Lavender!30}
\texttt{VoiceHD}~\cite{voiceHD}, \texttt{AdaptHD}~\cite{adapthd}, \texttt{PRID}~\cite{pridhd}, \texttt{MIMHD}~\cite{MIMHD}, \texttt{Dynamic-HDC}~\cite{Dynamic-HDC}, \texttt{HyperGRAF}~\cite{HyperGRAF}, \texttt{SemiHD}~\cite{SemiHD}, \texttt{BinHD}~\cite{BinHD}, \texttt{QubitHD}~\cite{QubitHD}, \texttt{QuantHD}~\cite{QuantHD}, \texttt{HDC-IM}~\cite{HDC-IM}, \texttt{EnHDC}~\cite{EnHDC}, \texttt{HDCluster}~\cite{HDCluster}, \texttt{TP-HDC}~\cite{lfsr_first_time}, \texttt{LeHDC}~\cite{LeHDC}, \texttt{XCelHD}~\cite{kang2022xcelhd}, \texttt{MulTa-HDC}~\cite{MulTa-HDC}, \texttt{HyperRec}~\cite{resSysHDC}, \texttt{ScaleHD}~\cite{ScaleHD}, \texttt{HyperSpec}~\cite{HyperSpec}, Imani et al.~\cite{imani-efficient}, Nazemi et al.~\cite{10.1145/3400302.3415696}, Montone et al.~\cite{montone2017hyperdimensional} 
& $\ast$ & $\ast$ & \textbf{C1, C2} & P. Random \\

\texttt{ST-HDC}~\cite{ST-HDC}, \texttt{RelHD}~\cite{RelHD}, \texttt{LookHD}~\cite{LookHD}, \texttt{SHDC}~\cite{shdc}, \texttt{DOMINO}~\cite{domino}, \texttt{GENERIC}~\cite{10.1145/3489517.3530669}, \texttt{BioHD}~\cite{BioHD}, \texttt{GrapHD}~\cite{cognitiveHDC}, \texttt{HAM}~\cite{HAM}, \texttt{PartialHD}~\cite{PartialHD}, Rahimi et al.~\cite{AbbasLang}, Datta et al.~\cite{HDC_Arch}, Li et al.~\cite{reramRahimi}, Wu et al.~\cite{FET}, Karunaratne et al.~\cite{Karunaratne2020}, Rahimi et al.~\cite{abbas-hdc-nano,10.1007/s11036-017-0942-6}, Kovalev et~al.~\cite{VQA} 
& $\ast$ & $\ast$ & \textbf{C1, C2, C3} & P. Random \\

\rowcolor{Lavender!30}
\texttt{FebHD}~\cite{febhd}, \texttt{DistHD}~\cite{disthd}, \texttt{DUAL}~\cite{Imani_Dual}, \texttt{TempHD}~\cite{TempHD}, \texttt{NeuralHD}~\cite{neuralhd}, \texttt{RE-FHDC}~\cite{re-hfdc}, \texttt{ManiHD}~\cite{manihd}, \texttt{HD-I-IoT}~\cite{HD-I-IoT}, \texttt{CascadeHD}~\cite{cascadehd}, \texttt{SupportHD}~\cite{supporthdc}, \texttt{CyberHD}~\cite{cyberhd}, Yu et al.~\cite{yu2022understanding} 
& $\ast$ & \textbf{M1} & $\ast$ & Gaussian \\

\texttt{SparseHD}~\cite{sparsehd}, \texttt{F5-HD}~\cite{F5-HD}, \texttt{compHD}~\cite{compHD}, \texttt{HD-Core}~\cite{HD-Core}, \texttt{HoloGN}~\cite{HoloGN}, Kleyko et al.~\cite{HDC_ind,Abbas_Class_recall} 
& $\ast$ & $\ast$ & \textbf{C2, C3} & P. Random \\

\rowcolor{Lavender!30}
\texttt{SecureHD}~\cite{Imani_cloud2019}, \texttt{OnlineHD}~\cite{onlinehd}, \texttt{FACH}~\cite{FACH}, \texttt{SHEARer}~\cite{SHEARer}, \texttt{SearcHD}~\cite{SearcHD} 
& $\ast$ & $\ast$ & \textbf{C1, C2} & Gaussian \\

\texttt{PULP-HD}~\cite{PULP-HD}, \texttt{MHD}~\cite{Imani_Hierarchical}, \texttt{PerfHD}~\cite{PerfHD}, Schmuck et al.~\cite{abbas-remat}, Rahimi et al.~\cite{Abbas_Biosignal,Abbas-ExG}, Moin et al.~\cite{moin2018emg} 
& $\ast$ & $\ast$ & \textbf{C1, C2\textsuperscript{\dag}, C3\textsuperscript{\dag}} & P. Random \\

\rowcolor{Lavender!30}
\texttt{ReHD}~\cite{LocalityEN}, \texttt{BRIC}~\cite{BRIC}, \texttt{HyperSpike}~\cite{HyperSpike}, \texttt{HyperSpikeASIC}~\cite{HyperSpikeASIC}, \texttt{HyDREA}~\cite{10.1145/3524067}, \texttt{FHDnn}~\cite{10.1145/3489517.3530394}, \texttt{FedHD}~\cite{10.1145/3495243.3558757}, \texttt{Prive-HD}~\cite{Prive-HD}, \texttt{THRIFTY}~\cite{THRIFTY}, \texttt{HDnn-PIM}~\cite{hdnn}, Gupta et al.~\cite{10.1145/3503541}, Morris et al.~\cite{multi_label} 
& $\ast$ & \textbf{M8} & $\ast$ & P. Random \\

\texttt{Laelaps}~\cite{sandwich}, Burrello et al.~\cite{abbas-ieeg} 
& \textbf{F2} & $\ast$ & \textbf{C1} & $\ast$ \\

\rowcolor{Lavender!30}
\texttt{GenieHD}~\cite{geniehd}, \texttt{HyperEmbed}~\cite{HyperEmbed}, Mitrokhin et al.~\cite{sensorimotor}, Rahimi et al.~\cite{rahimi3D} 
& $\ast$ & $\ast$ & \textbf{C1, C3} & P. Random \\

Menon et al.~\cite{CA-HV,Menon2022} 
& $\ast$ & \textbf{M3} & $\ast$ & P. Random \\

\rowcolor{Lavender!30}
\texttt{HDNA}~\cite{DNAimani} 
& $\ast$ & \textbf{M3} & \textbf{C1, C2\textsuperscript{\dag}, C3\textsuperscript{\dag}} & P. Random \\

Rosato et al.~\cite{9533805} 
& \textbf{F3} & $\ast$ & \textbf{C1, C2} & $\ast$ \\

\rowcolor{Lavender!30}
\texttt{QHD}~\cite{QHD} 
& $\ast$ & $\ast$ & \textbf{C1, C2} & \begin{tabular}[c]{@{}c@{}} Random Exp.\\ ($e^{i\theta}$) \end{tabular} \\

Mitrokhin et al.~\cite{symbolichdc} 
& \textbf{F4} & $\ast$ & \textbf{C1} & $\ast$ \\

\rowcolor{Lavender!30}
Burrello et al.~\cite{8584751} 
& \textbf{F2} & $\ast$ & \textbf{C1, C2} & $\ast$ \\

\texttt{tiny-HD}~\cite{BehnamIoT} 
& $\ast$ & \textbf{M2} & \textbf{C3} & P. Random \\

\rowcolor{Lavender!30}
Basaklar et al.~\cite{HV_Design} 
& $\ast$ & $\ast$ & \textbf{C2} & P. Random \\

Hersche et al.~\cite{AbbasEmbedding} 
& \textbf{F1} & \textbf{M2, M8} & $\ast$ & Gaussian \\

\rowcolor{Lavender!30}
Sutor et al.~\cite{gluing}, Daniels et al.~\cite{daniels2023efficient} 
& \textbf{F4} & $\ast$ & \textbf{C2, C5} & $\ast$ \\

Aygun et al.~\cite{HDC_Aygun_DATE2023} 
& $\ast$ & $\ast$ & \textbf{C1, C3} & LD Seq. \\

\rowcolor{Lavender!30}
\texttt{uHD}~\cite{uHD}, Moghadam et al.~\cite{Mehran-No-mul} 
& $\ast$ & $\ast$ & \textbf{C2} & LD Seq. \\

Chang et al.~\cite{8771622} 
& \textbf{F5} & $\ast$ & \textbf{C1, C2} & $\ast$ \\

\rowcolor{Lavender!30}
\texttt{HDCP}~\cite{HDCP} 
& $\ast$ & \textbf{M9} & $\ast$ & P. Random \\

\texttt{StocHD}~\cite{StocHD}, \texttt{Stochastic-HD}~\cite{StochasticHD} 
& $\ast$ & \textbf{M2, M9} & $\ast$ & P. Random \\

\rowcolor{Lavender!30}
Rasanen~\cite{okko} 
& \textbf{F5} & \textbf{M2} & $\ast$ & $\ast$ \\

Kirilenko et al.~\cite{10.1007/978-3-030-89820-5_3} 
& $\ast$ & $\ast$ & \textbf{C4} & P. Random \\

\rowcolor{Lavender!30}
Xu et al.~\cite{FSL-HD} 
& $\ast$ & \textbf{M7} & $\ast$ & P. Random \\

Hersche et al.~\cite{10.1145/3370748.3406560} 
& \textbf{F1} & $\ast$ & $\ast$ & Gaussian \\

\rowcolor{Lavender!30}
Heddes et al.~\cite{10.1145/3489517.3530553} 
& $\ast$ & $\ast$ & \textbf{C1} & P. Random \\

\texttt{LogHD}~\cite{LogHD_DATE2026}, \texttt{DystoHD}~\cite{DystoHD} 
& \textbf{F1} & \textbf{M1} & \textbf{C2} & P. Random \\

\rowcolor{Lavender!30}
\texttt{DecoHD}~\cite{DecoHD_DATE2026} 
& \textbf{F2, F3} & \textbf{M2, M3} & \textbf{C2} & Gaussian \\

Rajappa et al.~\cite{Radiation_HDC_ACM_TECS2026} 
& \textbf{F1} & \textbf{M1, M3} & \textbf{C2} & P. Random \\

\bottomrule
\multicolumn{5}{p{\textwidth}}{\tiny
\textbf{Stage-2 (F1-F5):} 
\textbf{F1}: Feature Quantization (Discretization); 
\textbf{F2}: LBP; 
\textbf{F3}: Thermometer-based Level Feature Encoding; 
\textbf{F4}: NN Output or DeepHash; 
\textbf{F5}: Non-linear or Domain-Specific Feature Transform. \newline
\textbf{Stage-4 (M1-M9):} 
\textbf{M1}: Radial Basis Function or Cosine Mapping; 
\textbf{M2}: Random Projection; 
\textbf{M3}: Cellular-Automata, Structured Mapping; 
\textbf{M4}: Sigmoidal Activation; 
\textbf{M5}: Scatter Code; 
\textbf{M6}: Fourier, DFT-based Mapping; 
\textbf{M7}: Selective Kanerva Coding, Randomized Activation; 
\textbf{M8}: End-to-End Learned Projection; 
\textbf{M9}: Stochastic Bitstream Mapping. \newline
\textbf{Stage-5 (C1-C5):} 
\textbf{C1}: Binding; 
\textbf{C2}: Bundling; 
\textbf{C3}: Permutation; 
\textbf{C4}: Circular Convolution; 
\textbf{C5}: Kronecker Product. \newline
\dag~Features a two-stage encoding approach. \quad 
$\ast$ Indicates that the module is bypassed or intrinsically handled by another stage. \quad 
P. Random: Pseudo-Random. \quad
The F/M/C tags indicate the dominant role of each method in the cited implementation; some methods may span multiple stages depending on whether a transform is used as preprocessing, mapping, or \HV{} source generation.
} \\
\end{tabularx}
}
\vspace{-1.5em}
\end{table*}

\color{black}

\subsubsection{Use-Case Examples based on Encoding Taxonomy. 
} 

Building on this taxonomy-based literature map, the following examples illustrate how these encoding choices translate into concrete design decisions for representative application scenarios.

\begin{itemize}[leftmargin=*, label=\ding{226}]
  \item \textbf{Record-based~\stfive~(Eq.~\ref{eq1}).} \emph{Wearable-sensor activity recognition:} Each quantized accelerometer sample is first mapped to its designated level $\boldsymbol{\mathcal{HV}}$ and then bound (\texttt{XOR}) with a unique time-step $\boldsymbol{\mathcal{HV}}$ that encodes temporal position. Superposing these bound vectors over all channels and sampling instants produces a single majority-voted $\boldsymbol{\mathcal{HV}}$ that compactly represents the entire activity sequence while inherently attenuating sensor noise~\cite{Imani_Hierarchical,Abbas_Biosignal,HDC_Arch,ParhiSurvey}.
  \item \textbf{N-gram~\stfive~(Eq.~\ref{eq2}).} \emph{Text language identification:} Within the Europarl dataset~\cite{Europarl}, each alphabetic symbol is first assigned a random seed $\boldsymbol{\mathcal{HV}}$. A cyclic-shift operator encodes positional information for every contiguous four-character (4-gram) window, after which the four position-encoded vectors are bound component-wise via \texttt{XOR} and accumulated into a class $\boldsymbol{\mathcal{HV}}$. This permutation-binding scheme preserves short-range character dependencies and achieves approximately 97\% language-identification accuracy when $D=10{,}000$~\cite{Imani_Hierarchical,Abbas_Biosignal,ParhiSurvey}. 

  \item \textbf{Scatter code~\stfour.} \emph{Associative memory indexing:} Scatter-coding generates distributed $\boldsymbol{\mathcal{HV}}$s as keys for storing and retrieving items in a distributed memory. For instance, in a cognitive computing database, each record’s index or key can be represented by a scatter-coded $\boldsymbol{\mathcal{HV}}$, ensuring that different keys map to well-separated points in the space. This property has been leveraged in HDC-based associative memories and hash tables to reliably look up stored data via partial matches, as the quasi-orthonormal addresses (keys) minimize interference between entries~\cite{thermometer, KleykoPART1, 10.1145/3489517.3530553}.


  \item \textbf{Sigmoidal activation function~\stfour.} \emph{Analog sensor anomaly detection:} Real-valued sensor readings can be encoded by feeding the feature vector through a set of randomly weighted summation units with a sigmoidal (e.g., logistic) activation. Each unit outputs 1 or 0 depending on whether the projection of the input exceeds its threshold, and the collection of these outputs forms the binary $\boldsymbol{\mathcal{HV}}$. This non-linear encoding (pioneered by early work on random threshold units) captures gradual variations in analog inputs with moderate accuracy and complexity, suitable for tasks like detecting anomalies in environmental sensor data~\cite{RBF,rbf2}.  

  \item \textbf{Random projection~\stfour.} \emph{CIFAR-10 image classification:} The raw pixel intensities of the images are linearly projected through a dense bipolar Gaussian matrix to yield image $\boldsymbol{\mathcal{HV}}$. Empirical analysis suggests that a dimensionality of $D\approx2{,}000-5{,}000$ effectively preserves inter-image similarity and classification accuracy, while keeping memory and computational overhead at a manageable level~\cite{RandomProj,AbbasEmbedding}.

  \item \textbf{Randomized activation function~\stfour.} \emph{Generic continuous-feature encoding:} A randomized activation encoder maps real-valued features into a $\boldsymbol{\mathcal{HV}}$ by applying a set of random nonlinear transformations. For example, in a gesture recognition application, a feature vector extracted from accelerometer signals can be fed into a layer of randomly initialized neurons (with random weights and biases) that output $+1$ if activated and $-1$ otherwise. The concatenation of all neuron outputs forms the $\boldsymbol{\mathcal{HV}}$ for that gesture sample. This approach, essentially using random threshold units for encoding, requires only moderate computation and no training, and is broadly applicable to continuous input data 
  where a quick, one-shot encoding is desired~\cite{travnik2017representing, 8771622}.

  \item \textbf{RBF-style (cosine)~\stfour.} \emph{Continuous-value regression (e.g., gas sensors):} Each continuous feature vector $F$ is transformed via $h_i=\operatorname{sgn}\bigl[\cos\bigl(B_i^{\top}F\bigr)\bigr]$, where every projection vector $B_i$ is sampled from $\mathcal{N}(0,1)$. It enables high-accuracy regression or function approximation in the downstream HDC pipeline by treating distance in feature space as Hamming distance in $\boldsymbol{\mathcal{HV}}$ space~\cite{cyberhd,neuralhd,RBF,rbf2,febhd,disthd,re-hfdc,manihd}.

  \item \textbf{Selective Kanerva coding~\stfour.} \emph{Robotics and continuous control:} Selective Kanerva coding encodes a continuous state by activating a sparse subset of prototype $\boldsymbol{\mathcal{HV}}$s from a large predefined pool (inspired by Kanerva’s sparse distributed memory). In robotic control, 
  each continuous state variable (like an angle or velocity) is first quantized coarsely, then represented by a few stored $\boldsymbol{\mathcal{HV}}$s ``covering'' the corresponding 
  bin or nearby values. These active $\boldsymbol{\mathcal{HV}}$s (often called landmarks) are bundled to produce the final state encoding $\boldsymbol{\mathcal{HV}}$. By varying the subset of prototypes smoothly as the state changes, this method yields representations that reflect continuous differences, using only simple table lookups and superposition—ideal for tasks like robot navigation or sensor fusion, where state variables are continuous but require a compact, noise-tolerant encoding~\cite{travnik2017representing}.

  \item \textbf{Circular convolution~\stfive.} \emph{Symbolic binding and reasoning:} Circular convolution (the binding operation in certain HDC models) is often used to encode structured data such as sequences or paired associations. As an example, to represent a semantic relation ``(entity$_1$ — relation — entity$_2$)'' in a knowledge base, one can bind the $\boldsymbol{\mathcal{HV}}$s for entity$_1$ and entity$_2$ with role $\boldsymbol{\mathcal{HV}}$s via circular convolution, yielding a single composite $\boldsymbol{\mathcal{HV}}$. Such encodings have been used in tasks like analogical reasoning and symbolic inference, where preserving the relational structure in the $\boldsymbol{\mathcal{HV}}$ is crucial. By using convolution-based encoding, these models achieve moderate accuracy on relational benchmarks while remaining compatible with fast Fourier transform (FFT) accelerations for efficient binding~\cite{9892462,Abbas_Class_recall}.

  \item \textbf{Discrete Fourier transform (DFT)~\sttwo.} \emph{Frequency-domain signal encoding:} Many signals are naturally analyzed in the frequency domain. For example, an EEG signal segment can be encoded by first computing its DFT and then mapping each frequency bin’s complex coefficient to a $\boldsymbol{\mathcal{HV}}$ representation (e.g., using phase to determine binary components). The resulting $\boldsymbol{\mathcal{HV}}$ captures the spectral signature of the signal; binding operations in the hyperdomain correspond to convolution in the original time domain, which aligns well with frequency-based feature integration. DFT-based $\boldsymbol{\mathcal{HV}}$s have been explored in audio and vibration analysis tasks, leveraging frequency content for improved classification performance in domains like speech recognition~\cite{9892462, voiceHD}.

  \item \textbf{Kronecker product~\stfive.} \emph{Multi-modal feature fusion:} The Kronecker (tensor) product can combine multiple modality-specific $\boldsymbol{\mathcal{HV}}$s into one joint representation. For instance, consider a system that processes both video and audio streams: a $\boldsymbol{\mathcal{HV}}$ from visual features (frame $\boldsymbol{\mathcal{HV}}$) and another from audio features (sound $\boldsymbol{\mathcal{HV}}$) can be tensored together to form a single $\boldsymbol{\mathcal{HV}}$ that encodes all cross-modal combinations. This fused representation enables the HDC model to capture interactions between modalities (e.g., lip movements and spoken words in an emotion recognition task) with a single high-dimensional vector. While the dimensionality grows multiplicatively (and thus can be very large), the approach has shown success in tasks where multi-modal correlations are key to accuracy~\cite{daniels2023efficient, RelHD}.

  \item \textbf{Feature quantization~\sttwo.} \emph{Quantized sensor signal encoding for wearable health monitoring:} Continuous-valued input features are discretized into fixed bins, with each bin assigned a unique $\boldsymbol{\mathcal{HV}}$ that preserves similarity across neighboring values. For example, in wearable IoT devices monitoring physiological signals, each quantized level of a sensor reading (e.g., temperature or heart rate) is mapped to a bipolar $\boldsymbol{\mathcal{HV}}$, and subsequent levels are generated by flipping a fixed number of bits from the prior one~\cite{AMS-HD_TBME2026}. These level $\boldsymbol{\mathcal{HV}}$s are then superposed to represent the full input sample. This strategy, demonstrated by Basaklar et al.~\cite{HV_Design}, significantly reduces dimensionality (by up to $30\times$) while maintaining competitive accuracy. Although discretization introduces quantization error, careful selection of bin granularity and $\boldsymbol{\mathcal{HV}}$ flipping rates balances performance and resource efficiency, making this method ideal for low-power, real-time health monitoring systems~\cite{AMS-HD_TBME2026, RHDC}.

  \item \textbf{Cellular automaton encoding~\stthree~\stfour.} \emph{Spatio-temporal pattern recognition:} An elementary cellular automaton (CA) can serve as a lightweight encoder by evolving an input pattern over time. For example, one can initialize a 1-D CA with a binary pattern derived from sensor readings (or a 2-D CA with an input image) and let it iterate for a fixed number of steps; the automaton’s state (cells) at the final step is then interpreted as the encoded $\boldsymbol{\mathcal{HV}}$. This process captures local spatio-temporal correlations in the data through the CA’s dynamics, and has been used for tasks like simple gesture or anomaly detection with extremely low hardware cost (CA updates involve only bit-wise operations)~\cite{Menon2022,Abbas_Class_recall,HoloGN}.

  \item \textbf{Stochastic computation~\stthree \stfour \stsix.} \emph{Ultra-low-power IoT classification:} In stochastic $\boldsymbol{\mathcal{HV}}$ encoders, analog values are first converted into stochastic bit-streams (bit sequences where the probability of 1s encodes the magnitude). These bit-streams are then combined using stochastic logic (e.g., bit-wise XNOR for binding and bit-wise OR for bundling) to produce the resulting $\boldsymbol{\mathcal{HV}}$. This approach drastically reduces hardware complexity and power consumption—making it attractive for IoT sensors that must classify events like ECG anomalies on-device at the expense of longer encoding times (since accuracy improves with longer bit-streams similar to $\boldsymbol{\mathcal{HV}}$s)~\cite{8049760,Alaghi_SCC1, ID-VSA_Moghadam_TVLSI2026}.

  \item \textbf{End-to-end projection~\stfour.} \emph{Learned $\boldsymbol{\mathcal{HV}}$ embeddings:} In applications requiring 
  high accuracy, one can learn the encoding function itself using an end-to-end model (often an NN) that outputs $\boldsymbol{\mathcal{HV}}$s optimized for the task. For example, a small NN might be trained to produce a $10{,}000$-dimensional binary $\boldsymbol{\mathcal{HV}}$ for each input image such that these $\boldsymbol{\mathcal{HV}}$s are easily separable for different classes. This end-to-end trained encoding, used in conjunction with simple HDC classifiers, has achieved accuracies close to DNNs on image and speech tasks. The trade-off is significantly higher encoding cost and complexity: the model needs offline training and typically requires more computation (or dedicated hardware) during inference. Nonetheless, for complex data modalities where manual encoding falls short, learned projection offers a path to fuse deep learning feature extraction with HDC’s efficient associative search~\cite{AbbasEmbedding,WANG2022125042}.


\end{itemize}

\subsubsection{Selection Criteria}
The taxonomy and use-case examples above can be translated into practical design guidelines for selecting an HDC encoding strategy. These criteria should be interpreted as non-exclusive design heuristics rather than rigid rules, since a practical HDC system may combine multiple stages depending on the input modality, target accuracy, dimensionality budget, and hardware constraints.

\begin{itemize}[leftmargin=*, label=\ding{114}]

 \item \textbf{Categorical symbols or tokens\blue{~\stone~\stfive}:} 
 Prefer simple record-based encoding for unordered sets of symbols, or $N$-gram encoding for symbol sequences because permutation and binding preserve local order and contextual dependencies, using orthogonal seed $\boldsymbol{\mathcal{HV}}$s for individual symbols in both cases.

\item \textbf{Continuous numeric data~\sttwo~\stfour:} 
If preserving numeric similarity is important, adopt a ``value-level'' encoding such as thermometer codes or RBF-based encoding, which map similar magnitudes to similar $\boldsymbol{\mathcal{HV}}$s. These encodings maintain metric relationships in the $\boldsymbol{\mathcal{HV}}$ space, aiding regression, anomaly detection, and continuous sensing tasks.

\item \textbf{Composite or structured data~\stone~\stfive:} 
For data with multiple modalities or inherent structure (e.g., multi-sensor inputs, graphs, role-filler pairs, or sequences with relationships), use binding operations, circular convolution, or tensor products to combine feature $\boldsymbol{\mathcal{HV}}$s. Such encoders help preserve relational information, including positional binding for sequences, at the cost of higher dimensionality or hardware complexity, and are recommended when relational information is crucial to the application.

\item \textbf{Resource-constrained hardware~\stsix~\stthree~\stfour:} 
In low-power or memory-limited settings (such as wearable devices or tiny IoT nodes), choose encoders that minimize computation and storage. Methods like CA encoding, stochastic or bit-stream mapping, sparse projections, or using low-dimensional scatter codes can substantially reduce hardware cost, trading off some accuracy for efficiency. Also consider reducing $D$ and using optimized base vectors (e.g., LD sequences) when reproducible initialization or improved low-dimensional behavior is beneficial for the target workload.

\item \textbf{High accuracy requirements~\stfour~\stfive:} 
When an application demands high accuracy and sufficient resources are available, employ more expressive encodings. For text or DNA sequences, $N$-gram encoding with large $D$ excels; for continuous data, RBF or even learned end-to-end encoders can significantly boost accuracy. These approaches come with higher complexity, so they are best reserved for cases where the gain justifies the cost.

\item \textbf{Tight dimensionality budgets~\stthree~\stfour:} 
If the chosen $\boldsymbol{\mathcal{HV}}$ dimensionality is very limited (e.g., $D \ll 10^3$ due to storage or bandwidth constraints), counteract the reduction in capacity by using structured $\boldsymbol{\mathcal{HV}}$ generation schemes. For instance, LD sequences or sparse $\boldsymbol{\mathcal{HV}}$s may help preserve separability at lower $D$, as conventional purely random $\boldsymbol{\mathcal{HV}}$s may collide in such low dimensions. Under severe dimensional limits, simpler encodings may also outperform complex ones, since complex encoders might overfit or produce correlated components when $D$ is small.

\end{itemize}


\subsection{From HDC Encoding to Training}
\label{section_encoding_to_training}
Once input data are encoded into $\boldsymbol{\mathcal{HV}}$s, 
the next critical step is training. 
Unlike traditional ML pipelines that rely on gradient-based optimization or iterative backpropagation, HDC leverages simple yet powerful operations, such as bundling and thresholding, to form class representations (recall Fig.~\ref{VSA_general} and Fig.~\ref{fig:hdc_vs_nn}). The goal is to construct a robust class $\boldsymbol{\mathcal{HV}}$ that captures the statistical patterns of all training samples belonging to the same category. The most common HDC training method is prototype-based learning, where all encoded $\boldsymbol{\mathcal{HV}}$s from a class are aggregated via element-wise addition (or bundling) to form a single prototype vector, often referred to as the class $\boldsymbol{\mathcal{HV}}$ ($C_{class}$). This process is sometimes normalized or thresholded to produce a binary or bipolar final representation. HDC training is highly parallelizable and memory efficient, enabling ultra-fast training even on hardware-constrained platforms~\cite{ScaleHD, adapthd}. Training can be performed in either batch or online mode. In batch training, all samples are encoded and then bundled per class before inference begins. In contrast, online HDC training updates class $\boldsymbol{\mathcal{HV}}$s incrementally as each new sample arrives. This is especially effective in real-time or streaming applications where memory and latency constraints are critical~\cite{onlinehd, adapthd}. Furthermore, strategies like weighted bundling and temporal decay can be applied to control the influence of older samples~\cite{HD-I-IoT}.

Epoch-based learning can also be applied by adopting a learning rate ($\eta$)-based model update for incremental learning. The model undergoes validation accuracy evaluation during each training sample's process, considering the impact of the final contributing \textit{training sample $\boldsymbol{\mathcal{HV}}$} ($h$). If validation improves based on the recent contribution to the class $\boldsymbol{\mathcal{HV}}$ ($C$), then the training sample's effect on the learning model is incorporated, contributing to the class $\boldsymbol{\mathcal{HV}}$ via accumulation. The formula for updating the class $\boldsymbol{\mathcal{HV}}$ in the event of validation accuracy improvement is $C_{new}$ = $C_{old} + (\eta \times h)$ (otherwise, $C_{new}$ = $C_{old} - (\eta \times h)$)~\cite{Sobol_Aygun_TCAD}.

To enhance generalization, recent works also introduce pre-training techniques using synthetic or auxiliary data to initialize the class $\boldsymbol{\mathcal{HV}}$s, which are then fine-tuned with application-specific data. Some studies apply transfer learning by adapting learned $\boldsymbol{\mathcal{HV}}$s across tasks, especially when underlying representations remain invariant, such as sensor modality or character structure~\cite{resSysHDC, AbbasLang}. In multi-class problems, class discrimination is improved by refining the $\boldsymbol{\mathcal{HV}}$s using associative memory updates or subtractive mechanisms, where negative samples are explicitly used to ``unlearn'' conflicting features from class vectors~\cite{HyperGRAF, Prive-HD}. This approach enables better decision boundaries when the classes share structural similarities in their input encoding.

Finally, the hardware-friendly nature of HDC during training is a major advantage. Since training primarily involves vector addition and simple element-wise operations, it can be efficiently implemented using IMC or low-complexity logic blocks. Several architectures, including \texttt{HDNA}~\cite{DNAimani}, \texttt{PerfHD}~\cite{PerfHD}, and \texttt{HyperSpikeASIC}~\cite{HyperSpikeASIC} leverage this property to deliver energy-efficient training directly on edge or embedded devices.

\color{black}

\vspace{-1em}
\section{Future Encoding Promises and Recommendations}
\label{challenges}
In this section, we discuss the future prospects of HDC with a focus on the vector encoding step as the central theme of this survey. The following discussion intentionally restricts future directions to issues that directly affect the proposed encoding taxonomy. HDC adoption in emerging platforms is most relevant to this survey when the platform changes how \HV{s} are generated, stored, mapped, or composed.

Significant research direction with the potential to yield fruitful results is the exploration of the analogy between HDC systems and NNs, particularly regarding encoding strategies based on NN outputs. Ma and Jiao~\cite{ma2022hyperdimensional} have provided valuable insights into this comparison by discussing the two learning strategies in a comparative manner. Notably, they highlight the effectiveness of NN-assisted HDC, which suggests that the generation of $\boldsymbol{\mathcal{HV}}$s can be effectively managed using NN approaches. Researchers could explore replacing custom distributions with lightweight AI methods for efficient $\boldsymbol{\mathcal{HV}}$ generation via NNs. Duan et al.~\cite{LeHDC} have previously proclaimed HDC as a single-layer binarized neural network (BNN) architecture. They identify two key limitations in current HDC training: (i) retraining updates only the misclassified class $\boldsymbol{\mathcal{HV}}$, unlike in conventional NNs, leaving correct ones unchanged, and (ii) retraining uses a fixed step size. They note that the derivative of loss, used in conventional NNs, is not considered in HDC training. Thus, NN-assisted encoders and learned projections should be interpreted as Stage-2 feature~\sttwo~and Stage-4 mapping~\stfour~design choices that improve \HV{} separability, rather than as a general replacement for the HDC pipeline. Recent works have also explored encoding synergy between HDC and ML by using neural encoders for $\boldsymbol{\mathcal{HV}}$ generation~\cite{AyarGLSVLSI24llm} or embedding HDC modules into deep learning pipelines~\cite{austin_ESL}.

Cross-paradigm computing should also be viewed through the lens of encoding rather than as a separate application direction. Among emerging computational paradigms, HDC can be seen as a complementary approach to SC. SC enables fault-tolerant computation through probabilistic bit-streams and has proven particularly effective in hardware-efficient designs. HDC, on the other hand, introduces high-dimensional representations and operations such as bundling and binding, enabling robust, symbolic, and distributed representations. SC is most relevant here when stochastic or unary bit-streams serve as Stage-3~\stthree~or Stage-4 mapping~\stfour~mechanisms; or Stage-6 hardware realizations~\stsix~for efficient encoder implementation owing to low-power SC.


In the broader context of emerging paradigms, HDC aligns well with IMC, particularly considering the inherent robustness. Karunaratne et al.~\cite{Karunaratne2020} explore the implementation of a full HDC system on an IMC platform. They explore an \textit{N-gram-based} encoding approach and similarity calculation through associative memory search, which operates exclusively in memory. On the other hand, Liu et al. propose an adapted method for Resistive Random-Access Memory (RRAM) devices for IMC in the \texttt{HDC-IM} framework~\cite{HDC-IM}. In this framework, they propose memory-based steps, such as bit-flipping, replication, shifting, and catenation, to generate item memory and continuous item memory, thereby offering a novel approach for handling the vector generation process in memory. Future efforts in in-memory $\boldsymbol{\mathcal{HV}}$ generation must consider the cost of steps like shifting and concatenation. It is also essential to measure the degree of orthogonality numerically after these steps, ensuring $\boldsymbol{\mathcal{HV}}$ quality.

Another promising avenue for future HDC research lies in custom processor design. Datta et al.~\cite{HDC_Arch} have taken the first step by proposing a $28nm$ Application-Specific Integrated Circuit (ASIC) processor architecture. This generic \texttt{HD} processor integrates multichannel sensor inputs and an $\boldsymbol{\mathcal{HV}}$ mapper, which facilitates vector generation into item and associative memory. The encoder executes essential learning operations. From the perspective of this survey, future processor designs should be evaluated primarily by how efficiently they support Stage-3 atomic \HV{} generation~\stthree, Stage-4 mapping~\stfour, and Stage-5 HDC composition operations~\stfive. Particularly, for lightweight wearable sensing of biomedical data and classification tasks, a custom processor optimized for HDC could prove highly valuable. HDC has already proven effective in various biomedical applications~\cite{sandwich,Abbas_Biosignal,Menon2022,CA-HV,AMS-HD_TBME2026}.

Representative IMC-oriented HDC works show that item-memory generation, bit-flipping, shifting, replication, and associative search can be implemented close to or inside memory~\cite{Karunaratne2020,HDC-IM,hdnn,ST-HDC}. A key challenge remains in generating high-quality $\boldsymbol{\mathcal{HV}}$s and integrating them into hardware while addressing orthogonality, correlation, and reproducibility issues. In parallel with these, the security and privacy dimensions of HDC are gaining increasing attention. As HDC finds applications in safety-critical domains such as healthcare, robotics, and edge-AI, ensuring model robustness against hardware-induced faults and adversarial manipulation becomes essential. Recent studies have explored the vulnerability of HDC models to targeted bit-flip attacks and proposed frameworks for attack simulation and defense. For instance, Zhang et al.~\cite{10908571} demonstrated how minimal, targeted memory errors can drastically degrade classification accuracy, and introduced mitigation techniques such as HD-BFS and HD-ABFS to enhance system reliability. Liu et al.~\cite{10247811} presented \texttt{HyperAttack}, a structured attack framework for evaluating HDC's resilience. 
On the privacy front, Khaleghi et al.~\cite{10720539} and Yu et al.~\cite{lightningtalk} proposed privacy-preserving HDC models for secure edge inference using homomorphic encryption and multi-party computation. These directions are directly tied to encoding as the choice of atomic \HV{} source, mapping mechanism, binarization rule, and memory layout can affect both robustness and privacy. Future secure HDC systems should therefore evaluate not only classifier accuracy but also how Stage-3 source~\stthree~and Stage-4 mapping~\stfour~choices affect the attack surface.

As we approach the conclusion of this survey, it is crucial to bring attention to a significant challenge that has been overlooked in the literature: the dynamic generation of $\boldsymbol{\mathcal{HV}}$s during the encoding procedure. Most HDC studies generate $\boldsymbol{\mathcal{HV}}$s by either using a seed vector with a bit-flip technique~\cite{Zou2022} or retrieving randomly generated orthogonal vectors from memory units, such as look-up tables~\cite{LookHD}. In online learning scenarios, where new data arrives continuously, a new vector set is required for each data instance. The dynamic generation of these new vector sequences becomes a key concern. This challenge directly affects Stage-3 source generation~\stthree~and Stage-6 hardware realization~\stsix, since the encoder must generate reproducible, sufficiently orthogonal, and hardware-efficient \HV{s} without excessive memory lookup overhead.

A recent study by Thomas et al.~\cite{thomas2023streaming} addresses the \textit{dynamicity} issue with an innovative hashing solution~\cite{10.1145/3489517.3530553} for on-the-fly vector generation, also known as \textit{streaming encoding}. SC, with its hardware-efficient nature, offers valuable potential for streaming encoding, particularly when exploring cross-paradigm computing solutions. In SC, the hardware structure for bit-stream generation utilizes a simple generator comprising an LFSR and a comparator~\cite{8049760}. This enables the production of correlated or uncorrelated sequences using the same or different LFSR modules, respectively, with ease~\cite{Alaghi_SCC1}. It is worth noting that Chang et al.~\cite{lfsr_first_time} have touched upon the possible utilization of a hardware-friendly LFSR approach in HDC systems. In summary, the dynamic generation of $\boldsymbol{\mathcal{HV}}$s during encoding presents a critical challenge for HDC systems, especially in scenarios involving online learning. 


To sum up, we recommend the following future efforts as new research opportunities in the frame of encoding-centric HDC: 
\ding{182} streaming encoding and dynamic \HV{} source generation for edge devices, directly targeting Stage-3 source generation~\stthree~and Stage-6 hardware realization~\stsix; 
\ding{183} encoding-aware biomedical HDC frameworks, where biosignal modality, feature extraction, and mapping choices should be jointly studied through Stage-1~\stone, Stage-2~\sttwo, and Stage-4~\stfour; \ding{184} new research efforts on the security and reliability side of HDC systems that evaluate how \HV{} source generation, mapping, binarization, and memory layout affect robustness and privacy, especially across Stage-3~\stthree, Stage-4~\stfour, and Stage-6~\stsix; \ding{185} cross-paradigm encoder designs, including SC, neuromorphic, and IMC-based approaches, only when they directly improve \HV{} generation, mapping, composition, or hardware realization; and \ding{186} new HDC efforts for video and event-stream processing that explicitly address spatio-temporal feature extraction, sequence composition, and encoder efficiency through Stage-1~\stone, Stage-2~\sttwo, Stage-5~\stfive, and Stage-6~\stsix.



\begin{table*}[!t]
\centering
\color{black}
\scriptsize
\caption{Representative HDC Frameworks and Implementations Mapped to the Encoding Taxonomy}
\label{hdc_imp}
\vspace{-1.2em}
\renewcommand{\arraystretch}{0.68}
\setlength{\tabcolsep}{2.5pt}
\setlength{\aboverulesep}{0pt} 
\setlength{\belowrulesep}{0pt} 
\setlength{\extrarowheight}{0pt}
\resizebox{\linewidth}{!}{
\begin{tabular}{@{} 
>{\raggedright\arraybackslash}p{0.31\linewidth}
>{\centering\arraybackslash}p{0.15\linewidth}
>{\raggedright\arraybackslash}p{0.50\linewidth}
@{}} 
\toprule
\textbf{HDC Projects} 
& 
\textbf{Relevant Taxonomy Stages}
& 
\textbf{Encoding-Relevant Details (\HV{} source, mapping, composition, or encoder realization)} \\ 
\midrule

\rowcolor{Lavender!30}
\textbf{\texttt{TorchHD}}~\cite{torchHD,heddesTorch}
& \sttwo~\stthree~\stfour~\stfive
& General HDC framework supporting random, level, thermometer, circular, RBF-kernel, \textit{N-gram-based}, and \textit{Record-based} encoding. \\ 

\textbf{\texttt{DistHD}}~\cite{disthd}
& \stthree~\stfour
& Gaussian \HV{} source and RBF-style mapping for distance-preserving continuous-feature encoding. \\ 

\rowcolor{Lavender!30}
\textbf{\texttt{OpenHD}}~\cite{openhd}
& \stfour~\stfive~\stsix
& GPU-oriented framework support for mapping, composition, and encoder execution. \\ 

\textbf{\texttt{HDTorch}}~\cite{hdTorch}
& \stthree~\stfour
& Random, sandwich, and scale-with-radius encoding for HDC model construction. \\ 

\rowcolor{Lavender!30}
\textbf{\texttt{Hyperdimensional Computing Library}}~\cite{HDlib}
& \stthree~\stfive~\stsix
& Random \HV{} source and \textit{N-gram-based} composition with CPU/GPU-oriented encoder support. \\

\textbf{\texttt{Laelaps}}~\cite{sandwich}
& \stone~\sttwo~\stfive~\stsix
& LBP-based feature extraction with composition-oriented encoder realization~\cite{kaya20141d}. \\

\rowcolor{Lavender!30}
\textbf{\texttt{PULP-HD}}~\cite{PULP-HD}
& \stthree~\stfive~\stsix
& Random \HV{} source with \textit{Record-based} and \textit{N-gram-based} composition for hardware-oriented encoding. \\

\textbf{\texttt{VoiceHD}}~\cite{voiceHD}
& \stone~\stthree~\stfive
& Random \HV{} source with binding and bundling operations for voice/time-series encoding. \\

\rowcolor{Lavender!30}
\textbf{\texttt{MHD}}~\cite{Imani_Hierarchical}
& \stthree~\stfive
& Random \HV{} source with \textit{Record-based} and \textit{N-gram-based} composition. \\

\textbf{\texttt{PerfHD}}~\cite{PerfHD}
& \stthree~\stfive
& Random \HV{} source with \textit{Record-based} and \textit{N-gram-based} composition. \\

\rowcolor{Lavender!30}
\textbf{\texttt{HyperSpec}}~\cite{HyperSpec}
& \stone~\stthree~\stfive~\stsix
& Random \HV{} source with \textit{Record-based} encoding and software-level encoder realization. \\

\textbf{\texttt{HDCluster}}~\cite{HDCluster}
& \stthree~\stfive~\stsix
& Random \HV{} source with \textit{Record-based} encoding and GPU/CUDA-oriented encoder realization. \\

\rowcolor{Lavender!30}
\textbf{\texttt{GraspHD}}~\cite{hassan2023efficient}
& \stone~\stthree~\stfive
& Random \HV{} source with \textit{Record-based}, spatial, and temporal composition. \\

\textbf{\texttt{uHD}}~\cite{uHD}
& \stthree~\stfive~\stsix
& LD-sequence \HV{} generation with bundling-oriented hardware realization. \\

\rowcolor{Lavender!30}
\textbf{\texttt{EmbHD}}~\cite{10.1145/3615991.3616404}
& \sttwo~\stthree~\stfive~\stsix
& Random and level \HV{s} with binding and bundling for microcontroller-oriented encoding. \\

\textbf{\texttt{PyBHV}}~\cite{PyBHV}
& \stthree
& Random \HV{} generation and symbolic \HV{} manipulation. \\ 

\rowcolor{Lavender!30}
\textbf{\texttt{HD Classification}}~\cite{HD-Classification}
& \stthree~\stfour~\stsix
& Random \HV{} generation and projection-oriented CPU/GPU encoder realization. \\ 

\textbf{\texttt{Constrained FSCIL}}~\cite{hersche2022cfscil}
& \stthree
& Quasi-random \HV{} generation for constrained learning settings. \\ 

\rowcolor{Lavender!30}
\textbf{\texttt{HDC Language Recognition}}~\cite{AbbasLang}
& \stone~\stthree~\stfive~\stsix
& Dense bipolar \HV{} generation with sequence-oriented language encoding and hardware realization. \\ 

\textbf{\texttt{Dynamic Vision Sensor with HDC}}~\cite{10.1145/3370748.3406560}
& \stone~\sttwo~\stfour
& Selective Kanerva coding~\cite{travnik2017representing} and randomized activation function for event-based encoding. \\

\rowcolor{Lavender!30}
\textbf{\texttt{HDC-MER}}~\cite{8771622}
& \stone~\stthree~\stfour
& Random \HV{} source with nonlinear mapping function. \\

\textbf{\texttt{EMG-based gesture recognition}}~\cite{moin2018emg}; \textbf{\texttt{Hand gesture recognition}}~\cite{Abbas_Biosignal}
& \stone~\stthree~\stfive
& Random \HV{} source with \textit{Record-based} and \textit{N-gram-based} composition. \\

\rowcolor{Lavender!30}
\textbf{\texttt{On-chip learning HDC library}}~\cite{abbas-remat}
& \stthree~\stfive~\stsix
& Random \HV{} source with \textit{Record-based} and \textit{N-gram-based} composition for on-chip encoder realization. \\

\textbf{\texttt{HD embedding-BCI}}~\cite{AbbasEmbedding}
& \stone~\sttwo~\stfour
& Feature quantization, random projection, and learned projection for biomedical encoding. \\

\rowcolor{Lavender!30}
\textbf{\texttt{HDC-EEG-ERP}}~\cite{10.1007/s11036-017-0942-6}
& \stone~\stthree~\stfive
& Random \HV{} source with binding, bundling, and permutation for EEG/ERP encoding. \\

\textbf{\texttt{HD pattern recognition}}~\cite{Abbas_Class_recall}
& \stthree~\stfive
& Random \HV{} source with bundling and permutation. \\

\rowcolor{Lavender!30}
\textbf{\texttt{HDC Parallel Single-pass Learning}}~\cite{yu2022understanding}
& \stthree~\stfour
& Gaussian \HV{} source with RBF-style mapping. \\

\textbf{\texttt{HDC for Out-of-Distribution Detection}}~\cite{10030789}
& \stthree~\stfour~\stfive
& Pseudo-random projection matrix with bundling-based class representation. \\

\rowcolor{Lavender!30}
\textbf{\texttt{HDC in Multiple-Input-Multiple-Output Neural Networks}}~\cite{menet2023mimonets}
& \stthree~\stfive
& Random keys with binding and bundling for multi-input neural/HDC integration. \\

\textbf{\texttt{HDC for Visual Abstract Reasoning}}~\cite{hersche2023probabilistic,neuro2023}
& \stone~\stthree~\stfive
& Random \HV{} source with binding and bundling for relational visual reasoning. \\

\bottomrule
\end{tabular}
}
\vspace{-0.5em}
\justify{\scriptsize{
Table emphasizes encoding-relevant aspects of prior HDC frameworks and implementations rather than sole software-platform statistics.Taxonomy symbols indicate stages in Fig.~\ref{fig:pipeline_taxonomy} are explicitly supported or emphasized by each framework: \stone~input-data modality, \sttwo~feature extraction, \stthree~atomic \HV{} source, \stfour~mapping, \stfive~composition, and \stsix~hardware realization.
}}
\vspace{-1.0em}
\end{table*}
\color{black}

\section{Extras: Available Platforms, Metrics, and SOTA Summary}
\label{extras}

\subsection{Development Platforms}

In what follows, we summarize recent platforms and highlight key prior works to inspire HDC researchers in refining encoding procedures.
Several open-source code projects on HDC have been released 
by various research 
groups. 
Table~\ref{hdc_imp} lists prior releases and their encoding methods by mapping each framework to the relevant stages of the proposed encoding taxonomy, including input-data modality~\stone, feature extraction~\sttwo, atomic \HV{} source~\stthree, mapping~\stfour, composition~\stfive, and hardware realization~\stsix.
For instance, 
the developers of \cite{torchHD, heddesTorch} implemented random, level, thermometer, circular, RBF kernel-based, \textit{N-gram-based}, and \textit{record-based} $\boldsymbol{\mathcal{HV}}$ encoding in Python using PyTorch. Another open-source project, 
\texttt{HDTorch}~\cite{hdTorch}, is also based on the PyTorch framework. \texttt{HDTorch} provides options for generating $\boldsymbol{\mathcal{HV}}$s using the \textit{random}, \textit{sandwich}, \textit{scale}, and \textit{scale with radius} approaches. Here, the sandwich approach refers to encoding two neighboring vectors similarly for half of their elements and setting the remaining elements randomly. The \textit{scale} method is an alternative term for level $\boldsymbol{\mathcal{HV}}$ generation, while \textit{scale with radius} utilizes a distance parameter called radius to encode vectors proportionally 
based on their proximity to the radius~\cite{sandwich}.
Kang et al.~\cite{openhd} propose \texttt{OpenHD}, an abstract framework designed for GPU-powered operations. 
Their framework allows designers to implement Python code, which is GPU-accelerated using the \texttt{HD} decorator and Just-In-Time (JIT) compilation~\cite{kang2022openhd, kang2022xcelhd}.
The \textit{Semantic Vectors Package} in \cite{semanticvectors} supports dense and sparse random vector encoding for concept-aware knowledge representation tasks in natural language processing. This package is implemented in both Java and Python. 
\texttt{Hyperdimensional Computing Library}~\cite{HDlib} 
is an open-source 
project offering 
\textit{N-gram-based} vector symbolic classification with both 
CPU and GPU-based optimizations. Additionally, a \texttt{PyTorch} implementation targeting few-shot learning with a combination of HDC systems and NNs is available in \cite{fewShot}. This repository employs quasi-random encoding for image data to obtain uncorrelated vectors \cite{hersche2022cfscil}. We highly recommend exploring the rest of the platforms for further investigation and insights. 



\vspace{-1em}
\subsection{Performance Metrics}

Evaluating HDC systems requires a complete view that balances accuracy, efficiency, and hardware friendliness. For encoding-focused evaluation, these metrics should also be interpreted with respect to the taxonomy stages that affect \HV{} quality, including feature extraction, atomic \HV{} source generation, mapping, composition, and hardware realization. The most widely used metric is classification accuracy, typically reported across standard datasets using cross-validation or test splits. Several recent works report additional confusion-matrix-derived metrics, including the F1 score, precision, recall/sensitivity, and specificity~\cite{Sobol_Aygun_TCAD}. HDC’s robustness, particularly under noise or reduced-precision conditions, is another key consideration. Metrics such as bit-flip resilience, performance under dimensionality ($D$) reduction, and tolerance to quantized operations (e.g., 4-bit counters) are recommended to highlight the overall robustness of the system~\cite{10908571}.

In addition to learning-based metrics, resource efficiency and latency are critical, especially for hardware deployment. Metrics such as inference latency, energy per classification, and hardware resource usage (e.g., LUTs, BRAMs, or GPU utilization) provide insight into suitability for edge or real-time systems. For example, memory footprint is often measured by the size of item memory and prototype vectors, and throughput is quantified as classifications per second. Since HDC often uses long $\boldsymbol{\mathcal{HV}}$s, it is recommended to consider latency, memory footprint, and memory-bandwidth-related costs. When evaluating encoding schemes specifically, similarity preservation (e.g., cosine or Hamming similarity) and orthogonality scores are used to assess encoding quality~\cite{gayler_vsa_2020}. These metrics ensure that encoded $\boldsymbol{\mathcal{HV}}$s retain useful semantic structure while remaining distinguishable, i.e., near-orthogonal.

\vspace{-1em}

\subsection{Selected SOTA Summary}

Finally, it is worth highlighting selected representative SOTA works with significant contributions to vector mapping and encoding. Table~\ref{extra_table} provides an overview of these works, emphasizing the relationship between application modality, dataset utilization, $\boldsymbol{\mathcal{HV}}$ generation, encoding strategy, and hardware efficiency.

\noindent\ding{171} Imani et al.~\cite{Imani_Hierarchical} 
reduce the classification cost by  
reducing the 
$\boldsymbol{\mathcal{HV}}$ size. 
They use a multi-encoder strategy to choose between \textit{record-based} or \textit{N-gram-based} encoding.
The specifics of vector generation on hardware remain undisclosed, with random generation utilizing nearly orthogonal vectors through bit-flipping. 

\noindent\ding{163} Nazemi et al.~\cite{10.1145/3400302.3415696}  
introduce the concept of NN-assisted HDC by
developing an 
automated feature extraction approach using NNs for 
HDC systems. 
Their approach employs hard-wired features and LUT-based levels via a DRAM module, with potential for further investigation into NN-based on-edge training independence.

\noindent\ding{102} Wu et al.~\cite{rahimi3D}  explore HDC's 
capacity with emerging device platforms such as nanotubes. Notably, in the realm of solid-state circuits, they present a CNFET (carbon nanotube field-effect transistor) and RRAM-based random projection unit for vector mapping, a unique contribution. 
Thus, further design efforts, following similar procedures with in-memory computing platforms, may yield fruitful outcomes.

\noindent\ding{161} Zou et al.~\cite{manihd}  propose an adaptive encoder facilitating non-linear interactions between features. This groundbreaking work transforms a static encoder into an adaptive one, aiming for maximum learning accuracy. They emphasize real-time encoding and learning properties, 
which are particularly crucial for streaming data and dynamic learning. 
They use an unsupervised manifold technique to multiply Gaussian-based random $\boldsymbol{\mathcal{HV}}$s with the manifold, creating an adaptive \texttt{HD} encoder.
Their approach enables 
reducing the $D$ size from $10{,}000$ to $4{,}000$ while maintaining the accuracy. 

\noindent\ding{72} 
Rasanen et al.~\cite{okko}  present a method for generating \HV{s} with local
similarities, improving generalization to new inputs.
The encoding maps quasi-orthogonal vectors for distant inputs by using raw data features, such as mel-frequency from speech data, to transform them into hyperspace.
The Weighted Accumulation of Random Projections (WARP) method and scatter code guide feature mapping for $\boldsymbol{\mathcal{HV}}$ generation. While the method introduces intriguing ideas for continuous data from sensory inputs, 
further discussion on hardware
implementation is needed, given the platform's resource limitations.

\noindent\ding{99} Chang et al.~\cite{8771622}  adopt a lightweight design for vector embedding using multiplexers. The generation step is simplified into a straightforward operation using the sign of the biomedical feature for the selection port of the multiplexer. 
Their study maps multimodal
sensor data into holistic representations.

\noindent\ding{86} 
Kim et al.~\cite{geniehd} propose chunk-processing ultra-long $\boldsymbol{\mathcal{HV}}$s to represent nucleotide structures in DNA sequences.
The approach focuses on parallelizing operations, especially for ultra-long $\boldsymbol{\mathcal{HV}}$s. 
Their method uses
approximate random generations and is not tied to specific hardware designs.
Notably, for applications with a limited number of 
symbols like DNA (Adenine, Guanine, Cytosine, and Thymine), the number of generated base $\boldsymbol{\mathcal{HV}}$s is limited, and it does not necessarily rely on dynamic vector generation. This work uniquely processes ultra-long vectors 
of up to $100{,}000$ dimensions.

\noindent\ding{69} Liu et al.~\cite{HDC-IM}  introduce an IMC technique with RRAM devices 
involving memory-based bit-flipping, replication, shifting, and catenation steps to generate both item memory and continuous item memory. 
Their method introduces a new
approach for handling vector generation in memory.

\begin{table*}
\centering
\caption{Summary of Noticeable SOTA Works Highlighting Applications, $\boldsymbol{\mathcal{HV}}$ Mapping, and Hardware}
\vspace{-0.75em}
\renewcommand{\arraystretch}{0.6}
\setlength{\aboverulesep}{0pt} 
\setlength{\belowrulesep}{0pt} 
\setlength{\extrarowheight}{0pt}
\resizebox{\linewidth}{!}{
\begin{tabular}{*{6}{l}} 
\toprule
\Large\textbf{Year, Authors} & \Large\textbf{Contribution} & \Large\begin{tabular}[l]{@{}l@{}}\textbf{Application,} \textbf{Dataset}\\\textbf{ (Accuracy, D)}\end{tabular} & \Large\textbf{Encoding} & \Large \begin{tabular}[l]{@{}l@{}}\textbf{\textit{Environment}, \textit{Model},}\\\textbf{ or \textit{Hardware}}\textbf{\textit{ }Details}\end{tabular} & \Large \begin{tabular}[l]{@{}l@{}}\textbf{\textit{Hardware}} \\\textbf{Efficiency}\end{tabular} \\ 
\midrule

\rowcolor{Lavender!30}
\Large \begin{tabular}[l]{@{}l@{}}2015, \\Räsänen \ding{72} \\ \cite{okko} \end{tabular} &\Large \begin{tabular}[l]{@{}l@{}}Vector generation for \\non-discrete data\end{tabular} &\Large \begin{tabular}[l]{@{}l@{}}Spoken word \\classification\\CAREGIVER Y2 UK corpus\\($>97$\%, $D$=4,000)\end{tabular} &\Large \begin{tabular}[l]{@{}l@{}}S-WARP Mapping and\\ Weighted Accumulation \\ of Random Projections\\ \blue{\sttwo~\stfour}\end{tabular} &\Large \begin{tabular}[l]{@{}l@{}}No information \\ given\end{tabular} &\Large \begin{tabular}[l]{@{}l@{}}No hardware \\implementation\end{tabular} \\ 

\Large \begin{tabular}[l]{@{}l@{}}2018, \\Wu et al. \ding{102} \\ \cite{rahimi3D} \end{tabular} &\Large \begin{tabular}[l]{@{}l@{}}Exploring the opportunities of \\ HDC for next generation \\ devices (nanotubes, RRAM)\end{tabular} &\Large \begin{tabular}[l]{@{}l@{}}21 European languages \\classification Europarl \\(98\%)\end{tabular} &\Large \begin{tabular}[l]{@{}l@{}}Evenly spaced \\time-delay \\encoding\\ \blue{\stfour~\stfive~\stsix}\end{tabular} &\Large \begin{tabular}[l]{@{}l@{}}Synopsys \\ Design Compiler \\CNFET and \\ RRAM-based design\end{tabular} &\Large \begin{tabular}[l]{@{}l@{}}\# of CNFET\\ (1952) \\\# of RRAM\\ (224)\end{tabular} \\ 

\rowcolor{Lavender!30}
 \Large \begin{tabular}[l]{@{}l@{}}2018, \\Imani et al. \ding{171} \\ \cite{Imani_Hierarchical} \end{tabular} & \Large \begin{tabular}[l]{@{}l@{}}Decreasing \\ classification cost\end{tabular} & \Large \begin{tabular}[l]{@{}l@{}}Speech recognition \\ISOLET \\(95.9\%, $D$=10,000)\end{tabular} & \Large \begin{tabular}[l]{@{}l@{}}Random-Bipolar $\boldsymbol{\mathcal{HV}}$ \\Bit-flipping applied\\  Hierarchical multi-encoder: \\I. \textit{Record-based} \\II. \textit{N-gram-based}\\ \blue{\stthree~\stfive~\stsix}\end{tabular} & \Large \begin{tabular}[l]{@{}l@{}}Synopsys \\ Design Compiler \\(TSMC 45nm)\end{tabular} & \Large \begin{tabular}[l]{@{}l@{}}Energy (6.6×) \\Speed (6.3×)\end{tabular} \\ 

\Large \begin{tabular}[l]{@{}l@{}}2019, \\Chang et al. \ding{99} \\ \cite{8771622} \end{tabular} &\Large \begin{tabular}[l]{@{}l@{}}$\boldsymbol{\mathcal{HV}}$ embedding using\\ random non-linearity \end{tabular} &\Large \begin{tabular}[l]{@{}l@{}}Galvanic Skin Response, \\Electroencephalogram, \\Electrocardiogram \\dataset (AMIGOS) \\binary emotion classification \\(76.6\%, $D$=10,000)\end{tabular} &\Large \begin{tabular}[l]{@{}l@{}}Multiplexer-based \\ ternary (-1, 0, +1) \\ feature mapping \\ into hyperspace\\ \blue{\sttwo~\stfour}\end{tabular} &\Large \begin{tabular}[l]{@{}l@{}}No information \\ given\end{tabular} &\Large \begin{tabular}[l]{@{}l@{}}Only system block\\ diagram is given\end{tabular} \\

\rowcolor{Lavender!30}
\Large \begin{tabular}[l]{@{}l@{}}2019 \\Liu et al. \ding{69} \\ \cite{HDC-IM} \end{tabular} & \Large \begin{tabular}[l]{@{}l@{}}RRAM-based HDC\\ in-memory architecture\end{tabular} &\Large \begin{tabular}[l]{@{}l@{}} ISOLET \\(92.1\%, $D$=8,192)\end{tabular} &\Large \begin{tabular}[l]{@{}l@{}}Block generation \\ of $\boldsymbol{\mathcal{HV}}$s on crossbar,\\ \texttt{XOR}-summation encoding\\ \blue{\stthree~\stfive~\stsix}\end{tabular} &\Large \begin{tabular}[l]{@{}l@{}}RRAM, CPU, \\ and GPU separately\end{tabular} &\Large \begin{tabular}[l]{@{}l@{}}2.28 J Energy\\ (in-memory)\\ 3960 J Energy (CPU)\end{tabular} \\ 
 
\Large \begin{tabular}[l]{@{}l@{}}2020, \\Nazemi et al. \ding{163} \\ \cite{10.1145/3400302.3415696} \end{tabular} & \Large \begin{tabular}[l]{@{}l@{}}Automatic \\ feature extraction\end{tabular} & \Large \begin{tabular}[l]{@{}l@{}}HAR (96.44\%, $D$=10,240) \\ISOLET (96.67\%, $D$=10,240)\end{tabular} &\Large \begin{tabular}[l]{@{}l@{}}Random seed $\boldsymbol{\mathcal{HV}}$ \\Bit-flipping \\~ Binding and \\Bundling\\ \blue{\sttwo~\stthree~\stfive~\stsix}\end{tabular} &\Large \begin{tabular}[l]{@{}l@{}}FPGA (Xilinx\\ UltraScale+ VU9P)\end{tabular} &\Large \begin{tabular}[l]{@{}l@{}}BRAMs (1.8\%) \\DSPs (15\%) \\FFs (0.8\%) \\LUTs (5.1\%) \\Latency (23.3 us) \\Power (5.3 W)\end{tabular} \\ 

\rowcolor{Lavender!30}
\Large \begin{tabular}[l]{@{}l@{}}2020, \\Kim et al. \ding{86} \\ \cite{geniehd} \end{tabular} &\Large \begin{tabular}[l]{@{}l@{}}DNA representation \\ in high-dimensional\\ $\boldsymbol{\mathcal{HV}}$s and \\ accelerated pattern match\end{tabular} &\Large \begin{tabular}[l]{@{}l@{}}DNA Pattern Matching\\ *Escherichia coli\\ *Human chromosome14\\ *Random sequence\\ (\textgreater{}99.99\%, $D$=100,000)\end{tabular} &\Large \begin{tabular}[l]{@{}l@{}}Random-Bipolar \\ base $\boldsymbol{\mathcal{HV}}$s, Shifting and \\ multiplication-based \\encoding\\ \blue{\stthree~\stfive~\stsix}\end{tabular} &\Large \begin{tabular}[l]{@{}l@{}}GPU \\ (NVIDIA 1080 Ti)\\ and\\ FPGA\\ (Kintex-7 KC705)\end{tabular} &\Large \begin{tabular}[l]{@{}l@{}}44.4$\times$ speedup\\ 54.1$\times$ energy-\\ efficiency\\ on FPGA\end{tabular} \\ 

\Large \begin{tabular}[l]{@{}l@{}}2021, \\Zou et al. \ding{161} \\ \cite{manihd} \end{tabular} &\Large \begin{tabular}[l]{@{}l@{}}Adaptive encoder \\ and non-linear interactions \\ between the features\\ during the encoding\end{tabular} &\Large \begin{tabular}[l]{@{}l@{}}MNIST (\textgreater{}97\%) PECAN (\textgreater{}97\%)\\  ISOLET (\textgreater{}95\%) EXTRA (\textgreater{}84\%) \\UCIHAR (\textgreater{}98\%)  \\($D$ = 4,000 for all)\end{tabular} &\Large \begin{tabular}[l]{@{}l@{}}Manifold \\ selection-based \\ encoding, Gaussian dist. \\ for initial vectors\\ \blue{\stthree~\stfour~\stsix}\end{tabular} &\Large \begin{tabular}[l]{@{}l@{}}In software: C++ \\~ In hardware: FPGA\\ (Kintex-7)\end{tabular} &\Large \begin{tabular}[l]{@{}l@{}}12.3× speed-up \\19.3× energy-\\ efficiency\end{tabular} \\ 

\bottomrule
\end{tabular}
}
\label{extra_table}
\vspace{-1.5em}
\end{table*}

Looking ahead, several promising directions emerge for the continued advancement of HDC, particularly in integrating it with ML paradigms~\cite{info15080490} and optimizing hardware~\cite{sparsehd}. However, in line with the encoding-centered scope of this survey, we keep this discussion limited to directions that directly affect $\boldsymbol{\mathcal{HV}}$ generation, mapping, composition, or encoder realization. On the ML front, a key opportunity lies in unifying HDC with deep learning architectures. Hybrid models that leverage HDC’s lightweight and noise-tolerant encoding, combined with NNs' representation learning, could lead to more efficient, interpretable, and robust AI systems. Recent works have started exploring the use of neural feature extractors to guide HDC encoding, and further exploration of end-to-end trainable HDC pipelines remains an open research frontier. These directions are most relevant here when NN-assisted feature extraction, adaptive encoders, or learned projections improve the separability, similarity preservation, or robustness of the resulting $\boldsymbol{\mathcal{HV}}$ representations. On the hardware side, the increasing adoption of HDC in IMC, RRAM, and FPGA platforms motivates deeper investigation into encoding schemes that are both accuracy-preserving and hardware-friendly. Vector generation schemes that minimize permutation or bundling costs and bit-stream-compatible representations will be crucial for scalable and low-energy deployment. Techniques such as structured randomization, quantized binding, and bit-level parallelism offer fertile ground for optimization. Similarly, emerging paradigms such as quantum-enhanced HDC are relevant to this survey only when they improve encoding-related operations, such as the manipulation, decomposition, or search of composite $\boldsymbol{\mathcal{HV}}$s~\cite{Cumbo2026,10655760,poduval2024hyperdimensionalquantumfactorization}.

Lastly, expanding HDC’s application space, particularly in multi-modal learning, symbolic reasoning, privacy-preserving AI, and real-time edge inference, has potential and will require encoding strategies that are adaptable, interpretable, and capable of continual learning. To avoid a broad application-level discussion, we emphasize that these directions should be evaluated through concrete encoding-level questions: \textit{How are input modalities converted into features?} \textit{How are atomic $\boldsymbol{\mathcal{HV}}$s generated?} \textit{How do mapping and composition preserve useful structure?} and \textit{How can the resulting encoders be deployed efficiently in hardware?} Therefore, future SOTA efforts should report not only accuracy but also encoding quality, orthogonality, memory footprint, latency, energy, and hardware efficiency.

\begin{table*}[!t]
\centering
\scriptsize
\caption{Glossary of Key Terms in Hyperdimensional Computing (HDC)}
\vspace{-1.5em}
\label{tab:hdc_glossary}
\setlength{\aboverulesep}{0pt} 
\setlength{\belowrulesep}{0pt} 
\setlength{\extrarowheight}{0pt}
\begin{tabular}{p{3.5cm}p{11.6cm}} 
\toprule
\textbf{Term} & \textbf{Description} \\
\midrule
\rowcolor{Lavender!30}
\textbf{Associative Memory (AM)} & A memory for storing learned class hypervectors used for similarity-based classification. \\
\textbf{Binary Spatter Coding} & A foundational VSA model using high-dimensional dense binary vectors; binding is performed via \texttt{XOR} and bundling via majority rule. \\
\rowcolor{Lavender!30}
\textbf{Binarization} & Transforms real-valued bundles into bipolar form (e.g., \texttt{sign()} to get $\pm 1$) or into binary form (e.g., majority thresholding to get $0$ or $1$). \\
\textbf{Binding} & Combines hypervectors (e.g., via XOR, circular convolution) to represent relationships or associations. \\
\rowcolor{Lavender!30}
\textbf{Bit-Flipping Encoding} & Generates hypervectors for neighboring scalar values by flipping a small number of bits. \\
\textbf{Bundling} & Aggregates multiple hypervectors into one, typically by majority vote or summation + sign thresholding. \\
\rowcolor{Lavender!30}
\textbf{Capacity} & The number of distinguishable patterns that can be stored/retrieved using a given dimension size $D$. \\
\textbf{Context Hypervector} & An encoded vector that represents temporal or positional context in sequences. \\
\rowcolor{Lavender!30}
\textbf{Contextual Binding} & Combines information with contextual cues (e.g., time or position) using binding + permutation. \\
\textbf{Dense Vector} & A hypervector with an equal mix of 1s and 0s (or +1s and -1s); typical in symbolic representation. \\
\rowcolor{Lavender!30}
\textbf{Dimensionality ($D$)} & Number of elements in a hypervector; higher $D$ increases robustness and capacity. \\
\textbf{Error Robustness} & HDC's tolerance to bit-flips and hardware faults due to redundancy in high-dimensional representations. \\
\rowcolor{Lavender!30}
\textbf{Fractional Power Encoding} & Encodes spatial (2D) information using powered base vectors \cite{KleykoPART1}. \\
\textbf{Hardware Efficiency} & HDC enables low-latency, energy-efficient operations using logic gates or in-memory computing. \\
\rowcolor{Lavender!30}
\textbf{Holographic Reduced Representation} & A foundational VSA model that utilizes continuous vectors and relies on circular convolution for the binding operation. \\
\textbf{Hypervector ($\boldsymbol{\mathcal{HV}}$)} & A high-dimensional vector (1K–10K bits) used to encode input data in HDC. Can be binary, bipolar, sparse, or dense. \\
\rowcolor{Lavender!30}
\textbf{Item Memory (IM)} & A memory storing fixed atomic hypervectors representing basic input features (e.g., letters, pixel locations). \\
\textbf{Multi-modal Encoding} & Combines different data types (text, image, time-series) into a unified hypervector space. \\
\rowcolor{Lavender!30}
\textbf{On-chip Learning} & Refers to real-time HDC training on hardware platforms (e.g., FPGA, RRAM). \\
\textbf{Orthogonality} & Degree of dissimilarity between hypervectors; ensures distinct features/symbols don’t interfere. \\
\rowcolor{Lavender!30}
\textbf{Permutation ($\rho$)} & Shifts elements in a hypervector to encode order/sequence (e.g., in $N$-gram modeling). \\
\textbf{Privacy-aware HDC} & Uses encoded data that prevents raw input exposure, enabling privacy-preserving computation. \\
\rowcolor{Lavender!30}
\textbf{Quasi-Random Sequences} & Used in some encodings (e.g., Sobol) to achieve low-discrepancy and near-orthogonality in fewer steps. \\
\textbf{Radial Basis Function (RBF)} & A kernel-inspired mapping that preserves similarity among continuous-valued inputs in the hyperdimensional space. \\
\rowcolor{Lavender!30}
\textbf{Random Seed} & An initial vector used to generate orthogonal/correlated hypervectors through structured randomness. \\
\textbf{Scalar Encoding} & Encoding continuous/numeric values using methods like thermometer codes, bit-flipping, or RBF projections. \\
\rowcolor{Lavender!30}
\textbf{Similarity Search} & Compares a test hypervector to stored class vectors (e.g., via cosine similarity or Hamming distance). \\
\textbf{Sparse Distributed Memory} & A neural-like model proposed by Kanerva for storing high-dimensional vectors using associative mechanisms. \\
\rowcolor{Lavender!30}
\textbf{Sparse Vector} & A hypervector with mostly 0s (or -1s), used for correlated numeric encodings. \\
\textbf{Symbolic Encoding} & Encoding categorical data (e.g., letters) into orthogonal or randomly assigned vectors. \\
\rowcolor{Lavender!30}
\textbf{Thermometer Encoding} & A scalar-to-vector encoding scheme where increasing values correspond to longer runs of 1s. \\
\textbf{Training Modes} & HDC models support single-pass (online) learning or batch-based multi-pass training. \\
\rowcolor{Lavender!30}
\textbf{Vector Generation} & The process of creating hypervectors using various encoding methods based on application and data type. \\
\textbf{Vector Symbolic Architecture} & A class of cognitive architectures using high-dimensional vectors to perform logic and reasoning. \\
\bottomrule
\end{tabular}
\vspace{-2em}
\end{table*}

\section{Conclusions}
\label{conclusions}
This study provides a comprehensive exploration of hypervector encoding, the primary and pivotal stage in Hyperdimensional Computing (HDC), using various methods and techniques, with a particular focus on the hypervector generation phase. Despite the promising potential of HDC, there has been relatively limited study of the critical step of hypervector generation, namely the initial stage of hypervector encoding. This encoding plays a fundamental role in the design of the architecture and the accuracy of HDC systems. Previous state-of-the-art (SOTA) research has focused on the holistic structure of HDC. Less attention has been given to exploring its potential for cross-paradigm computations, such as bit-stream processing, during the hypervector generation or initial vector mapping phase. In this study, we addressed this gap by providing an in-depth literature analysis of HDC and introducing readers to a structured perspective on initial vector mapping and encoding. We also explore the interplay between HDC and other emerging paradigms, emphasizing their effectiveness and the importance of this research for advancing the field. Our key contributions include: (i) a systematic categorization of encoding in HDC, supported by formal representation, operation conventions, and multi-stage encoding pipeline taxonomy, (ii) detailed comparative insights into symbolic and scalar encoding strategies, and (iii) a curated synthesis of state-of-the-art implementations with hands-on considerations, including platforms, evaluation metrics, selected SOTA comparisons, and hardware-aware implementation aspects. In addition, we recommend several forward-looking research directions, such as hybrid HDC-Machine Learning architectures, hardware-aware encoding optimizations, and privacy-preserving on-device learning.

\vspace{-0.25em}
\section*{Acknowledgments}
This work was supported by the National Science Foundation (NSF) under Grants 2616837, 2339701, 2609436, 2127780, and  2019511; NASA grants 80NSSC25C0335 and 80NSSC26C0321; the Semiconductor Research Corporation (SRC) Task under Grant 2988.001; the Department of the Navy, Office of Naval Research under Grants N00014-21-1-2225 and N00014-22-1-2067; the Air Force Office of Scientific Research; Vernon and Ruby Langlinais Non-Endowed Research Fund; Lockheed Martin Corporation Endowed Professorship Fund; and generous gifts from Cisco, Xilinx, NVIDIA, and Google (\textit{for education}).

\section*{Appendix}
\label{appendix:glossary}
 \textbf{\textit{Glossary}.} Table~\ref{tab:hdc_glossary} provides a glossary of key terms used in the HDC literature. 
\\
\\
\textbf{\textit{Datasets in Use for HDC Systems.}}
\label{appendix:2}
HDC systems have been predominantly applied to domains such as biomedical/biometrics, image and text processing, and continuous sensor data, including audio signals (e.g., \texttt{VoiceHD}~\cite{voiceHD}).
Commonly used datasets in prior HDC works include (i) \textit{UCIHAR} (Human Activity Recognition Using Smartphones \cite{UCIHAR}) like \cite{training-hdc}, \cite{imani-efficient}, \textbf{\texttt{THRIFTY}} \cite{THRIFTY}, \textbf{\texttt{MulTa-HDC}} \cite{MulTa-HDC}, \cite{10.1007/978-3-030-79150-6_35}, \cite{re-hfdc}, \cite{yu2022understanding}, (ii) \textit{ISOLET} (Spoken Letters \cite{misc_isolet_54}) like \cite{Kazemi_2022}, \textbf{\texttt{HDnn-PIM}} \cite{hdnn}, \textbf{\texttt{tiny-HD}} \cite{BehnamIoT}, (iii) \textit{PAMAP2} (Physical Activity Monitoring, \cite{misc_pamap2_physical_activity_monitoring_231}) like \textbf{\texttt{OpenHD}} \cite{openhd}, \textbf{\texttt{QuantHD}} \cite{QuantHD}, \textbf{\texttt{HDTorch}} \cite{hdTorch}, (iv) \textit{FACE} (Caltech256 dataset \cite{griffin_holub_perona_2022}) like \textbf{\texttt{AdaptHD}} \cite{adapthd}, \cite{THRIFTY}, \cite{openhd}, (v) \textit{MNIST} (Modified National Institute of Standards and Technology database \cite{726791}) like \cite{9195216}, \cite{LeHDC}, \cite{QuantHD}, \cite{ma2022hyperdimensional}, \cite{yan2023efficient}, \cite{duan2022braininspired}, \cite{10024980}, (vi) \textit{CARDIO} (Cardiotocography \cite{misc_cardiotocography_193}) like \cite{THRIFTY}, \textbf{\texttt{Store-n-Learn}} \cite{10.1145/3503541}, \cite{10050560}, (vii) CIFAR (Canadian Institute For Advanced Research \cite{Krizhevsky2009LearningML}) \cite{ma2022hyperdimensional}, \cite{Mehran-No-mul}, \cite{symbolichdc}, \cite{hdnn}, \cite{gluing}. 
Biomedical data processing applications with their specific datasets dominate the HDC use, with examples like 
EEG (electroencephalogram) \cite{10.3389/fneur.2022.816294, Abbas-ExG, 10.1007/s11036-017-0942-6}, ECG (electrocardiogram) \cite{8771622}, iEEG (intracranially recorded EEG signals) \cite{10.3389/fneur.2021.701791}, EMG (Electromyography) \cite{Abbas_Biosignal}, GSR (Galvanic Skin Response) \cite{Menon2022}, and ECoG (electrocorticography)~\cite{Abbas-ExG}. 


\bibliographystyle{unsrt}

\bibliography{sample-base, bibliography, hassan}


\end{document}